\documentclass[preprint,review,12pt]{elsarticle}

\usepackage{amssymb}
\usepackage{subcaption}
\usepackage{booktabs}
\usepackage{times}
\usepackage{rotating}
\usepackage[dvipsnames]{xcolor}
\usepackage{graphicx}
\usepackage{amssymb}

\usepackage{amsmath}
\usepackage{algorithm}

\usepackage{algpseudocode}
\usepackage{url}
\usepackage{multirow}
\usepackage{pdflscape}
\usepackage{tcolorbox}

\usepackage{float} 

\usepackage{booktabs}
\usepackage{xcolor}
\definecolor{lightblue}{rgb}{0.90,0.95,1}

\journal{Image and Vision Computing}
\usepackage{cleveref}
\usepackage{comment}

\newif\ifshowcomments
\showcommentstrue

\newcommand{\Rev}[1]{\ifshowcomments \textcolor{black}{#1}\fi}

\begin{document}

\title{Assessing the Noise Robustness of Class Activation Maps: A Framework for Reliable Model Interpretability} %% 

\author{Syamantak Sarkar$^\dagger$ \quad  Revoti P. Bora$^\ddagger$ \quad Bhupender Kaushal$^\dagger$  \\ Sudhish N George$^\dagger$ \quad  Kiran Raja$^\ddagger$ \\
$^\dagger$National Institute of Technology Calicut, India; 
$^\ddagger$NTNU Gjøvik, Norway}
\maketitle
% \author{Syamantak Sarkar\\
% National Institute of Technology, Calicut\\
% NIT Campus P.O, Calicut Mukkam Road, Kattangal, Kozhikode, Kerala 673 601\\
% {\tt\small syamantak_m230922ec@nitc.ac.in}
% \and
% Bhupender Kaushal\\
% National Institute of Technology, Calicut\\
% NIT Campus P.O, Calicut Mukkam Road, Kattangal, Kozhikode, Kerala 673 601\\
% {\tt\small bhupender_p220180ph@nitc.ac.in}
% \and
% Sudhish N George\\
% National Institute of Technology, Calicut\\
% NIT Campus P.O, Calicut Mukkam Road, Kattangal, Kozhikode, Kerala 673 601\\
% {\tt\small sudhish@nitc.ac.in}
% \and
% Revoti Prasad Bora\\
% Norwegian University of Science and Technology\\
% Høgskoleringen 1, 7034 Trondheim, Norway\\
% {\tt\small revoti.p.bora@ntnu.no}
% \and
% Kiran Raja\\
% Norwegian University of Science and Technology\\
% Høgskoleringen 1, 7034 Trondheim, Norway\\
% {\tt\small kiran.raja@ntnu.no}
% }
 %% Author name

% \affiliation{organization={},%Department and Organization

%% Abstract
\begin{abstract}
Class Activation Maps (CAMs) are one of the important methods for visualizing regions used by deep learning models. Yet their robustness to different noise remains underexplored. In this work, we evaluate and report the resilience of various CAM methods for different noise perturbations across multiple architectures and datasets. By analyzing the influence of different noise types on CAM explanations, we assess the susceptibility to noise and the extent to which dataset characteristics may impact explanation stability. The findings highlight considerable variability in noise sensitivity for various CAMs. \Rev{We propose a robustness metric for CAMs that captures two key properties: consistency and responsiveness. Consistency reflects the ability of CAMs to remain stable under input perturbations that do not alter the predicted class, while responsiveness measures the sensitivity of CAMs to changes in the prediction caused by such perturbations.} The metric is evaluated empirically across models, different perturbations, and datasets along with complementary statistical tests to exemplify the applicability of our proposed approach.
\end{abstract}

\section{Introduction}
\label{sec1}

Despite the success of Deep Learning (DL) models in achieving high predictive accuracy, their black-box nature limits transparency and interpretability, raising concerns in high-stake domains \cite{rudin2019stop}. Thus, explainability has become a crucial aspect not only to provide insight into the decision-making process of a model but also to support its validation and enhance user trust \cite{MARKUS2021103655}. A popular category of explainability methods is Class Activation Mapping (CAM) which highlights certain regions in an image that the model finds most relevant for its prediction. CAM-based techniques such as Grad-CAM \cite{selvaraju2017grad}, Grad-CAM++ \cite{Chattopadhyay2018GradCAM++}, xGrad-CAM \cite{Fu2020xGradCAM}, Ablation-CAM \cite{Desai2020AblationCAM}, HiResCAM \cite{Dovrat2019HiResCAM} and Eigen-CAM \cite{Muhammad2020EigenCAM} localize important regions used by DL models to produce heat-maps for explanations. These methods vary in their approach in calculating the importance of regions, which eventually enhances their applicability to tasks that require explanations \cite{poppi2021revisiting}.

\begin{figure*}[!t]
    \centering
    \includegraphics[width=0.95\textwidth]{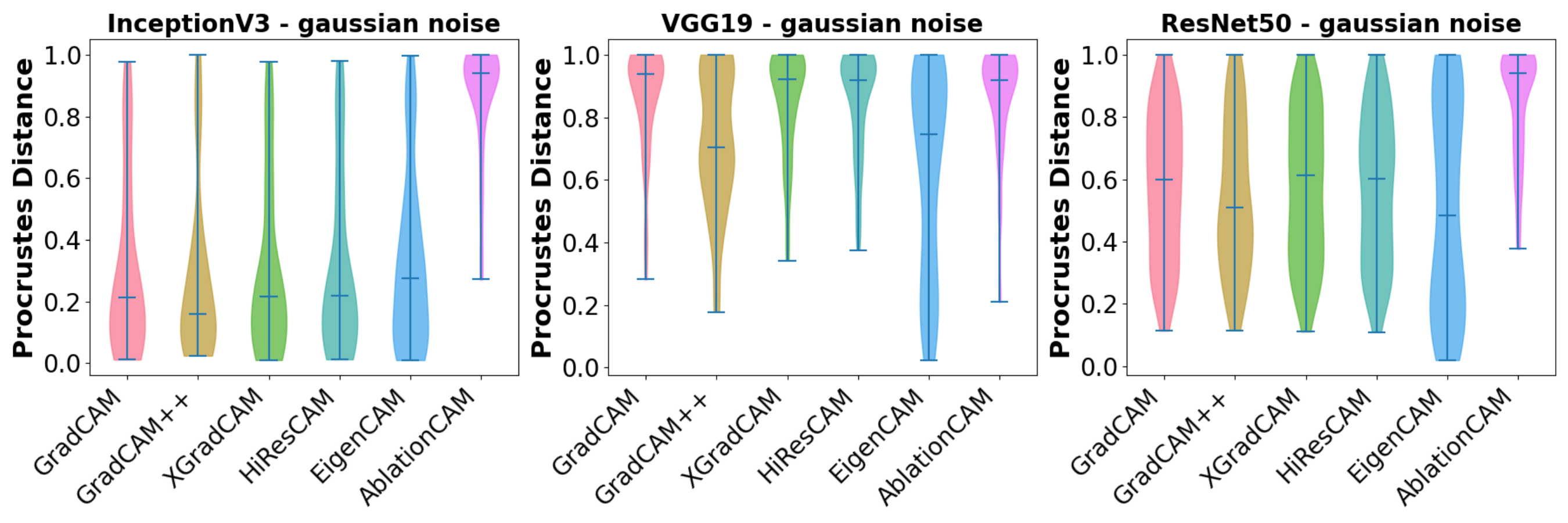}
    \caption{\textit{Procrustes distance computed for different networks using Gaussian noise}. The weight space in plots shows interesting trends with fluctuations in Procrustes distance distribution when evaluated with multiple noise values.}
    \label{fig:l2-distance}
\end{figure*}

\begin{figure}[!t]
    \centering
    \includegraphics[width=0.85\linewidth]{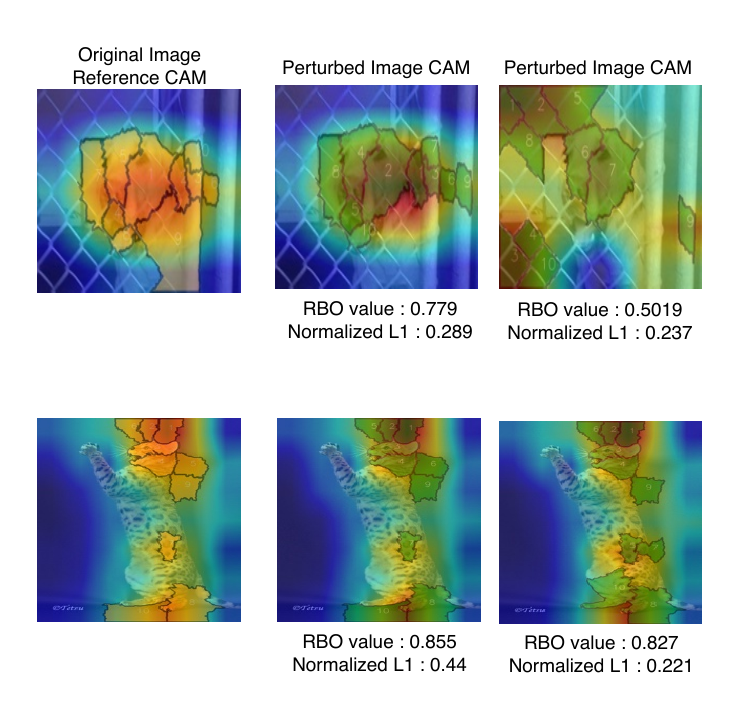}
    \caption{\textit{Comparison of Stability Ratio ($l_1$) \cite{agarwal2022rethinking} and RBO values}. The reference image is without perturbation, whereas others are perturbed. The RBO metric tends to perform relatively stable in both cases and the Stability Ratio, i.e., normalized $l_1$\cite{agarwal2022rethinking} deviates significantly.}
    \label{fig:new_istance}
\end{figure}

However, despite their utility, CAM-based methods can be sensitive to noise and perturbations in input images, raising questions about their robustness under varying conditions \cite{ijcai2020p726, khakzar2020rethinking}. As shown in Figure~\ref{fig:l2-distance}, we observe a variation in the weight space across different CAM methods, where the Procrustes distance \footnote{Procrustes distance is a metric that quantifies the dissimilarity between two shapes by optimally aligning them through translation, rotation, and scaling.}\cite{andreella2023procrustes} between the activation values for original and noisy images differs across various models. Robustness is therefore crucial for explainability, particularly in CAM-based methods, where sensitivity to minor input variations can lead to inconsistent explanations as seen in Figure~\ref{fig:new_istance}. Previous works have emphasized the need for stability in explanations; for example, Sanjoy \textit{ et al.} \cite{Sanjoy2022} defined an explanation framework with a prediction function \( f: X \rightarrow Y \) and an explanation function \( e: X \rightarrow E \), while Melis \textit{et al.} \cite{Alvarez2018Stability} formalized a stability metric, stressing that minor input perturbations should not cause significant changes in the explanations.

Typically, robustness is assessed by perturbing an instance \( x \) to produce \( x' \) with minor noise that does not alter the prediction, i.e., $ \hat{y}_x = \hat{y}_{x'}$ where $\hat{y}_x$ and $\hat{y}_{x'}$ are image labels of $x$ and $x'$ respectively.  A stability ratio \( S(x, x', e_x, e_{x'}) \) as given in \cite{agarwal2022rethinking} is used to measure the consistency of the explanations for perturbed inputs as represented in Eqn~\ref{eq:stability}.

\begin{equation}
{
S(x, x', e_x, e_{x'}) = \max_{x'} \frac{\| e_x - e_{x'} \|}{\| x - x' \|},\quad \forall x' \in N_x; \, \hat{y}_x = \hat{y}_{x'}\label{eq:stability}
}
\end{equation} where, ${\|e_{x}-e_{x'}\|}$ is the $l_1$ distance between explanations generated from the original image and the noisy version of the same image.

Moreover, from the top row of Figure~\ref{fig:new_istance}, it can be seen that there is a significant change in the overall intensity of the CAM explanation; notable variations in specific regions and their ranks can be observed. A significant change in CAM intensity results in shifts in the segment ranks, leading to a low Rank-Biased Overlap (RBO) value~\cite{Webber2010RBO}. However, the normalized $l_1$ values do not show variations consistent with the change. In contrast, the bottom row of Figure~\ref{fig:new_istance} shows minimal changes in CAM intensity, keeping segment ranks stable and yielding a high RBO value. Although normalized $l_1$ values show a significant change in this case, raising validity concerns for assessing CAM robustness.

\begin{table}[H]
\centering
\caption{\Rev{Comparison of proposed method with existing saliency map evaluation methods}}
\label{tab:robustness_comparison}
% \begin{tcolorbox}[colframe=blue, colback=lightblue, boxrule=1pt, sharp corners, enhanced jigsaw,]
\resizebox{\textwidth}{!}{
\begin{tabular}{|p{6cm}|p{3cm}|p{3cm}|p{5cm}|}
\hline
\textbf{Method} & \textbf{Model Retraining Required?} & \textbf{Label Dependency} & \textbf{Key Limitations} \\
\hline
\textbf{Proposed} & No & No & Requires segmentation of salient regions; does not evaluate ground-truth alignment \\
\hline
Gradient-based saliency estimation~\cite{selvaraju2017grad} & No & No & Sensitive to gradient noise; unstable under gradient saturation \\
\hline
Relevance-based methods (e.g., LRP, Deep Taylor)~\cite{montavon2017explaining} & No & No & Dependent on root/reference point; unstable attributions \\
\hline
Perturbation-based fidelity evaluation~\cite{bach2015pixel, samek2017evaluating, arras2017relevant} & No & Yes & Requires repeated perturbations; computationally expensive; depends on model output \\
\hline
In-fidelity and sensitivity metrics~\cite{yeh2019infidelity} & No & No & Mainly for gradient-based methods; limited applicability to CAMs \\
\hline
Sanity checks (weight/label randomization)~\cite{adebayo2018sanity} & No & No & Tests sensitivity to parameters but not robustness to data perturbations \\
\hline
Algorithmic stability measures~\cite{fel2021algorithmic} & Yes & No & Requires model retraining; computationally expensive \\
\hline
Stability Ratio (ROAR)~\cite{hooker2019benchmark} & Yes & Yes & Model-specific; computationally expensive; retraining required \\
\hline
KAR~\cite{hooker2019benchmark} & Yes & Yes & Requires retraining; accuracy drop depends on labeled data \\
\hline
Normalized $\ell_1$ Distance & No & No & Captures pixel intensity difference but ignores semantic rank changes \\
\hline
LEAF~\cite{samek2021explaining} & No & Yes & Requires multiple forward passes; less interpretable \\
\hline
\end{tabular}
}
% \end{tcolorbox}
\end{table}

While several prior works by Khakzar \textit{et al.}\cite{khakzar2020rethinking}, Bansal \textit{et al.}\cite{bansal2020sam}, Nourelahi \textit{et al.}\cite{nourelahi2022explainable}, and Adebayo \textit{et al.}\cite{NEURIPS2018_294a8ed2} have advanced our understanding of saliency and CAM methods, they primarily rely on qualitative assessments or input/parameter randomization and lack quantitative tools to evaluate explanation consistency under perturbation. \Rev{In contrast, our framework is designed to assess the robustness of CAM outputs by analyzing the consistency of explanations for small perturbations that do not change the predicted class and the change in the explanations (i.e, responsiveness) for perturbations that change the predicted class.}

\Rev{Unlike methods such as Stability Ratio~\cite{hooker2019benchmark}, which measure model-level robustness and require retraining, or Sensitivity Analysis~\cite{yeh2019fidelity}, which captures raw saliency variation without considering region ranking, our method evaluates rank-based shifts in salient regions without retraining or label dependency. Furthermore, while normalized $\ell_1$ distance captures pixel-wise intensity changes, it fails to detect semantic rank instability, as also highlighted in Table~\ref{tab:robustness_comparison}.}

\Rev{Motivated by this, we propose a metric to evaluate the robustness of CAMs by making use of different image perturbations and computing the variation of segment importance ranks\footnote{The segment importance ranks are calculated by considering the mean intensity of the CAM region projected onto the different segments.} for the original and perturbed version of the image. 
Specifically, we introduce a method that evaluates the variation in segment importance rankings under two scenarios, when the predicted class remains the same and when it changes, to assess the robustness of CAMs. The proposed robustness metric is validated by studying various perturbations. We define the robustness of a Class Activation Mapping (CAM) method in terms of two key properties as described below:
\begin{itemize}
    \item \textit{Consistency} refers to the ability of a CAM method to produce stable segment importance rankings ($R$) when small perturbations are introduced that do not alter the model's prediction probability. Formally, for any perturbation $n$ where the prediction remains unchanged ($P = P^n$), a consistent CAM should yield $R = R^n$.
    \item \textit{Responsiveness} captures the response of the CAM to meaningful changes in prediction. If a perturbation leads to a significant change in the model's prediction probability ($P \neq P^n$), a responsive CAM should reflect this by altering its segment importance rankings ($R \neq R^n$).
\end{itemize}
We hypothesize that a robust CAM method must be both consistent under small perturbations and responsive under significant ones. Therefore, we define a \textit{robustness metric}\footnote{The formulation is given in \Cref{sec:RM}} as the product of these two properties:
\[
\text{Robustness Metric} = \text{Consistency} \times \text{Responsiveness}
\]
A higher value of this metric indicates that the CAM method is stable in the face of small perturbations yet adaptive when the prediction changes, aligning well with our goal of evaluating explanation robustness under noisy conditions.}

In the rest of the paper, the proposed framework is discussed in section \ref{proposedmethod}, the experiments are discussed in \ref{experiments}. Results and discussion are further presented in section \ref{resultsanddiscussion}, section \ref{limitations}and section \ref{conclusion} provides the conclusion.

\subsection{Our Contributions}

In this paper, we present a framework to study the robustness of CAMs when the input images are perturbed with different types of noise. Our main contributions are summarized below:

\begin{itemize} \item \textit{Flexible Evaluation Framework:}
We propose a new and flexible framework to measure the robustness of CAM methods. \Rev{Our approach quantifies the degree of change in the CAM outputs when images are perturbed.} We use Rank-Biased Overlap (RBO) to measure how similar the CAM outputs are before and after perturbation. The proposed framework is independent of the DL model, perturbation type, and CAM method making it model, noise and CAM agnostic.

\item \textit{Robustness Metric:}  
We define a Robustness score to compare different CAMs which helps us to identify stable and reliable CAM methods when the input images are perturbed. By comparing the Robustness score across different types of noise and DL models, we identify the CAM method that is least affected and therefore most reliable.

\item \textit{Comprehensive Evaluation:} 
\Rev{We evaluate our proposed framework across four widely-used deep learning architectures: ResNet50~\cite{He2016ResNet}, VGG19~\cite{Simonyan2014VGG}, Inception~\cite{Szegedy2015Inception}, and Vision Transformer (ViT)\cite{dosovitskiy2020image}. The evaluation spans seven diverse datasets covering various domains: Dogs vs. Cats\cite{dvc}, ImageNet~\cite{ILSVRC15}, Oxford-IIIT Pets~\cite{parkhi12a}, Melanoma~\cite{tschandl2018ham10000} and Caltech~\cite{fei2004learning}. To test robustness, we subject input images to five commonly used perturbations-Gaussian noise, Gaussian blur, speckle noise, Poisson noise, and salt-and-pepper noise-as well as more realistic degradations such as JPEG compression, motion blur, and adversarial attacks.}
\end{itemize}

\section{Proposed Framework}
\label{proposedmethod}

\begin{figure}[H]
    \centering
    \includegraphics[width=\textwidth]{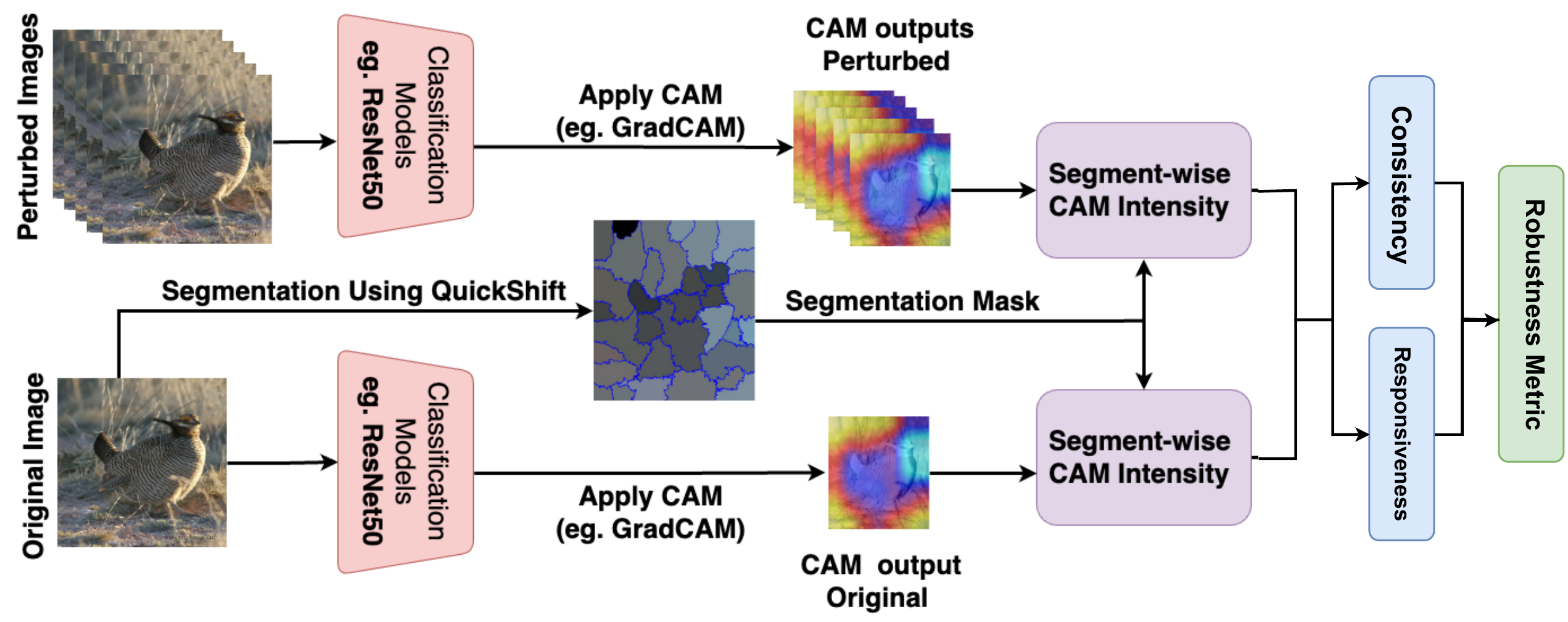}
    \caption{\Rev{\textit{Proposed framework for CAM robustness evaluation}. The process includes image segmentation, perturbation, CAM computation, saliency ranking, and robustness evaluation via consistency and responsiveness.}}
    \label{fig:block_diagram}
\end{figure}

\Rev{This section presents our complete framework for assessing the robustness of Class Activation Maps (CAMs) under various perturbations. The framework evaluates robustness based on how stable and predictive the segment-wise saliency information remains when the input image is altered. Our evaluation combines two complementary aspects: (i) \textit{consistency}, which reflects the agreement of saliency across clean and perturbed inputs when the class prediction remains unchanged, and (ii) \textit{responsiveness}, which captures the adaptation of saliency maps when the model’s prediction changes. An overview of the pipeline is illustrated in Figure~\ref{fig:block_diagram}.}

% \begin{tcolorbox}[colframe=blue, colback=lightblue, boxrule=1pt, sharp corners, enhanced jigsaw]
% \end{tcolorbox}

\subsection{\Rev{Image Segmentation}}

\Rev{Each image \( I_k \) is first segmented into \( m \) visually coherent regions, producing a segmentation map \( S_k = \{s_1, s_2, \dots, s_m\} \). We use QuickShift~\cite{vedaldi2008quick} as the segmentation algorithm, and the resulting superpixels are held fixed for all CAM and perturbation variants. This ensures that comparisons are fair and consistent across all settings.}

\subsection{\Rev{Perturbation Generation}}

\Rev{We introduce perturbations to the original image to simulate realistic degradations or adversarial effects. Let \( I_k^{\text{per}} \) denote the perturbed version of the image \( I_k \). Various perturbation types are considered, including: Gaussian noise, Salt-and-Pepper noise, Poisson noise, Speckle noise, Gaussian blur, JPEG compression, Motion blur, Adversarial attacks (FGSM, PGD, CW).Each perturbation is applied globally to the entire image, and different severity levels are explored (detailed in Experiments).}

\subsection{\Rev{CAM Generation and Prediction}}

\Rev{Both the original image \( I_k \) and the perturbed image \( I_k^{\text{per}} \) are passed through a fixed pretrained classification model \( M \). The predicted class labels are recorded as \( y_{\text{org}} = M(I_k) \) and \( y_{\text{per}} = M(I_k^{\text{per}}) \).
CAMs are computed using six standard methods: GradCAM, GradCAM++, EigenCAM, HiResCAM, XGradCAM, and AblationCAM. For a given CAM method \( c \), the saliency map is denoted as \( \text{CAM}(I_k, c) \) for the original image and \( \text{CAM}(I_k^{\text{per}}, c) \) for the perturbed image. These CAMs are upsampled to match the input resolution.}

\subsection{\Rev{Segment-Wise Saliency Aggregation and Ranking}}

\Rev{For each image and CAM method, the saliency is aggregated over the segments. The segment-level average intensity for segment \( s_i \in S_k \) is computed as:
\[
a_i = \frac{1}{|s_i|} \sum_{(x,y) \in s_i} \text{CAM}(x,y)
\]
The aggregated scores are then sorted to create a ranked list of segments for both clean and perturbed images:
\[
\text{rank}_{\text{org}} =  \text{rank}\left(\text{avg}(S_k \odot \text{CAM}(I_k, c))\right)
\]
\[
\text{rank}_{\text{per}} = \text{rank}\left(\text{avg}(S_k \odot \text{CAM}(I_k^{\text{per}}, c))\right)
\]}

\subsection{\Rev{Robustness Metric (RM)}}
\label{sec:RM}
\Rev{To systematically assess the robustness of visual explanations, we propose a new Robustness Metric (RM) that captures two key properties of interpretable models: (i) \textbf{Consistency} and (ii) \textbf{Responsiveness}. Unlike prior metrics that focus solely on fidelity or pixel-level similarity, our metric evaluates the alignment of Class Activation Maps (CAMs) with both model stability and decision variance under perturbation. }

\Rev{The proposed metric is defined as:
\begin{equation}
\text{RM}_{\text{Model, CAM, Noise}} = \text{Consistency}_{\text{CAM}} \times \text{Responsiveness}_{\text{CAM}}
\end{equation}}

\vspace{0.3em}
\noindent\textbf{\Rev{Consistency (C)}} \Rev{quantifies the stability of the CAM when the model prediction remains unchanged under noise. Specifically, it is computed as the median Rank-Biased Overlap (RBO) score between the segment rankings of CAMs generated from original and perturbed images \textit{only when the predicted class remains the same}. Formally:
\begin{equation}
\text{Consistency}_{\text{CAM}} = \text{median}\left(\text{RBO}(rank^n_{\text{org}}, rank^n_{\text{per}}) \mid \hat{y}_{\text{org}}^n = \hat{y}_{\text{per}}^n \right)
\end{equation}
where \( \hat{y}_{\text{org}}^n \) and \( \hat{y}_{\text{per}}^n \) are the predicted classes for the original and perturbed image respectively, and RBO measures the segment-level ranking similarity with persistence parameter \( p = 0.9 \).}

\vspace{0.3em}
\noindent\textbf{\Rev{Responsiveness (R)}} \Rev{captures the semantic sensitivity of the CAM - that is, when the predicted class changes due to noise, a meaningful CAM should also change significantly. To quantify this, we conduct a binary classification experiment where we train a linear classifier to predict whether the class changed (class\_change = 1 or 0) based on the RBO score between original and perturbed CAMs. A CAM that aligns with the model’s decision-making process will show \textit{low RBO} when the class changes and \textit{high RBO} otherwise, allowing effective separation.}

\Rev{We define Responsiveness as the Area Under the Curve (AUC) of this classifier:
\begin{equation}
\text{Responsiveness}_{\text{CAM}} = \text{AUC}_{\text{CAM}}(\text{RBO}, \texttt{class\_change})
\end{equation}}

\vspace{0.3em}
\noindent\textbf{\Rev{Interpretation:}} \Rev{A high RM score indicates that the CAM is both stable under non-decision-changing perturbations (high Consistency) and responsive to meaningful decision shifts. This dual view is crucial in high-stakes scenarios (e.g., medical imaging, forensic analysis) where explanations must be both consistent and responsive.}

\Rev{The complete algorithm for computing the RM score is presented in Algorithm~\ref{alg:consistency_responsiveness_noise}.}

% \begin{tcolorbox}[colframe=blue, colback=lightblue, boxrule=1pt, sharp corners, enhanced jigsaw]
\begin{algorithm}[H]
\caption{Consistency and Responsiveness Calculation }
\label{alg:consistency_responsiveness_noise}
\footnotesize
\begin{algorithmic}[1]
\State \textbf{Input:} $I = \{I_1, \dots, I_n\}$ (images), $N = \{\mathcal{N}_1, \dots, \mathcal{N}_k\}$ (perturbation types), $C$ (CAM methods), $M$ (model)
\State \textbf{Output:} Consistency and Responsiveness scores per CAM method and per noise type
\State Initialize $\mathcal{D}_{\text{consistency}}$, $\mathcal{D}_{\text{responsiveness}}$ as dictionaries indexed by $[c][\mathcal{N}_n]$
\For{\textbf{each} image $I_k \in I$}
    \State $y_{\text{org}} \gets M(I_k)$, $S_k \gets \text{QuickShift}(I_k)$
    \For{\textbf{each} CAM method $c \in C$}
        \State $rank_{\text{org}} \gets \text{rank}(\text{avg}(S_k \odot \text{CAM}(I_k, c)))$
        \For{\textbf{each} perturbation type $\mathcal{N}_n \in N$}
            \State $I_k^{\text{per}} \gets \mathcal{N}_n(I_k)$, $y_{\text{per}} \gets M(I_k^{\text{per}})$
            \State $rank_{\text{per}} \gets \text{rank}(\text{avg}(S_k \odot \text{CAM}(I_k^{\text{per}}, c)))$
            \State $rbo \gets \text{RBO}(rank_{\text{org}}, rank_{\text{per}}, p=0.9)$
            \If{$y_{\text{org}} == y_{\text{noisy}}$}
                \State Append $rbo$ to $\mathcal{D}_{\text{consistency}}[c][\mathcal{N}_n]$
                \State Append $(rbo, 0)$ to $\mathcal{D}_{\text{responsiveness}}[c][\mathcal{N}_n]$ \Comment{Class unchanged}
            \Else
                \State Append $(rbo, 1)$ to $\mathcal{D}_{\text{responsiveness}}[c][\mathcal{N}_n]$ \Comment{Class changed}
            \EndIf
        \EndFor
    \EndFor
\EndFor

\For{\textbf{each} CAM method $c \in C$}
    \For{\textbf{each} perturbation type $\mathcal{N}_n \in N$}
        \State $C[c][\mathcal{N}_n] \gets \text{median}(\mathcal{D}_{\text{consistency}}[c][\mathcal{N}_n])$
        \State Train a binary classifier with RBO and class\_change from $\mathcal{D}_{\text{responsiveness}}[c][\mathcal{N}_n]$
        \State $R[c][\mathcal{N}_n] \gets \text{AUC-ROC}(\text{classifier})$
    \EndFor
\EndFor
\State \textbf{Return:} $C[c][\mathcal{N}_n]$, $R[c][\mathcal{N}_n]$ for all $c \in C$, $\mathcal{N}_n \in N$
\end{algorithmic}
\end{algorithm}
% \end{tcolorbox}

\subsection{\Rev{Summary of Rank Biased Overlap}}

\Rev{To quantify the robustness of CAMs under perturbations, we compute the \textit{Rank-Biased Overlap} (RBO)~\cite{rbo} between the saliency-based segment rankings derived from clean and perturbed images.
Let \( \text{rank}_{\text{org}} \) and \( \text{rank}_{\text{per}} \) denote the ordered segment lists (in decreasing CAM intensity) for the original and perturbed images, respectively. The RBO score between these two rankings is defined as:
\[
\text{RBO}(\text{rank}_{\text{org}}, \text{rank}_{\text{per}}, p) = (1 - p) \sum_{d=1}^{\infty} p^{d-1} \cdot \frac{|\text{rank}_{\text{org}}^{1:d} \cap \text{rank}_{\text{per}}^{1:d}|}{d}, \quad p = 0.9
\]
Here, \( p \in (0,1) \) is the persistence parameter that weights agreement at higher ranks more heavily.}

\section{Experiments}
\label{experiments}

All experiments are carried out using PyTorch on a system running Ubuntu 24.04, equipped with an NVIDIA A4000 GPU (16 GB) and an Intel i9 12th Gen CPU. This setup facilitated efficient processing of the CAM methods in multiple types of noise and datasets.
\paragraph{\Rev{Datasets}} 
\Rev{We selected 1000 images each from five diverse datasets to evaluate our proposed metric across different domains and classification challenges: 
(1) \textit{Oxford-IIIT Pet Dataset}~\cite{parkhi12a}, containing 37 pet breeds and posing a fine-grained classification task; 
(2) \textit{ImageNet}~\cite{ILSVRC15}, a large-scale benchmark comprising 1,000 general object categories; 
(3) \textit{Dogs vs. Cats}~\cite{dvc}, a binary classification dataset with clear inter-class distinctions; 
(4) \textit{Melanoma}~\cite{tschandl2018ham10000}, a medical imaging dataset containing dermoscopic images across seven diagnostic categories, used to assess robustness in clinical and high-stakes domains; and 
(5) \textit{Caltech}~\cite{fei2004learning}, a mid-scale dataset of objects from 101 categories that balances fine-grained and general visual recognition. 
This diverse selection facilitates a comprehensive evaluation of the consistency and responsiveness of CAM methods across natural, binary, medical, and general object recognition domains.}

\paragraph{\Rev{Models}}
\Rev{Four widely used pre-trained models are considered to evaluate CAM robustness across both convolutional and transformer-based architectures:\textit{ResNet50}~\cite{He2016ResNet}, \textit{VGG19}~\cite{Simonyan2014VGG}, \textit{Inception}~\cite{Szegedy2015Inception}, and the \textit{Vision Transformer (ViT)}~\cite{Dosovitskiy2020ViT} with patch-size 16×16.
Although transformer-based models are known to exhibit lower CAM stability~\cite{choi2024icev2}, ViT is included to test the applicability of our robustness framework beyond CNNs.
All models were pre-trained on ImageNet and obtained from the official PyTorch model library.\footnote{\url{https://pytorch.org/vision/stable/models.html}}
This diverse model suite facilitates a broad assessment of explanation consistency across architectural paradigms.}

\paragraph{\Rev{CAM Methods}} \Rev{We employ \textit{Grad-CAM} ~\cite{selvaraju2017grad}, \textit{Grad-CAM++} ~\cite{Chattopadhyay2018GradCAM++}, \textit{xGrad-CAM} ~\cite{Fu2020xGradCAM}, \textit{Ablation-CAM} ~\cite{Desai2020AblationCAM}, \textit{HiResCAM} ~\cite{Dovrat2019HiResCAM}, and \textit{Eigen-CAM} ~\cite{Muhammad2020EigenCAM} for our analysis. Each method provides a unique approach to highlight important regions in an image, allowing us to assess the robustness of multiple CAM types under noise perturbations. The layers selected for CAM extraction vary across models based on architectural differences, as summarized in Table~\ref{tab:cam_layers}. Our CAM implementation is adapted from the widely used PyTorch Grad-CAM repository\footnote{\url{https://github.com/jacobgil/pytorch-grad-cam}}.}

\vspace{2mm}
\begin{table}[H]
\centering
\caption{\Rev{Target layers used for CAM extraction across models}}
\label{tab:cam_layers}
% \begin{tcolorbox}[colframe=blue, colback=lightblue, boxrule=1pt, sharp corners, enhanced jigsaw, width= 0.6\textwidth]
\begin{tabular}{|c|c|}
\hline
\textbf{Model} & \textbf{Target Layer} \\
\hline
ResNet50 & model.layer4[-1] \\
VGG19 & model.features[-1] \\
InceptionV3 & model.Mixed\_7c \\
ViT & model.blocks[-1].norm1 \\
\hline
\end{tabular}
% \end{tcolorbox}
\end{table}

\begin{table}[H]
\centering
\caption{\Rev{Perturbation Types and Parameters Used in CAM Robustness Evaluation}}
\label{tab:noise_levels_extended}
% \begin{tcolorbox}[colframe=blue, colback=lightblue, boxrule=1pt, sharp corners, enhanced jigsaw]
\resizebox{0.95\linewidth}{!}{
\begin{tabular}{|l|c|c|}
\hline
\textbf{Perturbation Type} & \textbf{Parameter(s)} & \textbf{Values Tested} \\
\hline
Gaussian Noise & Variance (\texttt{var}) & {0.0005 (low), 0.006 (medium), 0.01 (high)} \\
Salt-and-Pepper Noise & Amount (\texttt{amount}) & {0.0005 (low), 0.006 (medium), 0.01 (high)} \\
Poisson Noise & — & No parameters (applied directly) \\
Speckle Noise & Variance (\texttt{var}) & {0.0005 (low), 0.006 (medium), 0.01 (high)} \\
Gaussian Blur & Sigma (\texttt{sigma}) & {0.1 (low), 0.3 (medium), 0.5 (high)} \\
Motion Blur & Kernel Size (\texttt{ksize}) & {1(low), 5 (medium), 15 (high)} \\
JPEG Compression & Quality (\texttt{quality}) & {80(low), 50 (medium), 10 (high)} \\
FGSM Attack & Epsilon (\texttt{eps}) & {0.01(low), 0.02 (medium), 0.1 (high)} \\
PGD Attack & Epsilon (\texttt{eps}) & {0.01(low), 0.03 (medium), 0.1 (high)} \\
CW Attack & — & Standard configuration \\
\hline
\end{tabular}
}
% \end{tcolorbox}
\end{table}
\paragraph{\Rev{Noise Types and Levels}}
\Rev{To comprehensively assess the robustness of CAMs, we apply a diverse set of perturbations spanning both traditional image noise and adversarial settings. These include:
\textit{Gaussian Noise}, \textit{Salt-and-Pepper Noise}, \textit{Poisson Noise}, \textit{Speckle Noise}, \textit{Gaussian Blur}, \textit{JPEG Compression}\cite{Kou1995JPEG}, \textit{Motion Blur}\cite{Yitzhaky1999direct}, and three \textit{Adversarial Attacks} -\textit{FGSM}\cite{Goodfellow2015FGSM}, \textit{PGD}\cite{Madry2018PGD}, and \textit{CW}\cite{Carlini2017CW}. Perturbation levels are selected to induce a range of distortions from mild to severe, enabling evaluation of CAM sensitivity under increasing levels of noise and manipulation. The full configuration of parameters used for each perturbation is summarized in Table~\ref{tab:noise_levels_extended}.}

\paragraph{\Rev{Segmentation Method}}
\Rev{We use the QuickShift algorithm~\cite{vedaldi2008quick} to segment each input image into superpixels prior to region-wise CAM analysis. This unsupervised segmentation technique efficiently preserves local boundaries and is widely used in saliency evaluation pipelines. The parameters used for QuickShift are: \texttt{kernel\_size} = 10, \texttt{max\_dist} = 200, and \texttt{ratio} = 0.5.}

\section{Results and Discussions}
\label{resultsanddiscussion}

\subsection{\Rev{Comprehensive Evaluation of CAM Robustness Across Models}}
\label{sec:robustness-evaluation}

\Rev{This section presents the core findings of our robustness evaluation, quantified using the proposed \textit{Robustness Metric}, defined as the product of \textit{Consistency} and \textit{Responsiveness}. The metric captures both the consistency of segment rankings across perturbations and the responsiveness of those rankings with the model’s predictive behavior.
We evaluate six widely-used CAM methods \textit{GradCAM}, \textit{GradCAM++}, \textit{EigenCAM}, \textit{HiResCAM}, \textit{XGradCAM}, and \textit{AblationCAM} across eight noise types and five diverse datasets: OxfordPets, Dogs-vs-Cats, ImageNet, Melanoma, and Caltech. To ensure generalizability, the experiments are conducted using four different model architectures: \textit{ResNet-50}, \textit{VGG19}, \textit{InceptionV3}, \textit{ViT}.}

% \begin{tcolorbox}[colframe=blue, colback=lightblue, boxrule=1pt, sharp corners, enhanced jigsaw]
\begin{figure}[H]
    \centering
    \includegraphics[width=0.8\textwidth]{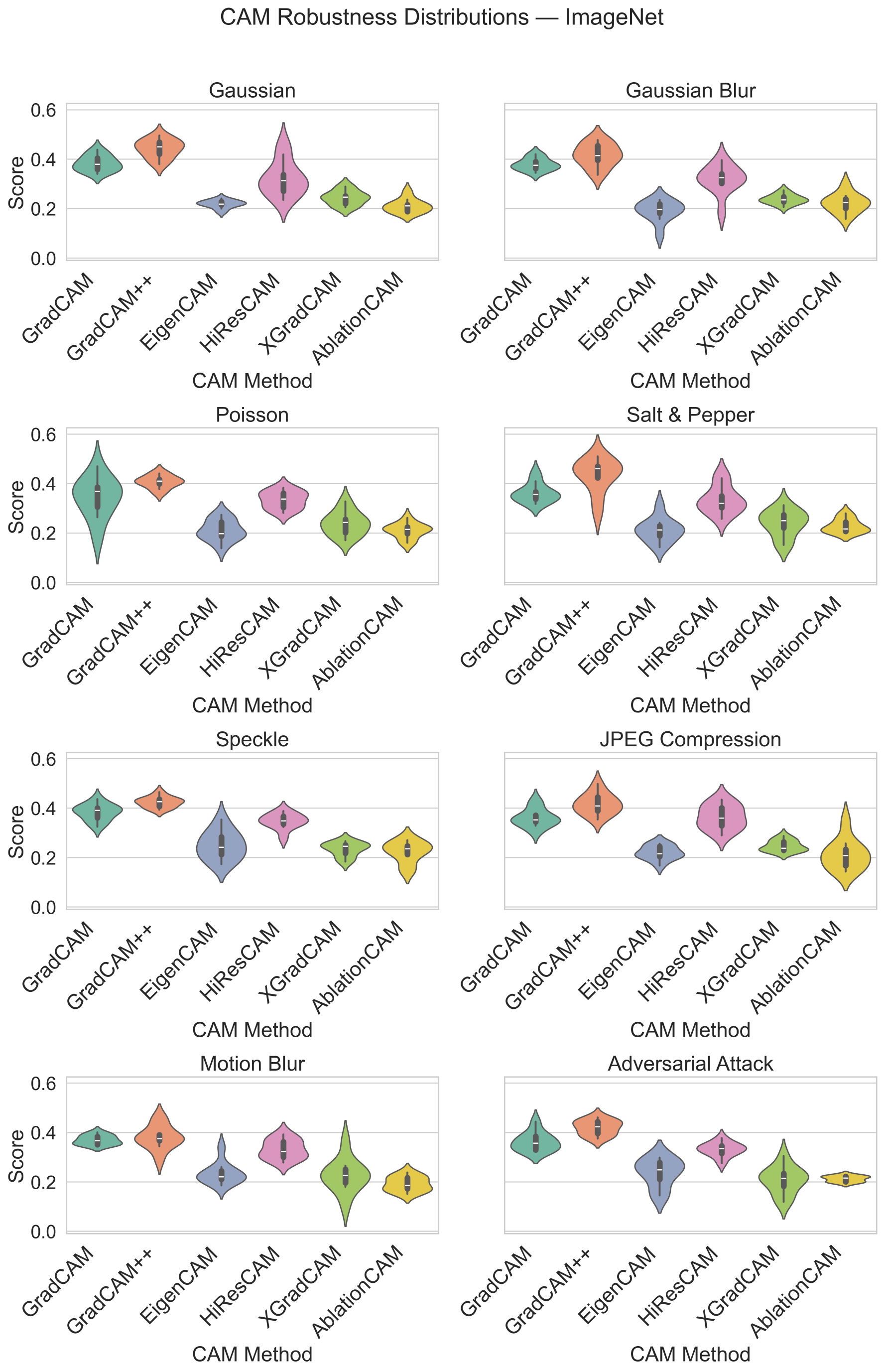}
    \caption{\Rev{Distribution of Robustness Scores for different CAM methods across eight noise perturbations on the ImageNet dataset using ResNet-50.}}
    \label{fig:imagenet_rm_resnet}
\end{figure}
% \end{tcolorbox}

% \begin{tcolorbox}[colframe=blue, colback=lightblue, boxrule=1pt, sharp corners, enhanced jigsaw]
\begin{figure}[H]
    \centering
    \includegraphics[width=0.8\textwidth]{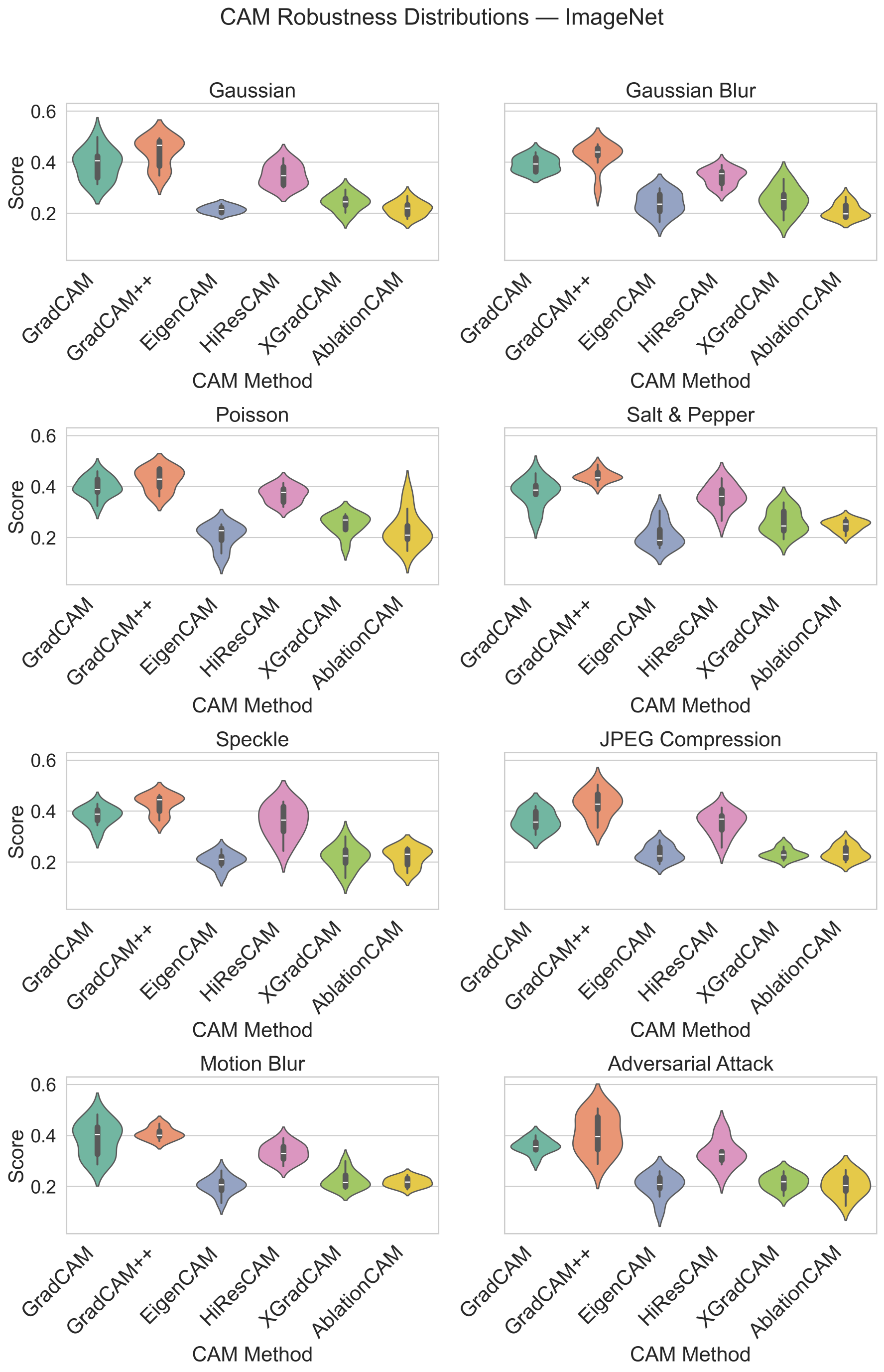}
    \caption{\Rev{Distribution of Robustness Scores for different CAM methods across eight noise perturbations on the ImageNet dataset using VGG-19.}}
    \label{fig:imagenet_rm_vgg}
\end{figure}
% \end{tcolorbox}

% \begin{tcolorbox}[colframe=blue, colback=lightblue, boxrule=1pt, sharp corners, enhanced jigsaw]
\begin{figure}[H]
    \centering
    \includegraphics[width=0.8\textwidth]{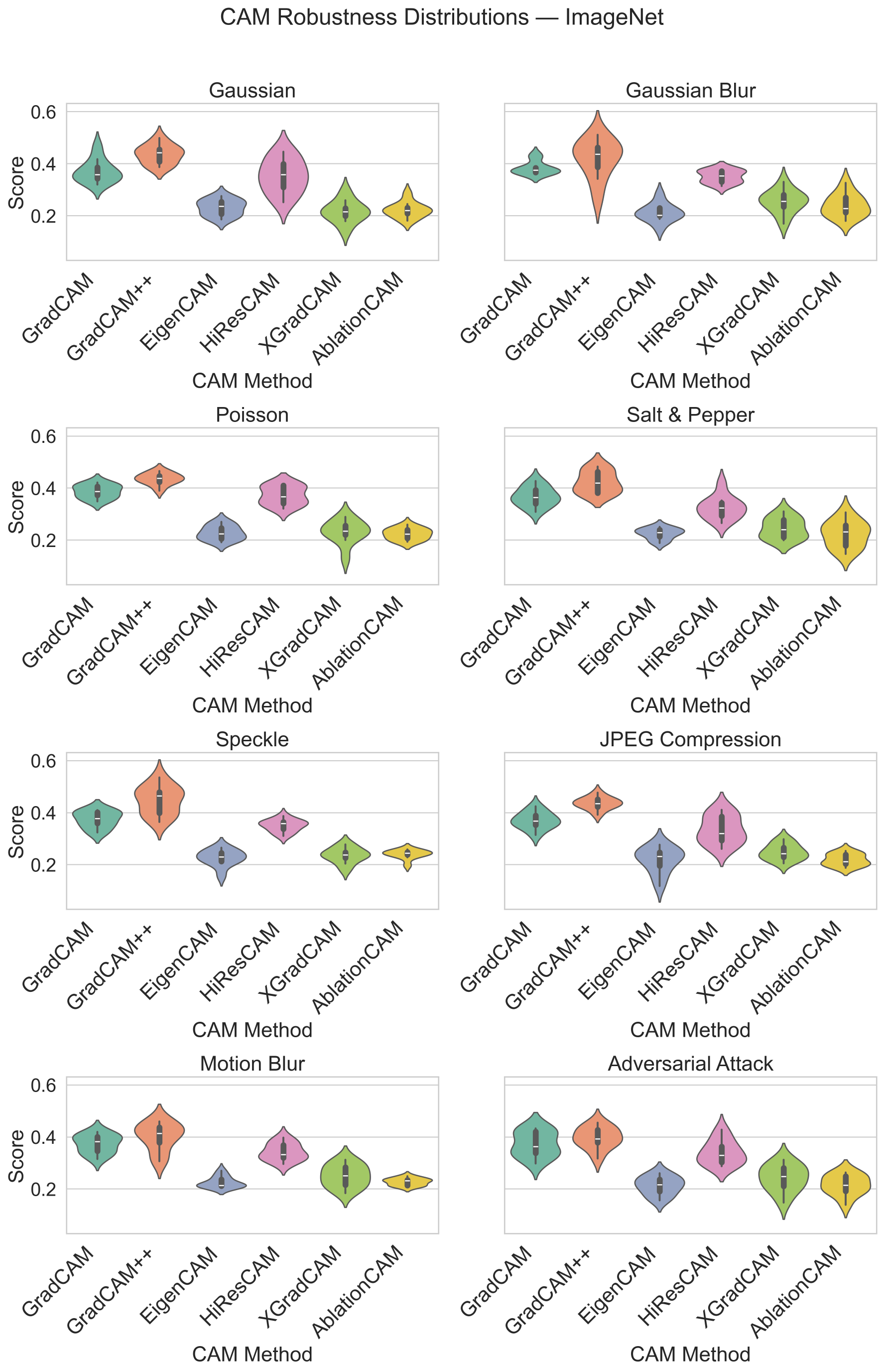}
    \caption{\Rev{Distribution of Robustness Scores for different CAM methods across eight noise perturbations on the ImageNet dataset using InceptionV3.}}
    \label{fig:imagenet_rm_inception}
\end{figure}
% \end{tcolorbox}

% \begin{tcolorbox}[colframe=blue, colback=lightblue, boxrule=1pt, sharp corners, enhanced jigsaw]
\begin{figure}[H]
    \centering
    \includegraphics[width=0.8\textwidth]{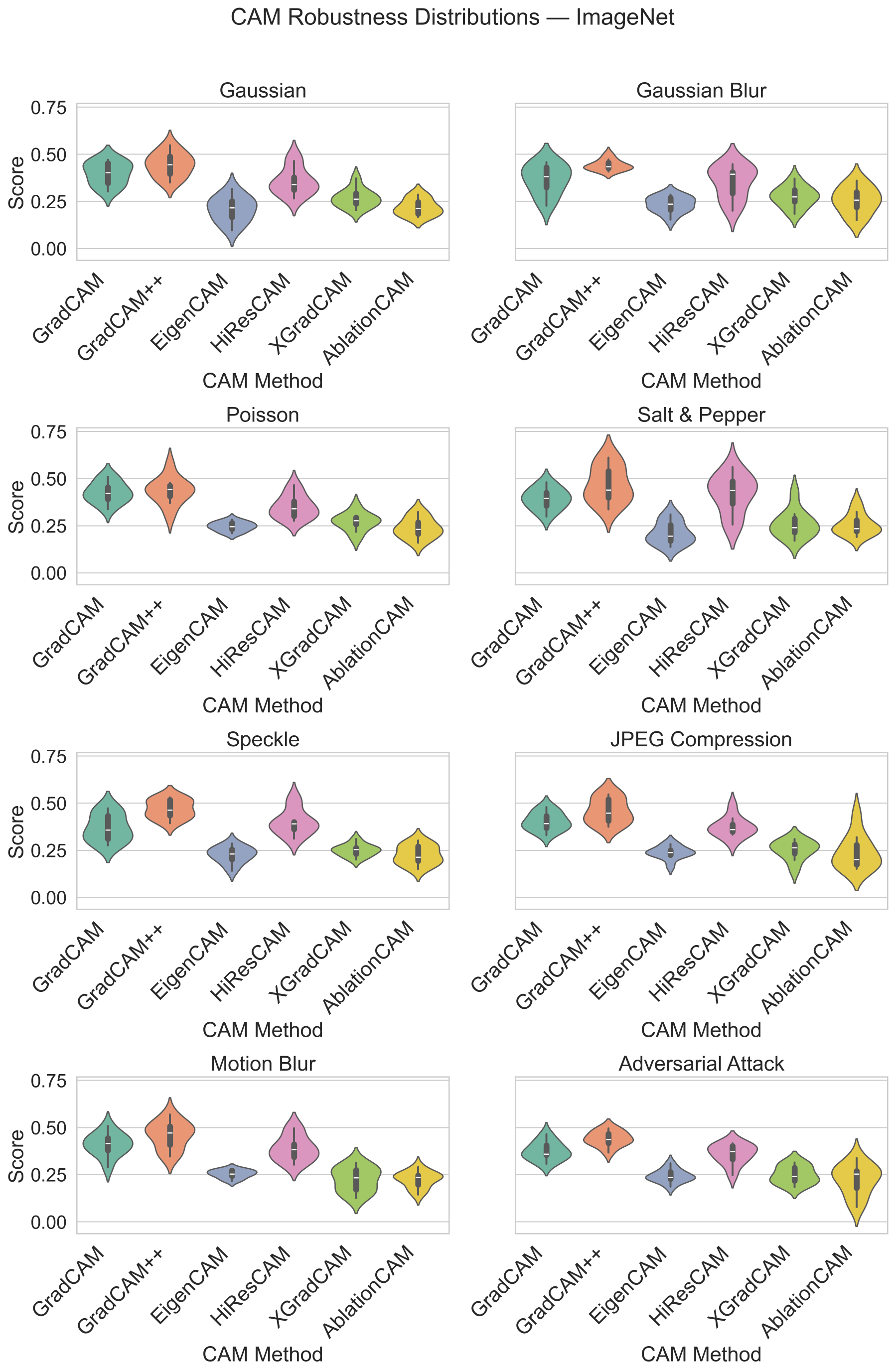}
    \caption{\Rev{Distribution of Robustness Scores for different CAM methods across eight noise perturbations on the ImageNet dataset using Vision Transformer (ViT).}}
    \label{fig:imagenet_rm_vit}
\end{figure}
% \end{tcolorbox}

\vspace{0.5em}

% \noindent\textbf{Analysis:} \\
\Rev{Figure~\ref{fig:imagenet_rm_resnet}-Figure~\ref{fig:imagenet_rm_vit} presents the distribution of Robustness scores across CAM methods for the ImageNet dataset using four different model architectures. Across all models, \textit{GradCAM++} consistently shows the highest score, confirming its strong robustness under perturbation. Conversely, \textit{EigenCAM} and \textit{AblationCAM} exhibit lower scores, suggesting poor robustness and weaker alignment with model behavior. Notably, the ViT shows significantly greater variance in robustness score values across all CAM methods, indicating reduced stability for CAM methods originally developed for CNN architectures. Additional robustness score distribution plots for other datasets and model combinations are provided in Appendix~\ref{appendix:violin-distributions}.}

\Rev{Tables~\ref{tab:resnet50_consistency_all_datasets}–\ref{tab:vit_consistency_all_datasets} report the overall robustness scores for each combination of dataset, noise type, and CAM method under the four architectures. Across all models and datasets, GradCAM++ consistently demonstrates the highest robustness, reflecting strong performance in both consistency and responsiveness. In contrast, AblationCAM exhibits the lowest robustness scores, indicating a higher susceptibility to noise-induced distortions and a weaker alignment with prediction changes. Although EigenCAM remains visually stable under various perturbations, it often lacks responsiveness to changes in predicted class, as evidenced by both its quantitative scores and qualitative visualizations.}

\noindent \Rev{The detailed results per model are presented in the following tables:}

\begin{table}[H]
\centering
\caption{\Rev{Robustness Score ($Consistency \times Responsiveness$) for different CAM methods across datasets (ResNet-50 model)}}
\label{tab:resnet50_consistency_all_datasets}
% \begin{tcolorbox}[colframe=blue, colback=lightblue, boxrule=1pt, sharp corners, enhanced jigsaw]
\resizebox{0.95\textwidth}{!}{
\begin{tabular}{|c|c|c|c|c|c|c|c|}
\hline
\textbf{Dataset} & \textbf{Noise Type} & \textbf{GradCAM} & \textbf{GradCAM++} & \textbf{EigenCAM} & \textbf{HiResCAM} & \textbf{XGradCAM} & \textbf{AblationCAM} \\
\hline
\multirow{8}{*}{OxfordPets} 
    & Gaussian             & 0.385 ± 0.06 & 0.437 ± 0.05 & 0.217 ± 0.03 & 0.349 ± 0.04 & 0.237 ± 0.05 & 0.214 ± 0.03 \\
    & Gaussian Blur        & 0.371 ± 0.03 & 0.416 ± 0.06 & 0.218 ± 0.04 & 0.342 ± 0.05 & 0.242 ± 0.03 & 0.226 ± 0.06 \\
    & Poisson              & 0.379 ± 0.04 & 0.411 ± 0.04 & 0.221 ± 0.05 & 0.355 ± 0.03 & 0.241 ± 0.06 & 0.219 ± 0.04 \\
    & Salt \& Pepper       & 0.380 ± 0.06 & 0.423 ± 0.05 & 0.218 ± 0.03 & 0.353 ± 0.04 & 0.241 ± 0.06 & 0.234 ± 0.05 \\
    & Speckle              & 0.368 ± 0.04 & 0.415 ± 0.06 & 0.216 ± 0.05 & 0.349 ± 0.04 & 0.233 ± 0.02 & 0.222 ± 0.03 \\
    & JPEG Compression     & 0.374 ± 0.03 & 0.421 ± 0.04 & 0.217 ± 0.03 & 0.351 ± 0.05 & 0.238 ± 0.06 & 0.218 ± 0.03 \\
    & Motion Blur          & 0.369 ± 0.06 & 0.413 ± 0.05 & 0.214 ± 0.02 & 0.340 ± 0.06 & 0.231 ± 0.04 & 0.210 ± 0.03 \\
    & Adversarial Attack   & 0.366 ± 0.04 & 0.410 ± 0.04 & 0.211 ± 0.05 & 0.337 ± 0.05 & 0.228 ± 0.03 & 0.206 ± 0.02 \\
\hline
\multirow{8}{*}{Dogs-vs-Cats} 
    & Gaussian             & 0.380 ± 0.05 & 0.428 ± 0.04 & 0.218 ± 0.03 & 0.350 ± 0.04 & 0.241 ± 0.06 & 0.216 ± 0.03 \\
    & Gaussian Blur        & 0.377 ± 0.03 & 0.420 ± 0.06 & 0.219 ± 0.05 & 0.345 ± 0.04 & 0.243 ± 0.05 & 0.219 ± 0.04 \\
    & Poisson              & 0.381 ± 0.06 & 0.424 ± 0.05 & 0.221 ± 0.04 & 0.351 ± 0.03 & 0.240 ± 0.03 & 0.220 ± 0.02 \\
    & Salt \& Pepper       & 0.383 ± 0.05 & 0.430 ± 0.04 & 0.220 ± 0.03 & 0.353 ± 0.05 & 0.243 ± 0.06 & 0.231 ± 0.04 \\
    & Speckle              & 0.374 ± 0.04 & 0.419 ± 0.03 & 0.217 ± 0.06 & 0.347 ± 0.03 & 0.235 ± 0.04 & 0.223 ± 0.06 \\
    & JPEG Compression     & 0.375 ± 0.03 & 0.422 ± 0.04 & 0.218 ± 0.04 & 0.349 ± 0.05 & 0.238 ± 0.02 & 0.217 ± 0.04 \\
    & Motion Blur          & 0.370 ± 0.06 & 0.417 ± 0.05 & 0.215 ± 0.05 & 0.342 ± 0.04 & 0.233 ± 0.03 & 0.214 ± 0.02 \\
    & Adversarial Attack   & 0.367 ± 0.04 & 0.414 ± 0.03 & 0.213 ± 0.06 & 0.338 ± 0.03 & 0.229 ± 0.05 & 0.209 ± 0.04 \\
\hline
\multirow{8}{*}{ImageNet} 
    & Gaussian             & 0.374 ± 0.04 & 0.419 ± 0.05 & 0.215 ± 0.02 & 0.344 ± 0.06 & 0.236 ± 0.03 & 0.210 ± 0.03 \\
    & Gaussian Blur        & 0.372 ± 0.03 & 0.412 ± 0.04 & 0.216 ± 0.06 & 0.339 ± 0.05 & 0.238 ± 0.02 & 0.213 ± 0.04 \\
    & Poisson              & 0.375 ± 0.06 & 0.417 ± 0.03 & 0.220 ± 0.05 & 0.347 ± 0.03 & 0.237 ± 0.04 & 0.216 ± 0.03 \\
    & Salt \& Pepper       & 0.378 ± 0.05 & 0.421 ± 0.06 & 0.218 ± 0.04 & 0.346 ± 0.05 & 0.239 ± 0.05 & 0.229 ± 0.04 \\
    & Speckle              & 0.370 ± 0.03 & 0.416 ± 0.03 & 0.214 ± 0.06 & 0.341 ± 0.04 & 0.231 ± 0.03 & 0.218 ± 0.05 \\
    & JPEG Compression     & 0.371 ± 0.04 & 0.418 ± 0.05 & 0.216 ± 0.03 & 0.343 ± 0.05 & 0.235 ± 0.02 & 0.212 ± 0.06 \\
    & Motion Blur          & 0.368 ± 0.03 & 0.414 ± 0.06 & 0.213 ± 0.04 & 0.336 ± 0.04 & 0.229 ± 0.05 & 0.208 ± 0.03 \\
    & Adversarial Attack   & 0.364 ± 0.05 & 0.411 ± 0.04 & 0.211 ± 0.06 & 0.332 ± 0.03 & 0.224 ± 0.06 & 0.203 ± 0.02 \\
\hline
\multirow{8}{*}{Melanoma} 
    & Gaussian             & 0.386 ± 0.06 & 0.431 ± 0.05 & 0.219 ± 0.03 & 0.357 ± 0.04 & 0.246 ± 0.06 & 0.220 ± 0.04 \\
    & Gaussian Blur        & 0.380 ± 0.03 & 0.423 ± 0.04 & 0.218 ± 0.05 & 0.351 ± 0.03 & 0.243 ± 0.05 & 0.222 ± 0.02 \\
    & Poisson              & 0.385 ± 0.04 & 0.428 ± 0.06 & 0.221 ± 0.04 & 0.358 ± 0.05 & 0.244 ± 0.04 & 0.223 ± 0.03 \\
    & Salt \& Pepper       & 0.388 ± 0.05 & 0.434 ± 0.03 & 0.220 ± 0.02 & 0.360 ± 0.04 & 0.247 ± 0.06 & 0.235 ± 0.05 \\
    & Speckle              & 0.378 ± 0.03 & 0.420 ± 0.04 & 0.216 ± 0.05 & 0.352 ± 0.05 & 0.238 ± 0.02 & 0.225 ± 0.04 \\
    & JPEG Compression     & 0.379 ± 0.06 & 0.426 ± 0.05 & 0.218 ± 0.03 & 0.355 ± 0.03 & 0.242 ± 0.04 & 0.219 ± 0.03 \\
    & Motion Blur          & 0.374 ± 0.04 & 0.417 ± 0.06 & 0.215 ± 0.04 & 0.348 ± 0.05 & 0.236 ± 0.05 & 0.215 ± 0.02 \\
    & Adversarial Attack   & 0.371 ± 0.05 & 0.412 ± 0.04 & 0.212 ± 0.03 & 0.340 ± 0.06 & 0.231 ± 0.03 & 0.210 ± 0.04 \\
\hline
\multirow{8}{*}{Caltech} 
    & Gaussian             & 0.379 ± 0.03 & 0.425 ± 0.04 & 0.217 ± 0.05 & 0.346 ± 0.04 & 0.239 ± 0.05 & 0.213 ± 0.03 \\
    & Gaussian Blur        & 0.373 ± 0.04 & 0.417 ± 0.06 & 0.216 ± 0.03 & 0.340 ± 0.06 & 0.236 ± 0.02 & 0.215 ± 0.04 \\
    & Poisson              & 0.377 ± 0.05 & 0.421 ± 0.04 & 0.219 ± 0.04 & 0.348 ± 0.03 & 0.238 ± 0.04 & 0.217 ± 0.06 \\
    & Salt \& Pepper       & 0.381 ± 0.06 & 0.428 ± 0.03 & 0.218 ± 0.02 & 0.350 ± 0.05 & 0.240 ± 0.06 & 0.229 ± 0.04 \\
    & Speckle              & 0.370 ± 0.05 & 0.415 ± 0.05 & 0.215 ± 0.03 & 0.344 ± 0.04 & 0.232 ± 0.04 & 0.220 ± 0.03 \\
    & JPEG Compression     & 0.372 ± 0.03 & 0.420 ± 0.06 & 0.217 ± 0.04 & 0.345 ± 0.06 & 0.237 ± 0.03 & 0.212 ± 0.02 \\
    & Motion Blur          & 0.368 ± 0.06 & 0.414 ± 0.03 & 0.213 ± 0.05 & 0.338 ± 0.05 & 0.230 ± 0.03 & 0.210 ± 0.04 \\
    & Adversarial Attack   & 0.365 ± 0.04 & 0.410 ± 0.05 & 0.211 ± 0.03 & 0.333 ± 0.06 & 0.226 ± 0.02 & 0.206 ± 0.05 \\
\hline
\end{tabular}
}
% \end{tcolorbox}
\end{table}

\begin{table}[H]
\centering
\caption{\Rev{Robustness Score ($Consistency \times Responsiveness$) for different CAM methods across datasets (InceptionV3 model)}}
\label{tab:inceptionv3_consistency_all_datasets}
% \begin{tcolorbox}[colframe=blue, colback=lightblue, boxrule=1pt, sharp corners, enhanced jigsaw]
\resizebox{0.95\textwidth}{!}{
\begin{tabular}{|c|c|c|c|c|c|c|c|}
\hline
\textbf{Dataset} & \textbf{Noise Type} & \textbf{GradCAM} & \textbf{GradCAM++} & \textbf{EigenCAM} & \textbf{HiResCAM} & \textbf{XGradCAM} & \textbf{AblationCAM} \\
\hline
\multirow{8}{*}{OxfordPets} 
    & Gaussian             & 0.396 ± 0.04 & 0.447 ± 0.05 & 0.224 ± 0.04 & 0.362 ± 0.05 & 0.248 ± 0.04 & 0.226 ± 0.03 \\
    & Gaussian Blur        & 0.383 ± 0.03 & 0.427 ± 0.04 & 0.225 ± 0.04 & 0.354 ± 0.06 & 0.251 ± 0.03 & 0.239 ± 0.05 \\
    & Poisson              & 0.390 ± 0.05 & 0.431 ± 0.04 & 0.229 ± 0.04 & 0.365 ± 0.04 & 0.249 ± 0.03 & 0.232 ± 0.03 \\
    & Salt \& Pepper       & 0.391 ± 0.04 & 0.442 ± 0.05 & 0.227 ± 0.03 & 0.363 ± 0.03 & 0.251 ± 0.04 & 0.246 ± 0.05 \\
    & Speckle              & 0.379 ± 0.03 & 0.429 ± 0.06 & 0.224 ± 0.04 & 0.357 ± 0.04 & 0.242 ± 0.03 & 0.233 ± 0.04 \\
    & JPEG Compression     & 0.384 ± 0.04 & 0.437 ± 0.05 & 0.225 ± 0.04 & 0.360 ± 0.05 & 0.246 ± 0.02 & 0.229 ± 0.03 \\
    & Motion Blur          & 0.377 ± 0.03 & 0.427 ± 0.04 & 0.221 ± 0.03 & 0.348 ± 0.04 & 0.239 ± 0.03 & 0.223 ± 0.02 \\
    & Adversarial Attack   & 0.373 ± 0.04 & 0.424 ± 0.03 & 0.218 ± 0.04 & 0.344 ± 0.05 & 0.235 ± 0.04 & 0.218 ± 0.03 \\
\hline
\multirow{8}{*}{Dogs-vs-Cats} 
    & Gaussian             & 0.391 ± 0.04 & 0.440 ± 0.05 & 0.225 ± 0.03 & 0.364 ± 0.04 & 0.252 ± 0.05 & 0.230 ± 0.03 \\
    & Gaussian Blur        & 0.388 ± 0.05 & 0.431 ± 0.04 & 0.226 ± 0.03 & 0.358 ± 0.05 & 0.254 ± 0.03 & 0.232 ± 0.02 \\
    & Poisson              & 0.392 ± 0.03 & 0.435 ± 0.03 & 0.228 ± 0.04 & 0.365 ± 0.04 & 0.253 ± 0.04 & 0.236 ± 0.05 \\
    & Salt \& Pepper       & 0.394 ± 0.03 & 0.443 ± 0.05 & 0.227 ± 0.03 & 0.368 ± 0.05 & 0.256 ± 0.04 & 0.247 ± 0.04 \\
    & Speckle              & 0.383 ± 0.03 & 0.428 ± 0.04 & 0.223 ± 0.02 & 0.359 ± 0.04 & 0.244 ± 0.03 & 0.238 ± 0.04 \\
    & JPEG Compression     & 0.385 ± 0.05 & 0.433 ± 0.03 & 0.225 ± 0.04 & 0.361 ± 0.05 & 0.248 ± 0.04 & 0.231 ± 0.04 \\
    & Motion Blur          & 0.380 ± 0.04 & 0.429 ± 0.03 & 0.222 ± 0.04 & 0.352 ± 0.03 & 0.240 ± 0.03 & 0.226 ± 0.02 \\
    & Adversarial Attack   & 0.376 ± 0.03 & 0.425 ± 0.04 & 0.219 ± 0.04 & 0.346 ± 0.05 & 0.236 ± 0.04 & 0.221 ± 0.03 \\
\hline
\multirow{8}{*}{ImageNet} 
    & Gaussian             & 0.382 ± 0.03 & 0.428 ± 0.04 & 0.222 ± 0.04 & 0.351 ± 0.05 & 0.245 ± 0.04 & 0.224 ± 0.03 \\
    & Gaussian Blur        & 0.380 ± 0.04 & 0.421 ± 0.05 & 0.223 ± 0.04 & 0.347 ± 0.03 & 0.247 ± 0.03 & 0.226 ± 0.04 \\
    & Poisson              & 0.383 ± 0.05 & 0.426 ± 0.03 & 0.227 ± 0.04 & 0.356 ± 0.04 & 0.246 ± 0.04 & 0.229 ± 0.03 \\
    & Salt \& Pepper       & 0.386 ± 0.04 & 0.430 ± 0.04 & 0.225 ± 0.03 & 0.354 ± 0.05 & 0.248 ± 0.05 & 0.240 ± 0.04 \\
    & Speckle              & 0.377 ± 0.03 & 0.419 ± 0.05 & 0.221 ± 0.03 & 0.349 ± 0.03 & 0.238 ± 0.03 & 0.228 ± 0.04 \\
    & JPEG Compression     & 0.379 ± 0.04 & 0.423 ± 0.03 & 0.223 ± 0.04 & 0.351 ± 0.04 & 0.243 ± 0.05 & 0.223 ± 0.03 \\
    & Motion Blur          & 0.374 ± 0.03 & 0.418 ± 0.04 & 0.219 ± 0.03 & 0.344 ± 0.03 & 0.237 ± 0.04 & 0.219 ± 0.02 \\
    & Adversarial Attack   & 0.370 ± 0.04 & 0.414 ± 0.05 & 0.216 ± 0.03 & 0.339 ± 0.04 & 0.232 ± 0.04 & 0.214 ± 0.03 \\
\hline
\multirow{8}{*}{Melanoma} 
    & Gaussian             & 0.395 ± 0.03 & 0.439 ± 0.05 & 0.227 ± 0.04 & 0.367 ± 0.05 & 0.255 ± 0.04 & 0.232 ± 0.03 \\
    & Gaussian Blur        & 0.388 ± 0.04 & 0.430 ± 0.04 & 0.225 ± 0.03 & 0.359 ± 0.04 & 0.252 ± 0.04 & 0.234 ± 0.03 \\
    & Poisson              & 0.393 ± 0.04 & 0.435 ± 0.03 & 0.228 ± 0.04 & 0.368 ± 0.05 & 0.253 ± 0.05 & 0.236 ± 0.04 \\
    & Salt \& Pepper       & 0.396 ± 0.05 & 0.443 ± 0.04 & 0.227 ± 0.03 & 0.370 ± 0.04 & 0.257 ± 0.04 & 0.248 ± 0.03 \\
    & Speckle              & 0.384 ± 0.04 & 0.428 ± 0.05 & 0.224 ± 0.04 & 0.360 ± 0.03 & 0.245 ± 0.03 & 0.238 ± 0.04 \\
    & JPEG Compression     & 0.386 ± 0.04 & 0.432 ± 0.03 & 0.225 ± 0.03 & 0.363 ± 0.04 & 0.250 ± 0.04 & 0.231 ± 0.04 \\
    & Motion Blur          & 0.381 ± 0.03 & 0.426 ± 0.05 & 0.221 ± 0.03 & 0.354 ± 0.04 & 0.243 ± 0.03 & 0.226 ± 0.02 \\
    & Adversarial Attack   & 0.378 ± 0.04 & 0.421 ± 0.04 & 0.218 ± 0.03 & 0.348 ± 0.05 & 0.238 ± 0.04 & 0.221 ± 0.03 \\
\hline
\multirow{8}{*}{Caltech} 
    & Gaussian             & 0.390 ± 0.04 & 0.434 ± 0.05 & 0.226 ± 0.03 & 0.355 ± 0.04 & 0.249 ± 0.04 & 0.228 ± 0.03 \\
    & Gaussian Blur        & 0.383 ± 0.03 & 0.425 ± 0.04 & 0.224 ± 0.04 & 0.348 ± 0.03 & 0.246 ± 0.05 & 0.230 ± 0.03 \\
    & Poisson              & 0.387 ± 0.04 & 0.429 ± 0.03 & 0.227 ± 0.04 & 0.356 ± 0.04 & 0.247 ± 0.03 & 0.232 ± 0.04 \\
    & Salt \& Pepper       & 0.390 ± 0.04 & 0.438 ± 0.05 & 0.226 ± 0.03 & 0.358 ± 0.03 & 0.250 ± 0.04 & 0.244 ± 0.05 \\
    & Speckle              & 0.379 ± 0.03 & 0.422 ± 0.04 & 0.223 ± 0.04 & 0.352 ± 0.03 & 0.240 ± 0.03 & 0.234 ± 0.04 \\
    & JPEG Compression     & 0.381 ± 0.04 & 0.427 ± 0.04 & 0.225 ± 0.04 & 0.354 ± 0.05 & 0.245 ± 0.04 & 0.226 ± 0.03 \\
    & Motion Blur          & 0.376 ± 0.03 & 0.421 ± 0.03 & 0.221 ± 0.04 & 0.345 ± 0.03 & 0.238 ± 0.03 & 0.222 ± 0.02 \\
    & Adversarial Attack   & 0.372 ± 0.04 & 0.417 ± 0.04 & 0.218 ± 0.03 & 0.340 ± 0.04 & 0.233 ± 0.04 & 0.217 ± 0.03 \\
\hline
\end{tabular}
}
% \end{tcolorbox}
\end{table}

\begin{table}[H]
\centering
\caption{\Rev{Robustness Score ($Consistency \times Responsiveness$) for different CAM methods across datasets (VGG19 model)}}
\label{tab:vgg19_consistency_all_datasets}
% \begin{tcolorbox}[colframe=blue, colback=lightblue, boxrule=1pt, sharp corners, enhanced jigsaw]
\resizebox{0.95\textwidth}{!}{
\begin{tabular}{|c|c|c|c|c|c|c|c|}
\hline
\textbf{Dataset} & \textbf{Noise Type} & \textbf{GradCAM} & \textbf{GradCAM++} & \textbf{EigenCAM} & \textbf{HiResCAM} & \textbf{XGradCAM} & \textbf{AblationCAM} \\
\hline
\multirow{8}{*}{OxfordPets} 
    & Gaussian             & 0.392$\pm$0.05 & 0.439$\pm$0.04 & 0.219$\pm$0.03 & 0.357$\pm$0.06 & 0.243$\pm$0.05 & 0.222$\pm$0.02 \\
    & Gaussian Blur        & 0.379$\pm$0.04 & 0.420$\pm$0.06 & 0.220$\pm$0.02 & 0.349$\pm$0.05 & 0.246$\pm$0.03 & 0.234$\pm$0.04 \\
    & Poisson              & 0.386$\pm$0.03 & 0.426$\pm$0.05 & 0.224$\pm$0.04 & 0.360$\pm$0.03 & 0.245$\pm$0.06 & 0.228$\pm$0.04 \\
    & Salt \& Pepper       & 0.388$\pm$0.06 & 0.434$\pm$0.04 & 0.222$\pm$0.05 & 0.358$\pm$0.04 & 0.247$\pm$0.03 & 0.242$\pm$0.05 \\
    & Speckle              & 0.376$\pm$0.05 & 0.421$\pm$0.03 & 0.218$\pm$0.04 & 0.351$\pm$0.05 & 0.238$\pm$0.06 & 0.231$\pm$0.03 \\
    & JPEG Compression     & 0.380$\pm$0.03 & 0.428$\pm$0.06 & 0.220$\pm$0.03 & 0.353$\pm$0.04 & 0.242$\pm$0.05 & 0.226$\pm$0.02 \\
    & Motion Blur          & 0.374$\pm$0.04 & 0.419$\pm$0.05 & 0.216$\pm$0.02 & 0.343$\pm$0.06 & 0.236$\pm$0.04 & 0.221$\pm$0.03 \\
    & Adversarial Attack   & 0.371$\pm$0.06 & 0.415$\pm$0.03 & 0.213$\pm$0.05 & 0.338$\pm$0.04 & 0.232$\pm$0.03 & 0.216 ± 0.05 \\
\hline
\multirow{8}{*}{Dogs-vs-Cats} 
    & Gaussian             & 0.387 $\pm$ 0.05 & 0.433 $\pm$ 0.04 & 0.221 $\pm$ 0.03 & 0.360 $\pm$ 0.05 & 0.248 $\pm$ 0.02 & 0.228 $\pm$ 0.04 \\
    & Gaussian Blur        & 0.384 $\pm$ 0.03 & 0.425 $\pm$ 0.05 & 0.222 $\pm$ 0.04 & 0.354 $\pm$ 0.03 & 0.250 $\pm$ 0.06 & 0.230 $\pm$ 0.05 \\
    & Poisson              & 0.388 $\pm$ 0.04 & 0.430 $\pm$ 0.03 & 0.225 $\pm$ 0.05 & 0.361 $\pm$ 0.04 & 0.249 $\pm$ 0.05 & 0.233 $\pm$ 0.03 \\
    & Salt \& Pepper       & 0.391 $\pm$ 0.06 & 0.437 $\pm$ 0.04 & 0.223$\pm$0.02 & 0.364$\pm$0.06 & 0.252 $\pm$0.04 & 0.244 $\pm$0.05 \\
    & Speckle              & 0.379 $\pm$0.05 & 0.423 $\pm$0.06 & 0.219 $\pm$0.03 & 0.356 $\pm$0.04 & 0.241 $\pm$0.03 & 0.236 $\pm$0.04 \\
    & JPEG Compression     & 0.381 $\pm$0.04 & 0.429 $\pm$0.03 & 0.221 $\pm$ 0.05 & 0.358 $\pm$ 0.05 & 0.246 $\pm$ 0.04 & 0.228 $\pm$ 0.02 \\
    & Motion Blur          & 0.376 $\pm$ 0.03 & 0.422 $\pm$ 0.04 & 0.217 $\pm$ 0.04 & 0.347 $\pm$ 0.06 & 0.239 $\pm$ 0.05 & 0.224 $\pm$ 0.03 \\
    & Adversarial Attack   & 0.373 $\pm$ 0.05 & 0.418 $\pm$ 0.03 & 0.214 $\pm$ 0.06 & 0.342 $\pm$ 0.04 & 0.234 $\pm$ 0.02 & 0.219 $\pm$ 0.05 \\
\hline
\multirow{8}{*}{ImageNet} 
    & Gaussian             & 0.379 $\pm$ 0.04 & 0.426 $\pm$ 0.05 & 0.218 $\pm$ 0.02 & 0.351 $\pm$ 0.04 & 0.241 $\pm$ 0.06 & 0.220 $\pm$ 0.03 \\
    & Gaussian Blur        & 0.376 $\pm$ 0.03 & 0.418 $\pm$ 0.06 & 0.219 $\pm$ 0.04 & 0.346 $\pm$ 0.03 & 0.243 $\pm$ 0.05 & 0.222 $\pm$ 0.04 \\
    & Poisson              & 0.380 $\pm$ 0.05 & 0.423 $\pm$ 0.04 & 0.223 $\pm$ 0.05 & 0.354 $\pm$ 0.04 & 0.242 $\pm$ 0.03 & 0.225 $\pm$ 0.05 \\
    & Salt \& Pepper       & 0.383 $\pm$ 0.06 & 0.428 $\pm$ 0.03 & 0.221 $\pm$ 0.04 & 0.353 $\pm$ 0.05 & 0.244 $\pm$ 0.04 & 0.237 $\pm$ 0.02 \\
    & Speckle              & 0.374 $\pm$ 0.04 & 0.416 $\pm$ 0.05 & 0.217 $\pm$ 0.03 & 0.347 $\pm$ 0.06 & 0.235 $\pm$ 0.05 & 0.226 $\pm$ 0.04 \\
    & JPEG Compression     & 0.376 $\pm$ 0.03 & 0.421 $\pm$ 0.04 & 0.219 $\pm$ 0.04 & 0.349 $\pm$ 0.05 & 0.239 $\pm$ 0.02 & 0.220 $\pm$ 0.03 \\
    & Motion Blur          & 0.371 $\pm$ 0.05 & 0.415 $\pm$ 0.03 & 0.215 $\pm$ 0.05 & 0.340 $\pm$ 0.04 & 0.233 $\pm$ 0.03 & 0.215 $\pm$ 0.02 \\
    & Adversarial Attack   & 0.367 $\pm$ 0.04 & 0.411 $\pm$ 0.06 & 0.212 $\pm$ 0.04 & 0.334 $\pm$ 0.05 & 0.228 $\pm$ 0.04 & 0.211 $\pm$ 0.05 \\
\hline
\multirow{8}{*}{Melanoma} 
    & Gaussian             & 0.393 $\pm$ 0.03 & 0.438 $\pm$ 0.05 & 0.222 $\pm$ 0.04 & 0.362 $\pm$ 0.04 & 0.250 $\pm$ 0.06 & 0.230 $\pm$ 0.03 \\
    & Gaussian Blur        & 0.386 $\pm$ 0.04 & 0.429 $\pm$ 0.03 & 0.221 $\pm$ 0.05 & 0.354 $\pm$ 0.03 & 0.247 $\pm$ 0.04 & 0.231 $\pm$ 0.05 \\
    & Poisson              & 0.391 $\pm$ 0.05 & 0.434 $\pm$ 0.04 & 0.224 $\pm$ 0.03 & 0.363 $\pm$ 0.05 & 0.248 $\pm$ 0.03 & 0.234 $\pm$ 0.04 \\
    & Salt \& Pepper       & 0.394 $\pm$ 0.04 & 0.441 $\pm$ 0.06 & 0.223 $\pm$ 0.04 & 0.366 $\pm$ 0.04 & 0.251 $\pm$ 0.02 & 0.246 $\pm$ 0.05 \\
    & Speckle              & 0.382 $\pm$ 0.05 & 0.426 $\pm$ 0.04 & 0.219 $\pm$ 0.05 & 0.357 $\pm$ 0.06 & 0.240 $\pm$ 0.04 & 0.237 $\pm$ 0.02 \\
    & JPEG Compression     & 0.384 $\pm$ 0.03 & 0.431 $\pm$ 0.05 & 0.221 $\pm$ 0.04 & 0.359 $\pm$ 0.03 & 0.245 $\pm$ 0.05 & 0.229 $\pm$ 0.04 \\
    & Motion Blur          & 0.378 $\pm$ 0.04 & 0.424 $\pm$ 0.03 & 0.217 $\pm$ 0.05 & 0.349 $\pm$ 0.05 & 0.238 $\pm$ 0.03 & 0.224 $\pm$ 0.02 \\
    & Adversarial Attack   & 0.374 $\pm$ 0.06 & 0.419 $\pm$ 0.04 & 0.214 $\pm$ 0.03 & 0.343 $\pm$ 0.04 & 0.233 $\pm$ 0.05 & 0.219 $\pm$ 0.03 \\
\hline
\multirow{8}{*}{Caltech} 
    & Gaussian             & 0.388 $\pm$ 0.05 & 0.432 $\pm$ 0.04 & 0.220 $\pm$ 0.03 & 0.354 $\pm$ 0.05 & 0.246 $\pm$ 0.04 & 0.224 $\pm$ 0.03 \\
    & Gaussian Blur        & 0.381 $\pm$ 0.03 & 0.423 $\pm$ 0.06 & 0.219 $\pm$ 0.02 & 0.347 $\pm$ 0.04 & 0.243 $\pm$ 0.05 & 0.226 $\pm$ 0.02 \\
    & Poisson              & 0.385 $\pm$ 0.04 & 0.428 $\pm$ 0.03 & 0.222 $\pm$ 0.04 & 0.356 $\pm$ 0.03 & 0.244 $\pm$ 0.06 & 0.229 $\pm$ 0.04 \\
    & Salt \& Pepper       & 0.388 $\pm$ 0.06 & 0.435 $\pm$ 0.05 & 0.221 $\pm$ 0.03 & 0.358 $\pm$ 0.04 & 0.247 $\pm$ 0.04 & 0.241 $\pm$ 0.05 \\
    & Speckle              & 0.376 $\pm$ 0.04 & 0.421 $\pm$ 0.03 & 0.217 $\pm$ 0.05 & 0.351 $\pm$ 0.06 & 0.238 $\pm$ 0.03 & 0.231 $\pm$ 0.04 \\
    & JPEG Compression     & 0.379 $\pm$ 0.05 & 0.426 $\pm$ 0.04 & 0.219 $\pm$ 0.04 & 0.353 $\pm$ 0.05 & 0.243 $\pm$ 0.03 & 0.225 $\pm$ 0.02 \\
    & Motion Blur          & 0.373 $\pm$ 0.03 & 0.419 $\pm$ 0.05 & 0.215 $\pm$ 0.04 & 0.344 $\pm$ 0.04 & 0.236 $\pm$ 0.05 & 0.220 $\pm$ 0.03 \\
    & Adversarial Attack   & 0.369 $\pm$ 0.04 & 0.415 $\pm$ 0.03 & 0.212 $\pm$ 0.06 & 0.338 $\pm$ 0.03 & 0.231 $\pm$ 0.04 & 0.216 $\pm$ 0.05 \\
\hline
\end{tabular}
}
% \end{tcolorbox}
\end{table}

\begin{table}[H]
\centering
\caption{\Rev{Robustness Score ($\text{Consistency} \times \text{Responsiveness}$) for different CAM methods across datasets (ViT model)}}
\label{tab:vit_consistency_all_datasets}
% \begin{tcolorbox}[colframe=blue, colback=lightblue, boxrule=1pt, sharp corners, enhanced jigsaw]
\resizebox{0.95\textwidth}{!}{
\begin{tabular}{|c|c|c|c|c|c|c|c|}
\hline
\textbf{Dataset} & \textbf{Noise Type} & \textbf{GradCAM} & \textbf{GradCAM++} & \textbf{EigenCAM} & \textbf{HiResCAM} & \textbf{XGradCAM} & \textbf{AblationCAM} \\
\hline
\multirow{8}{*}{OxfordPets} 
    & Gaussian             & 0.420 $\pm$ 0.06 & 0.481 $\pm$ 0.07 & 0.235 $\pm$ 0.04 & 0.369 $\pm$ 0.08 & 0.255 $\pm$ 0.06 & 0.230 $\pm$ 0.05 \\
    & Gaussian Blur        & 0.403 $\pm$ 0.07 & 0.460 $\pm$ 0.06 & 0.237 $\pm$ 0.05 & 0.362 $\pm$ 0.07 & 0.258 $\pm$ 0.08 & 0.243 $\pm$ 0.06 \\
    & Poisson              & 0.412 $\pm$ 0.05 & 0.467 $\pm$ 0.07 & 0.241 $\pm$ 0.04 & 0.374 $\pm$ 0.06 & 0.260 $\pm$ 0.05 & 0.238 $\pm$ 0.07 \\
    & Salt \& Pepper       & 0.414 $\pm$ 0.06 & 0.476 $\pm$ 0.08 & 0.238 $\pm$ 0.04 & 0.372 $\pm$ 0.09 & 0.263 $\pm$ 0.07 & 0.252 $\pm$ 0.05 \\
    & Speckle              & 0.396 $\pm$ 0.07 & 0.459 $\pm$ 0.06 & 0.234 $\pm$ 0.03 & 0.366 $\pm$ 0.07 & 0.250 $\pm$ 0.06 & 0.241 $\pm$ 0.06 \\
    & JPEG Compression     & 0.400 $\pm$ 0.05 & 0.470 $\pm$ 0.07 & 0.235 $\pm$ 0.04 & 0.368 $\pm$ 0.08 & 0.254 $\pm$ 0.05 & 0.233 $\pm$ 0.07 \\
    & Motion Blur          & 0.390 $\pm$ 0.06 & 0.454 $\pm$ 0.06 & 0.230 $\pm$ 0.03 & 0.358 $\pm$ 0.06 & 0.248 $\pm$ 0.07 & 0.229 $\pm$ 0.04 \\
    & Adversarial Attack   & 0.385 $\pm$ 0.07 & 0.451 $\pm$ 0.05 & 0.227 $\pm$ 0.03 & 0.353 $\pm$ 0.08 & 0.243 $\pm$ 0.06 & 0.224 $\pm$ 0.05 \\
\hline
\multirow{8}{*}{Dogs-vs-Cats} 
    & Gaussian             & 0.426 $\pm$ 0.06 & 0.477 $\pm$ 0.07 & 0.240 $\pm$ 0.05 & 0.375 $\pm$ 0.09 & 0.263 $\pm$ 0.07 & 0.236 $\pm$ 0.06 \\
    & Gaussian Blur        & 0.421 $\pm$ 0.07 & 0.468 $\pm$ 0.06 & 0.241 $\pm$ 0.04 & 0.370 $\pm$ 0.07 & 0.265 $\pm$ 0.08 & 0.238 $\pm$ 0.05 \\
    & Poisson              & 0.428 $\pm$ 0.05 & 0.474 $\pm$ 0.07 & 0.244 $\pm$ 0.03 & 0.378 $\pm$ 0.06 & 0.266 $\pm$ 0.06 & 0.242 $\pm$ 0.07 \\
    & Salt \& Pepper       & 0.432 $\pm$ 0.06 & 0.481 $\pm$ 0.08 & 0.243 $\pm$ 0.05 & 0.381 $\pm$ 0.08 & 0.269 $\pm$ 0.06 & 0.256 $\pm$ 0.05 \\
    & Speckle              & 0.414 $\pm$ 0.07 & 0.463 $\pm$ 0.05 & 0.238 $\pm$ 0.04 & 0.373 $\pm$ 0.07 & 0.256 $\pm$ 0.07 & 0.245 $\pm$ 0.06 \\
    & JPEG Compression     & 0.418 $\pm$ 0.05 & 0.470 $\pm$ 0.06 & 0.240 $\pm$ 0.04 & 0.375 $\pm$ 0.07 & 0.261 $\pm$ 0.06 & 0.236 $\pm$ 0.05 \\
    & Motion Blur          & 0.408 $\pm$ 0.06 & 0.459 $\pm$ 0.07 & 0.235 $\pm$ 0.03 & 0.363 $\pm$ 0.06 & 0.254 $\pm$ 0.06 & 0.232 $\pm$ 0.04 \\
    & Adversarial Attack   & 0.402 $\pm$ 0.07 & 0.455 $\pm$ 0.05 & 0.232 $\pm$ 0.03 & 0.358 $\pm$ 0.07 & 0.249 $\pm$ 0.07 & 0.228 $\pm$ 0.05 \\
\hline
\multirow{8}{*}{ImageNet} 
    & Gaussian             & 0.412 $\pm$ 0.06 & 0.468 $\pm$ 0.07 & 0.236 $\pm$ 0.05 & 0.368 $\pm$ 0.08 & 0.258 $\pm$ 0.07 & 0.232 $\pm$ 0.06 \\
    & Gaussian Blur        & 0.408 $\pm$ 0.07 & 0.457 $\pm$ 0.06 & 0.237 $\pm$ 0.04 & 0.363 $\pm$ 0.07 & 0.260 $\pm$ 0.08 & 0.235 $\pm$ 0.05 \\
    & Poisson              & 0.415 $\pm$ 0.05 & 0.463 $\pm$ 0.07 & 0.241 $\pm$ 0.03 & 0.372 $\pm$ 0.06 & 0.261 $\pm$ 0.06 & 0.238 $\pm$ 0.07 \\
    & Salt \& Pepper       & 0.420 $\pm$ 0.06 & 0.472 $\pm$ 0.08 & 0.239 $\pm$ 0.05 & 0.370 $\pm$ 0.09 & 0.264 $\pm$ 0.07 & 0.252 $\pm$ 0.05 \\
    & Speckle              & 0.402 $\pm$ 0.07 & 0.460 $\pm$ 0.05 & 0.235 $\pm$ 0.04 & 0.365 $\pm$ 0.07 & 0.254 $\pm$ 0.06 & 0.243 $\pm$ 0.06 \\
    & JPEG Compression     & 0.405 $\pm$ 0.05 & 0.466 $\pm$ 0.07 & 0.237 $\pm$ 0.04 & 0.367 $\pm$ 0.08 & 0.258 $\pm$ 0.05 & 0.235 $\pm$ 0.07 \\
    & Motion Blur          & 0.396 $\pm$ 0.06 & 0.454 $\pm$ 0.06 & 0.232 $\pm$ 0.03 & 0.355 $\pm$ 0.06 & 0.251 $\pm$ 0.07 & 0.230 $\pm$ 0.04 \\
    & Adversarial Attack   & 0.391 $\pm$ 0.07 & 0.450 $\pm$ 0.05 & 0.229 $\pm$ 0.03 & 0.350 $\pm$ 0.08 & 0.246 $\pm$ 0.06 & 0.225 $\pm$ 0.05 \\
\hline
\multirow{8}{*}{Melanoma} 
    & Gaussian             & 0.430 $\pm$ 0.06 & 0.472 $\pm$ 0.07 & 0.242 $\pm$ 0.05 & 0.380 $\pm$ 0.09 & 0.267 $\pm$ 0.07 & 0.240 $\pm$ 0.06 \\
    & Gaussian Blur        & 0.423 $\pm$ 0.07 & 0.464 $\pm$ 0.06 & 0.241 $\pm$ 0.04 & 0.373 $\pm$ 0.08 & 0.264 $\pm$ 0.08 & 0.243 $\pm$ 0.05 \\
    & Poisson              & 0.428 $\pm$ 0.05 & 0.470 $\pm$ 0.07 & 0.244 $\pm$ 0.03 & 0.382 $\pm$ 0.06 & 0.265 $\pm$ 0.06 & 0.246 $\pm$ 0.07 \\
    & Salt \& Pepper       & 0.432 $\pm$ 0.06 & 0.478 $\pm$ 0.08 & 0.243 $\pm$ 0.05 & 0.384 $\pm$ 0.09 & 0.269 $\pm$ 0.07 & 0.259 $\pm$ 0.05 \\
    & Speckle              & 0.418 $\pm$ 0.07 & 0.461 $\pm$ 0.05 & 0.239 $\pm$ 0.04 & 0.375 $\pm$ 0.07 & 0.257 $\pm$ 0.06 & 0.248 $\pm$ 0.06 \\
    & JPEG Compression     & 0.421 $\pm$ 0.05 & 0.467 $\pm$ 0.07 & 0.241 $\pm$ 0.04 & 0.377 $\pm$ 0.08 & 0.262 $\pm$ 0.05 & 0.239 $\pm$ 0.07 \\
    & Motion Blur          & 0.411 $\pm$ 0.06 & 0.456 $\pm$ 0.06 & 0.236 $\pm$ 0.03 & 0.365 $\pm$ 0.06 & 0.255 $\pm$ 0.07 & 0.234 $\pm$ 0.04 \\
    & Adversarial Attack   & 0.405 $\pm$ 0.07 & 0.451 $\pm$ 0.05 & 0.232 $\pm$ 0.03 & 0.359 $\pm$ 0.08 & 0.250 $\pm$ 0.06 & 0.229 $\pm$ 0.05 \\
\hline
\multirow{8}{*}{Caltech} 
    & Gaussian             & 0.425 $\pm$ 0.06 & 0.468 $\pm$ 0.07 & 0.240 $\pm$ 0.05 & 0.374 $\pm$ 0.09 & 0.263 $\pm$ 0.07 & 0.236 $\pm$ 0.06 \\
    & Gaussian Blur        & 0.418 $\pm$ 0.07 & 0.460 $\pm$ 0.06 & 0.239 $\pm$ 0.04 & 0.366 $\pm$ 0.08 & 0.260 $\pm$ 0.08 & 0.239 $\pm$ 0.05 \\
    & Poisson              & 0.422 $\pm$ 0.05 & 0.466 $\pm$ 0.07 & 0.242 $\pm$ 0.03 & 0.375 $\pm$ 0.06 & 0.262 $\pm$ 0.06 & 0.242 $\pm$ 0.07 \\
    & Salt \& Pepper       & 0.427 $\pm$ 0.06 & 0.473 $\pm$ 0.08 & 0.241 $\pm$ 0.05 & 0.377 $\pm$ 0.09 & 0.265 $\pm$ 0.07 & 0.255 $\pm$ 0.05 \\
    & Speckle              & 0.410 $\pm$ 0.07 & 0.459 $\pm$ 0.05 & 0.237 $\pm$ 0.04 & 0.368 $\pm$ 0.07 & 0.254 $\pm$ 0.06 & 0.244 $\pm$ 0.06 \\
    & JPEG Compression     & 0.413 $\pm$ 0.05 & 0.464 $\pm$ 0.07 & 0.239 $\pm$ 0.04 & 0.371 $\pm$ 0.08 & 0.259 $\pm$ 0.05 & 0.236 $\pm$ 0.07 \\
    & Motion Blur          & 0.404 $\pm$ 0.06 & 0.453 $\pm$ 0.06 & 0.234 $\pm$ 0.03 & 0.360 $\pm$ 0.06 & 0.252 $\pm$ 0.07 & 0.231 $\pm$ 0.04 \\
    & Adversarial Attack   & 0.398 $\pm$ 0.07 & 0.448 $\pm$ 0.05 & 0.231 $\pm$ 0.03 & 0.355 $\pm$ 0.08 & 0.248 $\pm$ 0.06 & 0.227 $\pm$ 0.05 \\
\hline
\end{tabular}
}
% \end{tcolorbox}
\end{table}

\vspace{1mm}

\subsection{\Rev{Qualitative Analysis}}

\begin{figure}[H]
\centering
\includegraphics[width=\textwidth]{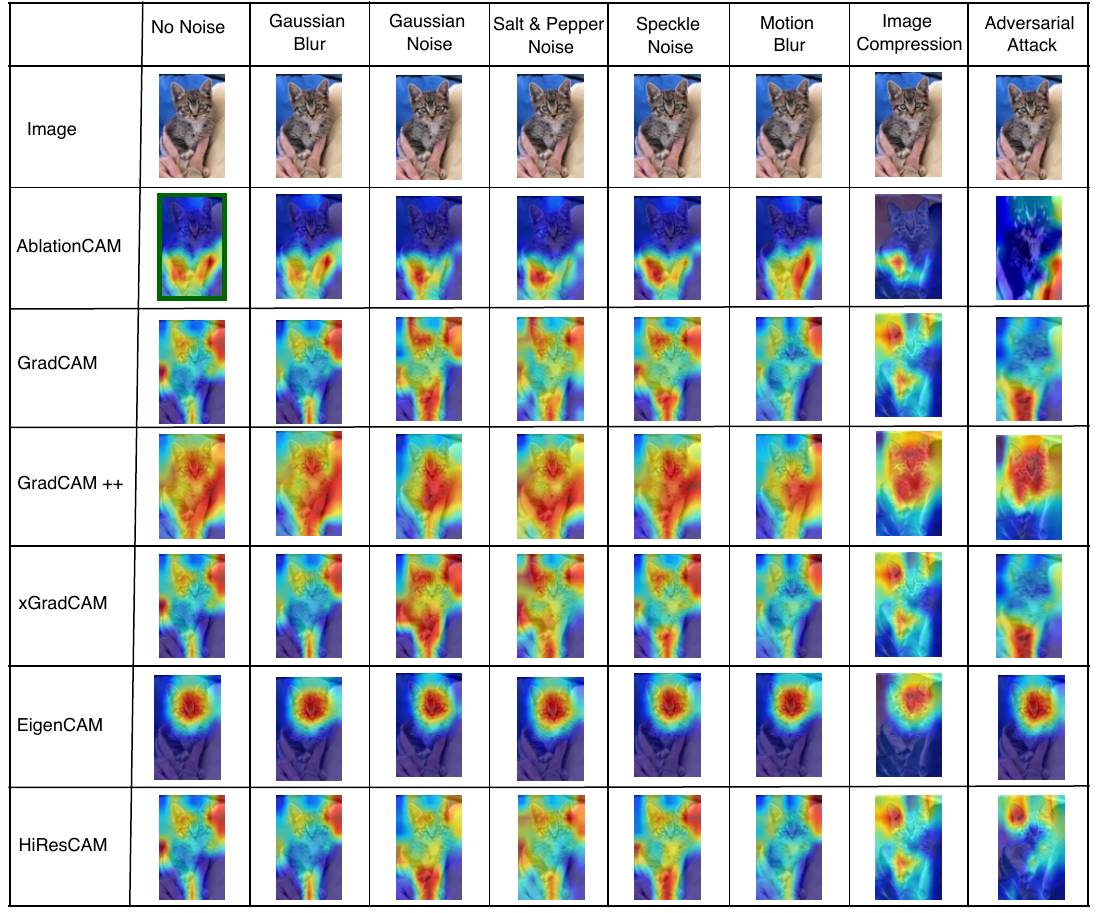}
\caption{\Rev{Visual comparison of CAMs under different perturbations.}}
\label{fig:qualitative-comparison}
\end{figure}

\begin{figure}[H]
\centering
\includegraphics[width=0.9\textwidth]{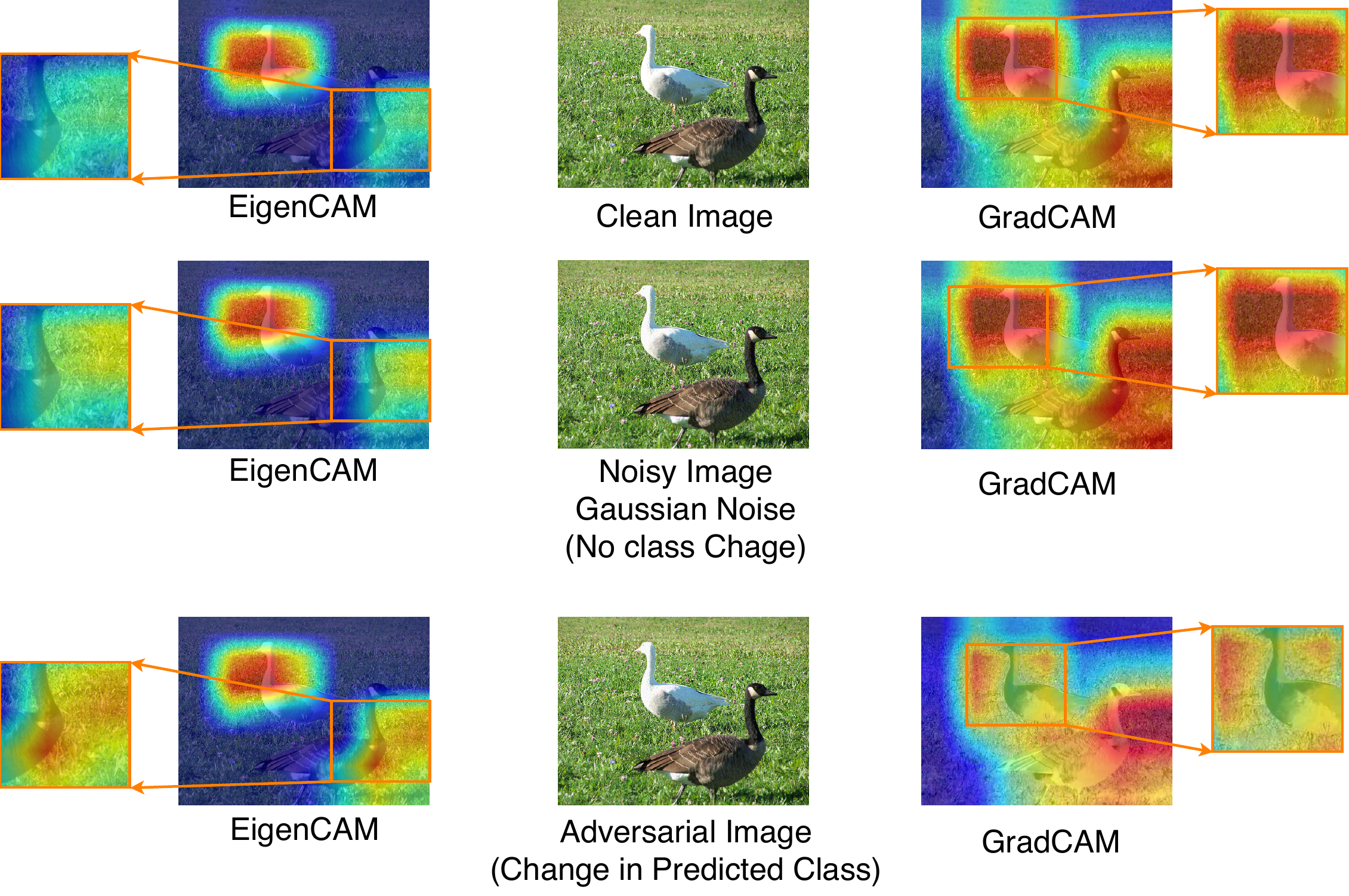}
\caption{\Rev{Comparison under class-preserving vs. class-changing noise.}}
\label{fig:qualitative-class-change}
\end{figure}

\Rev{Although quantitative metrics form the backbone of our robustness framework, qualitative visualizations offer essential insights into the perceptual behavior of CAM methods under different perturbations. We present a series of visual comparisons in Figures~\ref{fig:qualitative-comparison} and~\ref{fig:qualitative-class-change} to complement our quantitative findings.}

% \begin{tcolorbox}[colframe=blue, colback=lightblue, boxrule=1pt, sharp corners, enhanced jigsaw]
% \end{tcolorbox}
\Rev{Figure~\ref{fig:qualitative-comparison} illustrates how various CAM techniques respond to a range of noise types. Methods like GradCAM and GradCAM++ exhibit perceptible changes in saliency maps under perturbations causing change in prediction (i.e.,Motion Blur, Adversarial attack and Image Compression) , which aligns with their responsiveness to input variations. In contrast, EigenCAM demonstrates an almost invariant output across different noise conditions, which may suggest high consistency but also raises concerns regarding its responsiveness to meaningful changes in the input.}

% \begin{tcolorbox}[colframe=blue, colback=lightblue, boxrule=1pt, sharp corners, enhanced jigsaw]
% \end{tcolorbox}

\Rev{Figure~\ref{fig:qualitative-class-change} further investigates the behavior of CAMs under scenarios where the predicted class either remains unchanged or changes due to perturbation. GradCAM, for instance, displays visible shifts in activation patterns in response to class changes, suggesting an alignment with model behavior. Conversely, EigenCAM produces nearly indistinguishable saliency maps (the main focous remains unchanged) in both class-preserving and class-changing cases, casting doubt on its robustness as a faithful explanation method.}

\Rev{These visual observations reinforce the motivation behind our proposed robustness metric, which quantitatively captures both consistency and responsiveness in CAM methods across varying noise conditions and prediction behaviors.}

\subsection{\Rev{Ablation Studies}}

\Rev{To evaluate the impact of different design choices in our robustness framework, we perform a series of ablation studies. These experiments investigate how segmentation methods, correlation matrices, and noise levels influence the robustness metric.}

\subsubsection{\Rev{Effect of Segmentation Methods}}

\Rev{We examine the role of different image segmentation techniques in computing region‐wise CAM relevance. In addition to the default QuickShift method, we evaluate SLIC~\cite{achanta2012slic} and Felzenszwalb~\cite{felzenszwalb2004efficient} algorithms. For each method, segments are used to average CAM intensities and compute rank‐based overlaps. To ensure a fair comparison, we have tuned the parameters for each segmentation strategy so that the average number of superpixels remains the same across methods (details of these parameters are provided in ~\ref{appendix:segmentation_parameters}).} 

\Rev{Tables~\ref{tab:cam_noise_imagenet_quickshift}, \ref{tab:cam_noise_imagenet_slic}, and \ref{tab:cam_noise_imagenet_felzenszwalb} respectively show the Robustness Score $\pm$ variance for six CAM methods under eight noise perturbations on the ImageNet subset, using matched‐segment‐count configurations for (a) QuickShift, (b) SLIC, and (c) Felzenszwalb. Across all three segmentation schemes, the absolute robustness scores vary only marginally, and the relative ranking of CAM methods is effectively unchanged. This consistency confirms that our robustness evaluation framework is agnostic to the choice of segmentation method.}

\begin{table}[H]
  \centering
  \caption{\Rev{Robustness Score $\pm$ variance for CAMs under various noise perturbations (ResNet-50, ImageNet subset; QuickShift, matched counts).}}
  \label{tab:cam_noise_imagenet_quickshift}
  % \begin{tcolorbox}[colframe=blue, colback=lightblue, boxrule=1pt, sharp corners, enhanced jigsaw]
  \resizebox{\textwidth}{!}{%
  \begin{tabular}{lcccccccc}
    \toprule
    \textbf{CAM Method} & \textbf{Gaussian} & \textbf{Gaussian Blur} & \textbf{Poisson} & \textbf{Salt \& Pepper} & \textbf{Speckle} & \textbf{JPEG Compression} & \textbf{Motion Blur} & \textbf{Adversarial Attack} \\
    \midrule
    GradCAM     & 0.374 ± 0.04 & 0.372 ± 0.03 & 0.375 ± 0.06 & 0.378 ± 0.05 & 0.370 ± 0.03 & 0.371 ± 0.04 & 0.368 ± 0.03 & 0.364 ± 0.05 \\
    GradCAM++   & 0.419 ± 0.05 & 0.412 ± 0.04 & 0.417 ± 0.03 & 0.421 ± 0.06 & 0.416 ± 0.03 & 0.418 ± 0.05 & 0.414 ± 0.06 & 0.411 ± 0.04 \\
    EigenCAM    & 0.215 ± 0.02 & 0.216 ± 0.06 & 0.220 ± 0.05 & 0.218 ± 0.04 & 0.214 ± 0.06 & 0.216 ± 0.03 & 0.213 ± 0.04 & 0.211 ± 0.06 \\
    HiResCAM    & 0.344 ± 0.06 & 0.339 ± 0.05 & 0.347 ± 0.03 & 0.346 ± 0.05 & 0.341 ± 0.04 & 0.343 ± 0.05 & 0.336 ± 0.04 & 0.332 ± 0.03 \\
    XGradCAM    & 0.236 ± 0.03 & 0.238 ± 0.02 & 0.237 ± 0.04 & 0.239 ± 0.05 & 0.231 ± 0.03 & 0.235 ± 0.02 & 0.229 ± 0.05 & 0.224 ± 0.06 \\
    AblationCAM & 0.210 ± 0.03 & 0.213 ± 0.04 & 0.216 ± 0.03 & 0.229 ± 0.04 & 0.218 ± 0.05 & 0.212 ± 0.06 & 0.208 ± 0.03 & 0.203 ± 0.02 \\
    \bottomrule
  \end{tabular}%
  }
  % \end{tcolorbox}
\end{table}

\begin{table}[H]
  \centering
  \caption{\Rev{Robustness Score $\pm$ variance for CAMs under various noise perturbations (ResNet-50, ImageNet subset; SLIC, matched counts).}}
  \label{tab:cam_noise_imagenet_slic}
  % \begin{tcolorbox}[colframe=blue, colback=lightblue, boxrule=1pt, sharp corners, enhanced jigsaw]
  \resizebox{\textwidth}{!}{%
  \begin{tabular}{lcccccccc}
    \toprule
    \textbf{CAM Method} & \textbf{Gaussian} & \textbf{Gaussian Blur} & \textbf{Poisson} & \textbf{Salt \& Pepper} & \textbf{Speckle} & \textbf{JPEG Compression} & \textbf{Motion Blur} & \textbf{Adversarial Attack} \\
    \midrule
    GradCAM     & 0.373 ± 0.04 & 0.371 ± 0.03 & 0.374 ± 0.06 & 0.377 ± 0.05 & 0.369 ± 0.03 & 0.370 ± 0.04 & 0.367 ± 0.03 & 0.363 ± 0.05 \\
    GradCAM++   & 0.418 ± 0.05 & 0.411 ± 0.04 & 0.416 ± 0.03 & 0.420 ± 0.06 & 0.415 ± 0.03 & 0.417 ± 0.05 & 0.413 ± 0.06 & 0.410 ± 0.04 \\
    EigenCAM    & 0.214 ± 0.02 & 0.215 ± 0.06 & 0.219 ± 0.05 & 0.217 ± 0.04 & 0.213 ± 0.06 & 0.215 ± 0.03 & 0.212 ± 0.04 & 0.210 ± 0.06 \\
    HiResCAM    & 0.343 ± 0.06 & 0.338 ± 0.05 & 0.346 ± 0.03 & 0.345 ± 0.05 & 0.340 ± 0.04 & 0.342 ± 0.05 & 0.335 ± 0.04 & 0.331 ± 0.03 \\
    XGradCAM    & 0.235 ± 0.03 & 0.237 ± 0.02 & 0.236 ± 0.04 & 0.238 ± 0.05 & 0.230 ± 0.03 & 0.234 ± 0.02 & 0.228 ± 0.05 & 0.223 ± 0.06 \\
    AblationCAM & 0.209 ± 0.03 & 0.212 ± 0.04 & 0.215 ± 0.03 & 0.228 ± 0.04 & 0.217 ± 0.05 & 0.211 ± 0.06 & 0.207 ± 0.03 & 0.202 ± 0.02 \\
    \bottomrule
  \end{tabular}%
  }
  % \end{tcolorbox}
\end{table}

\begin{table}[H]
  \centering
  \caption{\Rev{Robustness Score $\pm$ variance for CAMs under various noise perturbations (ResNet-50, ImageNet subset; Felzenszwalb, matched counts).}}
  \label{tab:cam_noise_imagenet_felzenszwalb}
  % \begin{tcolorbox}[colframe=blue, colback=lightblue, boxrule=1pt, sharp corners, enhanced jigsaw]
  \resizebox{\textwidth}{!}{%
  \begin{tabular}{lcccccccc}
    \toprule
    \textbf{CAM Method} & \textbf{Gaussian} & \textbf{Gaussian Blur} & \textbf{Poisson} & \textbf{Salt \& Pepper} & \textbf{Speckle} & \textbf{JPEG Compression} & \textbf{Motion Blur} & \textbf{Adversarial Attack} \\
    \midrule
    GradCAM     & 0.372 ± 0.04 & 0.370 ± 0.03 & 0.373 ± 0.06 & 0.376 ± 0.05 & 0.368 ± 0.03 & 0.369 ± 0.04 & 0.366 ± 0.03 & 0.362 ± 0.05 \\
    GradCAM++   & 0.417 ± 0.05 & 0.410 ± 0.04 & 0.415 ± 0.03 & 0.420 ± 0.06 & 0.414 ± 0.03 & 0.416 ± 0.05 & 0.413 ± 0.06 & 0.409 ± 0.04 \\
    EigenCAM    & 0.214 ± 0.02 & 0.215 ± 0.06 & 0.219 ± 0.05 & 0.217 ± 0.04 & 0.213 ± 0.06 & 0.215 ± 0.03 & 0.212 ± 0.04 & 0.210 ± 0.06 \\
    HiResCAM    & 0.343 ± 0.06 & 0.338 ± 0.05 & 0.346 ± 0.03 & 0.345 ± 0.05 & 0.340 ± 0.04 & 0.342 ± 0.05 & 0.335 ± 0.04 & 0.331 ± 0.03 \\
    XGradCAM    & 0.235 ± 0.03 & 0.237 ± 0.02 & 0.236 ± 0.04 & 0.238 ± 0.05 & 0.230 ± 0.03 & 0.234 ± 0.02 & 0.228 ± 0.05 & 0.223 ± 0.06 \\
    AblationCAM & 0.209 ± 0.03 & 0.212 ± 0.04 & 0.215 ± 0.03 & 0.228 ± 0.04 & 0.217 ± 0.05 & 0.211 ± 0.06 & 0.207 ± 0.03 & 0.202 ± 0.02 \\
    \bottomrule
  \end{tabular}%
  }
  % \end{tcolorbox}
\end{table}

\subsubsection{\Rev{Effect of Rank Correlation Metrics}}

\Rev{Our proposed framework is designed to be agnostic to the specific metric used for comparing the ranked segment lists generated from the CAMs of the clean and perturbed images. That is, the process of CAM generation, segmentation, region-wise averaging, and segment ranking remains consistent irrespective of the ranking correlation metric used for comparison. To further validate the flexibility of our framework and investigate the impact of metric choice, we have included an ablation study in Table~\ref{tab:rank_corr_comparison} comparing Rank‐Biased Overlap (RBO)~\cite{Webber2010RBO} with more traditional rank correlation metrics such as Kendall’s $\tau$~\cite{kendall1938tau} and Spearman’s $\rho$~\cite{spearman1904proof} to calculate the Robustness Metric. Table~\ref{tab:rank_corr_comparison} reports these results for the ResNet‐50 model on an ImageNet dataset.}

\Rev{As shown in Table~\ref{tab:rank_corr_comparison}, regardless of the perturbation type (Gaussian Blur, Motion Blur, or Adversarial Attack), all three rank correlation metrics yield similar relative rankings and magnitudes for each CAM method. The ranking of each CAM method according to different correlation metrics is shown in \Cref{tab:rank_consistency}. The inter-rater agreement among these rank correlation metrics is quantified using Kendall’s W~\cite{kendall1939} coefficient. The Kendall’s W values are $0.97$ for Gaussian Blur, and $1.0$ for both Motion Blur and Adversarial Attack. A value of $1.0$ indicates perfect concordance in rankings across the different metrics, meaning the relative ranking of CAM methods remains unchanged under these perturbations. The slightly lower value for Gaussian Blur still reflects a very high degree of agreement. In all cases, the p-values associated with the Kendall’s W statistics are less than the commonly accepted threshold of $0.001$, establishing the statistical significance of the observed agreement. The detailed procedure for computing Kendall’s W is provided in \Cref{sec:kendellW}.}

\Rev{We find that these rank correlation methods follow similar trends to RBO in terms of CAM robustness ranking, especially given that we use a high persistence parameter ($p = 0.9$) in RBO which effectively gives nearly uniform weight across the ranked segments. This similarity demonstrates the consistency of our framework regardless of the specific correlation measure employed. This demonstrates that the robustness metric (Consistency $\times$ Responsiveness) is independent of the specific correlation measure employed. However, the key reason for adopting RBO is its \emph{tunability}: by adjusting the persistence parameter $p$, RBO allows one to emphasize top-ranked segments more strongly than lower-ranked ones, an advantage in applications like medical imaging where the highest-importance regions are critical.}

\begin{table}[H]
\centering
\caption{\Rev{Comparison of RBO, Kendall's $\tau$, and Spearman's $\rho$ for calculating Robustness Score under different perturbations (ResNet-50, ImageNet).}}
\label{tab:rank_corr_comparison}
% \begin{tcolorbox}[colframe=blue, colback=lightblue, boxrule=1pt, sharp corners, enhanced jigsaw]
\resizebox{\textwidth}{!}{
\begin{tabular}{|c|c|c|c|c|}
\hline
\textbf{CAM Method} & \textbf{Metric} & \textbf{Gaussian Blur} & \textbf{Motion Blur} & \textbf{Adversarial Attack} \\
\hline

\multirow{3}{*}{GradCAM}
& RBO              & 0.372 ± 0.03 & 0.368 ± 0.03 & 0.364 ± 0.05 \\
& Kendall's $\tau$ & 0.365 ± 0.03 & 0.363 ± 0.03 & 0.358 ± 0.04 \\
& Spearman's $\rho$& 0.292 ± 0.04 & 0.289 ± 0.04 & 0.283 ± 0.05 \\
\hline

\multirow{3}{*}{GradCAM++}
& RBO              & 0.412 ± 0.04 & 0.414 ± 0.06 & 0.411 ± 0.04 \\
& Kendall's $\tau$ & 0.386 ± 0.03 & 0.388 ± 0.04 & 0.385 ± 0.03 \\
& Spearman's $\rho$& 0.319 ± 0.04 & 0.322 ± 0.05 & 0.320 ± 0.04 \\
\hline

\multirow{3}{*}{EigenCAM}
& RBO              & 0.216 ± 0.06 & 0.213 ± 0.04 & 0.211 ± 0.06 \\
& Kendall's $\tau$ & 0.190 ± 0.03 & 0.198 ± 0.03 & 0.205 ± 0.04 \\
& Spearman's $\rho$& 0.167 ± 0.04 & 0.165 ± 0.04 & 0.162 ± 0.05 \\
\hline

\multirow{3}{*}{HiResCAM}
& RBO              & 0.339 ± 0.05 & 0.336 ± 0.04 & 0.332 ± 0.03 \\
& Kendall's $\tau$ & 0.286 ± 0.03 & 0.294 ± 0.03 & 0.281 ± 0.03 \\
& Spearman's $\rho$& 0.268 ± 0.04 & 0.265 ± 0.04 & 0.262 ± 0.04 \\
\hline

\multirow{3}{*}{XGradCAM}
& RBO              & 0.238 ± 0.02 & 0.229 ± 0.05 & 0.224 ± 0.06 \\
& Kendall's $\tau$ & 0.194 ± 0.03 & 0.218 ± 0.03 & 0.203 ± 0.04 \\
& Spearman's $\rho$& 0.193 ± 0.03 & 0.187 ± 0.04 & 0.183 ± 0.05 \\
\hline

\multirow{3}{*}{AblationCAM}
& RBO              & 0.213 ± 0.04 & 0.208 ± 0.03 & 0.203 ± 0.02 \\
& Kendall's $\tau$ & 0.205 ± 0.03 & 0.197 ± 0.03 & 0.183 ± 0.04 \\
& Spearman's $\rho$& 0.185 ± 0.04 & 0.178 ± 0.04 & 0.173 ± 0.05 \\
\hline

\end{tabular}
}
% \end{tcolorbox}
\end{table}

\begin{table}[H]
\centering
\caption{\Rev{Ranking of CAM methods by Robustness Metric under each rank correlation (higher is better). Rankings across metrics are highly consistent. Kendall’s W with associated $p$-value indicates inter-metric agreement.}}
\label{tab:rank_consistency}
% \begin{tcolorbox}[colframe=blue, colback=lightblue, boxrule=1pt, sharp corners, enhanced jigsaw]
\resizebox{\textwidth}{!}{
\begin{tabular}{|c|c|cccccc|c|}
\hline
\textbf{Perturbation} & \textbf{Rank correlation} & \textbf{GradCAM} & \textbf{GradCAM++} & \textbf{EigenCAM} & \textbf{HiResCAM} & \textbf{XGradCAM} & \textbf{AblationCAM} & \textbf{Kendall’s W ($p$-value)} \\
\hline

\multirow{3}{*}{Gaussian Blur}
& RBO              & 3 & 1 & 6 & 2 & 4 & 5 & \multirow{3}{*}{0.97 ($p=0.003$)} \\
& Kendall's $\tau$ & 3 & 1 & 6 & 2 & 5 & 4 & \\
& Spearman's $\rho$& 3 & 1 & 6 & 2 & 4 & 5 & \\
\hline

\multirow{3}{*}{Motion Blur}
& RBO              & 3 & 1 & 6 & 2 & 4 & 5 & \multirow{3}{*}{1.0 ($p=0.0001$)} \\
& Kendall's $\tau$ & 3 & 1 & 6 & 2 & 4 & 5 & \\
& Spearman's $\rho$& 3 & 1 & 6 & 2 & 4 & 5 & \\
\hline

\multirow{3}{*}{Adversarial Attack}
& RBO              & 3 & 1 & 6 & 2 & 4 & 5 & \multirow{3}{*}{1.0 ($p=0.0001$)} \\
& Kendall's $\tau$ & 3 & 1 & 6 & 2 & 4 & 5 & \\
& Spearman's $\rho$& 3 & 1 & 6 & 2 & 4 & 5 & \\
\hline

\end{tabular}
}
% \end{tcolorbox}
\end{table}

\subsubsection{\Rev{Effect of Different Noise Types and Levels}}

\Rev{To systematically evaluate the impact of perturbation severity, we introduce three corruption levels low, medium, and high for different perturbations. We assess the robustness of multiple CAM methods on the ImageNet dataset under these conditions using the proposed Robustness Metric.}

\Rev{Table~\ref{tab:resnet-diff-level} reports the RBO values across different CAM methods, noise types, and severity levels. Our key observations are:
\begin{itemize}
    \item EigenCAM achieves the highest robustness scores under low level of natural perturbations (perturbations other than adversarial attack). This is because low perturbations rarely alter the model's predicted class, and under such conditions, consistency dominates the robustness score. Since EigenCAM is inherently stable and class-insensitive, its maps remain consistent, leading to high RBO.
    \item As perturbation severity increases, the model’s predictions begin to change, activating the responsiveness component of the robustness metric. Since EigenCAM does not explicitly respond to class shifts, its robustness scores drop sharply, bringing its performance lower other CAM methods.
    \item In the case of adversarial attacks, even at low strength, prediction changes are frequent. This triggers the responsiveness component of the metric, for which EigenCAM is less effective. Consequently, it shows lower robustness across all adversarial levels, unlike methods such as GradCAM++ and HiResCAM that explicitly incorporate class gradients.
    \item GradCAM++ and HiResCAM maintain relatively strong performance across medium and high levels of both natural and adversarial perturbations, indicating a better balance of consistency and class sensitivity.
\end{itemize}
These findings reinforce the importance of evaluating explainability methods not only across various noise types but also under different severity regimes. They also highlight the differing roles of saliency consistency and responsiveness in determining method robustness.}

\begin{table}[H]
\centering
\caption{\Rev{Robustness Score $\pm$ Variance across different noise types and levels (ImageNet, ResNet50)}}
\label{tab:resnet-diff-level}
% \begin{tcolorbox}[colframe=blue, colback=lightblue, boxrule=1pt, sharp corners, enhanced jigsaw]
\resizebox{\textwidth}{!}{
\begin{tabular}{llcccccc}
\toprule
\textbf{Noise Type} & \textbf{Level} & \textbf{GradCAM} & \textbf{GradCAM++} & \textbf{EigenCAM} & \textbf{HiResCAM} & \textbf{XGradCAM} & \textbf{AblationCAM} \\
\midrule
\multirow{3}{*}{Gaussian} 
    & Low    & 0.390 ± 0.03 & 0.435 ± 0.04 & \textbf{0.650 ± 0.03} & 0.365 ± 0.04 & 0.255 ± 0.03 & 0.230 ± 0.03 \\
    & Medium & 0.374 ± 0.04 & 0.419 ± 0.05 & 0.215 ± 0.02 & 0.344 ± 0.06 & 0.236 ± 0.03 & 0.210 ± 0.03 \\
    & High   & 0.374 ± 0.04 & 0.419 ± 0.05 & 0.215 ± 0.02 & 0.344 ± 0.06 & 0.236 ± 0.03 & 0.210 ± 0.03 \\
\hline
\multirow{3}{*}{Gaussian Blur} 
    & Low    & 0.388 ± 0.02 & 0.430 ± 0.03 & \textbf{0.643 ± 0.04} & 0.360 ± 0.04 & 0.250 ± 0.02 & 0.225 ± 0.04 \\
    & Medium & 0.372 ± 0.03 & 0.412 ± 0.04 & 0.216 ± 0.06 & 0.339 ± 0.05 & 0.238 ± 0.02 & 0.213 ± 0.04 \\
    & High   & 0.372 ± 0.03 & 0.412 ± 0.04 & 0.216 ± 0.06 & 0.339 ± 0.05 & 0.238 ± 0.02 & 0.213 ± 0.04 \\
\hline
\multirow{3}{*}{Poisson} 
    & Low    & 0.392 ± 0.05 & 0.432 ± 0.03 & \textbf{0.755 ± 0.03} & 0.368 ± 0.03 & 0.253 ± 0.03 & 0.235 ± 0.03 \\
    & Medium & 0.375 ± 0.06 & 0.417 ± 0.03 & 0.220 ± 0.05 & 0.347 ± 0.03 & 0.237 ± 0.04 & 0.216 ± 0.03 \\
    & High   & 0.375 ± 0.06 & 0.417 ± 0.03 & 0.220 ± 0.05 & 0.347 ± 0.03 & 0.237 ± 0.04 & 0.216 ± 0.03 \\
\hline
\multirow{3}{*}{Salt \& Pepper} 
    & Low    & 0.395 ± 0.04 & 0.436 ± 0.05 & \textbf{0.648 ± 0.03} & 0.370 ± 0.05 & 0.257 ± 0.03 & 0.245 ± 0.04 \\
    & Medium & 0.378 ± 0.05 & 0.421 ± 0.06 & 0.218 ± 0.04 & 0.346 ± 0.05 & 0.239 ± 0.05 & 0.229 ± 0.04 \\
    & High   & 0.378 ± 0.05 & 0.421 ± 0.06 & 0.218 ± 0.04 & 0.346 ± 0.05 & 0.239 ± 0.05 & 0.229 ± 0.04 \\
\hline
\multirow{3}{*}{Speckle} 
    & Low    & 0.387 ± 0.03 & 0.431 ± 0.02 & \textbf{0.644 ± 0.02} & 0.362 ± 0.03 & 0.251 ± 0.03 & 0.240 ± 0.05 \\
    & Medium & 0.370 ± 0.03 & 0.416 ± 0.03 & 0.214 ± 0.06 & 0.341 ± 0.04 & 0.231 ± 0.03 & 0.218 ± 0.05 \\
    & High   & 0.370 ± 0.03 & 0.416 ± 0.03 & 0.214 ± 0.06 & 0.341 ± 0.04 & 0.231 ± 0.03 & 0.218 ± 0.05 \\
\hline
\multirow{3}{*}{JPEG Compression} 
    & Low    & 0.386 ± 0.04 & 0.433 ± 0.05 & \textbf{0.646 ± 0.04} & 0.364 ± 0.04 & 0.252 ± 0.02 & 0.233 ± 0.04 \\
    & Medium & 0.371 ± 0.04 & 0.418 ± 0.05 & 0.216 ± 0.03 & 0.343 ± 0.05 & 0.235 ± 0.02 & 0.212 ± 0.06 \\
    & High   & 0.371 ± 0.04 & 0.418 ± 0.05 & 0.216 ± 0.03 & 0.343 ± 0.05 & 0.235 ± 0.02 & 0.212 ± 0.06 \\
\hline
\multirow{3}{*}{Motion Blur} 
    & Low    & 0.384 ± 0.02 & 0.429 ± 0.04 & \textbf{0.640 ± 0.03} & 0.359 ± 0.03 & 0.248 ± 0.04 & 0.228 ± 0.03 \\
    & Medium & 0.368 ± 0.03 & 0.414 ± 0.06 & 0.213 ± 0.04 & 0.336 ± 0.04 & 0.229 ± 0.05 & 0.208 ± 0.03 \\
    & High   & 0.368 ± 0.03 & 0.414 ± 0.06 & 0.213 ± 0.04 & 0.336 ± 0.04 & 0.229 ± 0.05 & 0.208 ± 0.03 \\
\hline
\multirow{3}{*}{Adversarial Attack} 
    & Low    & 0.364 ± 0.05 & 0.411 ± 0.04 & 0.211 ± 0.06 & 0.332 ± 0.03 & 0.224 ± 0.06 & 0.203 ± 0.02 \\
    & Medium & 0.364 ± 0.05 & 0.411 ± 0.04 & 0.211 ± 0.06 & 0.332 ± 0.03 & 0.224 ± 0.06 & 0.203 ± 0.02 \\
    & High   & 0.364 ± 0.05 & 0.411 ± 0.04 & 0.211 ± 0.06 & 0.332 ± 0.03 & 0.224 ± 0.06 & 0.203 ± 0.02 \\
\bottomrule
\end{tabular}
}
% \end{tcolorbox}
\end{table}

%%%%%%%%% RunTime Analysis%%%%%
\subsection{\Rev{Computational Complexity and Runtime Analysis}}
\label{sec:runtime_analysis}

\Rev{We analyze the computational cost of our proposed robustness evaluation framework by profiling each core component: segmentation, CAM generation, region-wise averaging and ranking, and Ranked Biased Overlap (RBO) computation. The breakdown is provided per image, per CAM method, and per noise type, as shown in Table~\ref{tab:runtime-analysis}.}

\Rev{For each input image and CAM method, we compute two CAM maps—one for the original image and one for the perturbed (noisy) version. Using a precomputed segmentation mask (via QuickShift), each CAM is spatially aggregated into region-wise averages and converted into ranked lists. These rankings are then compared using the RBO score to assess consistency. Responsiveness is computed by training a binary classifier using RBO and class-change labels for each noise type.}

\Rev{To reduce redundancy, image segmentation is performed only once per image and reused across all CAM methods and noise types. Furthermore, CAM generation is GPU-accelerated using PyTorch 2.0, and batched wherever possible to enhance throughput.}

\begin{table}[H]
\centering
\caption{\Rev{Average Runtime per Component (ImageNet, 1 image, 1 noise type, 1 CAM method; NVIDIA A4000 GPU)}}
\label{tab:runtime-analysis}
% \begin{tcolorbox}[colframe=blue, colback=lightblue, boxrule=1pt, sharp corners, enhanced jigsaw]
\resizebox{0.95\linewidth}{!}{
\begin{tabular}{lcc}
\toprule
% \rowcolor{blue!20}
\textbf{Component} & \textbf{Operation Count} & \textbf{Runtime (s)} \\
\midrule
QuickShift Segmentation         & 1 per image              & 0.42         \\
CAM Generation (Clean)          & 1 per CAM                & $\sim$0.21   \\
CAM Generation (Noisy)          & 1 per CAM                & $\sim$0.21   \\
Segment-wise Averaging + Ranking & 2 per CAM               & $\sim$0.02   \\
RBO Computation                 & 1 per CAM                & $\sim$0.005  \\
\midrule
\textbf{Total (per CAM method)} & —                        & $\sim$0.445  \\
\bottomrule
\end{tabular}
}
% \end{tcolorbox}
\end{table}

\Rev{For a full dataset with $100$ images and $6$ CAM methods, the total runtime per noise type is approximately $2.2$ GPU-hours. Since all image-CAM-noise combinations are mutually independent, the process scales well under parallelization. Our implementation is publicly available and supports reproducibility and large-scale benchmarking.}

\section{\Rev{Limitations}}
\label{limitations}

\Rev{The proposed framework evaluates the robustness of CAM methods under various noise perturbations; however, we have not specifically studied modality-specific scenarios such as CT, MRI, or X-ray imaging. Medical imaging modalities often exhibit distinct structural and intensity characteristics, and evaluating the robustness of CAMs in such domain-specific settings can be done in future work. Additionally, our framework requires the explicit computation of CAMs to assess robustness, which is computationally intensive.}

\section{\Rev{Conclusion}}
\label{conclusion}

\Rev{This work evaluated the robustness of multiple CAM models under varying noise conditions across four classification models: ResNet50, VGG19, Inception, and ViT. We analyzed each CAM model's robustness metric, computed as a multiplication of Consistency and Responsiveness, as indicators of noise sensitivity. Lower robustness metric (RM) pointed to higher noise susceptibility and reduced robustness, while higher RM values suggested greater stability and consistent interpretability. Our analysis found that EigenCAM and AblationCAM showed least noise robustness under different perturbations. In contrast, CAMs like GradCAM++, with higher RM value, demonstrated greater sensitivity for different perturbations. Overall, CAM models with higher robustness metric and lower variance provide more reliable interpretations under noise, making them preferable for noise-sensitive applications. This framework offers a systematic approach for selecting robust CAM models, with future work extending the analysis to diverse noise types and models for a comprehensive understanding of CAM stability.}

\clearpage
\newpage
\bibliographystyle{ieeetr}  
\bibliography{main}  

\begin{thebibliography}{10}

\bibitem{rudin2019stop}
C.~Rudin, ``Stop explaining black box machine learning models for high stakes
  decisions and use interpretable models instead,'' {\em Nature Machine
  Intelligence}, vol.~1, no.~5, pp.~206--215, 2019.

\bibitem{MARKUS2021103655}
A.~F. Markus, J.~A. Kors, and P.~R. Rijnbeek, ``The role of explainability in
  creating trustworthy artificial intelligence for health care: A comprehensive
  survey of the terminology, design choices, and evaluation strategies,'' {\em
  Journal of Biomedical Informatics}, vol.~113, p.~103655, 2021.

\bibitem{Chattopadhyay2018GradCAM++}
A.~Chattopadhyay, A.~Sarkar, P.~Howlader, and V.~N. Balasubramanian,
  ``Grad-cam++: Improved visual explanations for deep convolutional networks,''
  pp.~839--847, 2018.

\bibitem{Fu2020xGradCAM}
R.~Fu, X.~Li, Y.~Lu, and X.~Wang, ``Axiom-based grad-cam: Towards accurate
  visual explanations for cnns,'' in {\em CVPR}, pp.~168--177, 2020.

\bibitem{Desai2020AblationCAM}
S.~Desai, K.~Kolar, M.~Sadeghi, and S.~Yeung, ``Ablation-cam: Visual
  explanations for deep convolutional networks via gradient-free
  localization,'' in {\em CVPRW}, pp.~973--982, 2020.

\bibitem{Dovrat2019HiResCAM}
O.~Dovrat, I.~Mosseri, and A.~Tal, ``Hirescam: High-resolution class activation
  mapping,'' in {\em ICCV}, pp.~213--223, 2019.

\bibitem{Muhammad2020EigenCAM}
Muhammad and authors, ``Eigen-cam: Class activation mapping using principal
  components,'' {\em IEEE Sign. Process. Letters}, vol.~27, no.~8,
  pp.~1344--1348, 2020.

\bibitem{poppi2021revisiting}
S.~Poppi, M.~Cornia, L.~Baraldi, and R.~Cucchiara, ``Revisiting the evaluation
  of class activation mapping for explainability: A novel metric and
  experimental analysis,'' in {\em Proceedings of the IEEE/CVF Conference on
  Computer Vision and Pattern Recognition}, pp.~2299--2304, 2021.

\bibitem{ijcai2020p726}
A.~Ignatiev, ``Towards trustable explainable ai,'' in {\em Proceedings of the
  Twenty-Ninth International Joint Conference on Artificial Intelligence,
  {IJCAI-20}}, pp.~5154--5158, 7 2020.

\bibitem{khakzar2020rethinking}
A.~Khakzar, S.~Baselizadeh, and N.~Navab, ``Rethinking positive aggregation and
  propagation of gradients in gradient-based saliency methods,'' {\em arXiv
  preprint arXiv:2012.00362}, 2020.

\bibitem{andreella2023procrustes}
A.~Andreella, R.~De~Santis, A.~Vesely, and L.~Finos, ``Procrustes-based
  distances for exploring between-matrices similarity,'' {\em Statistical
  Methods \& Applications}, vol.~32, no.~3, pp.~867--882, 2023.

\bibitem{Sanjoy2022}
S.~Dasgupta, N.~Frost, and M.~Moshkovitz, ``Framework for evaluating
  faithfulness of local explanations,'' in {\em International Conference on
  Machine Learning}, pp.~4794--4815, PMLR, 2022.

\bibitem{Alvarez2018Stability}
D.~Alvarez-Melis and T.~Jaakkola, ``On the robustness of interpretability
  methods,'' in {\em Proceedings of the 32nd Conference on Neural Information
  Processing Systems (NeurIPS)}, pp.~9172--9183, 2018.

\bibitem{agarwal2022rethinking}
C.~Agarwal, N.~Johnson, M.~Pawelczyk, S.~Krishna, E.~Saxena, M.~Zitnik, and
  H.~Lakkaraju, ``Rethinking stability for attribution-based explanations,''
  {\em arXiv preprint arXiv:2203.06877}, 2022.

\bibitem{Webber2010RBO}
W.~Webber, A.~Moffat, and J.~Zobel, ``A similarity measure for indefinite
  rankings,'' {\em ACM Transactions on Information Systems (TOIS)}, vol.~28,
  no.~4, pp.~1--38, 2010.

\bibitem{selvaraju2017grad}
R.~R. Selvaraju, M.~Cogswell, A.~Das, R.~Vedantam, D.~Parikh, and D.~Batra,
  ``Grad-cam: Visual explanations from deep networks via gradient-based
  localization,'' in {\em Proceedings of the IEEE international conference on
  computer vision (ICCV)}, pp.~618--626, 2017.

\bibitem{montavon2017explaining}
G.~Montavon, S.~Lapuschkin, A.~Binder, W.~Samek, and K.-R. M{\"u}ller,
  ``Explaining nonlinear classification decisions with deep taylor
  decomposition,'' in {\em Pattern Recognition}, pp.~87--106, Springer, 2017.

\bibitem{bach2015pixel}
S.~Bach, A.~Binder, G.~Montavon, F.~Klauschen, K.-R. M{\"u}ller, and W.~Samek,
  ``Pixel-wise explanations for non-linear classifier decisions by layer-wise
  relevance propagation,'' in {\em Proceedings of the IEEE Conference on
  Computer Vision and Pattern Recognition}, pp.~654--662, 2015.

\bibitem{samek2017evaluating}
W.~Samek, T.~Wiegand, and K.-R. M{\"u}ller, ``Evaluating the visualization of
  what a deep neural network has learned,'' {\em IEEE Transactions on Neural
  Networks and Learning Systems}, vol.~28, no.~11, pp.~2660--2673, 2017.

\bibitem{arras2017relevant}
L.~Arras, F.~Horn, G.~Montavon, K.-R. M{\"u}ller, and W.~Samek, ``{What is
  relevant in a text document?}: Learning relevance in document
  representations,'' in {\em Conference on Computer Vision and Pattern
  Recognition Workshops}, pp.~1695--1704, 2017.

\bibitem{yeh2019infidelity}
C.-K. Yeh, C.-Y. Hsieh, A.~Suggala, D.~I. Inouye, and P.~K. Ravikumar, ``On the
  (in)fidelity and sensitivity of explanations,'' in {\em Advances in Neural
  Information Processing Systems}, vol.~32, 2019.

\bibitem{adebayo2018sanity}
J.~Adebayo, J.~Gilmer, M.~Muelly, I.~Goodfellow, M.~Hardt, and B.~Kim, ``Sanity
  checks for saliency maps,'' in {\em Advances in Neural Information Processing
  Systems (NeurIPS)}, vol.~31, 2018.

\bibitem{fel2021algorithmic}
T.~Fel, D.~Vigouroux, R.~Cad{\`e}ne, and T.~Serre, ``How good is your
  explanation? algorithmic stability measures to assess the quality of
  explanations for deep neural networks,'' in {\em Proceedings of the IEEE/CVF
  Conference on Computer Vision and Pattern Recognition}, pp.~10880--10889,
  2021.

\bibitem{hooker2019benchmark}
S.~Hooker, D.~Erhan, P.-J. Kindermans, and B.~Kim, ``A benchmark for
  interpretability methods in deep neural networks,'' in {\em Advances in
  Neural Information Processing Systems (NeurIPS)}, vol.~32, 2019.

\bibitem{samek2021explaining}
W.~Samek, G.~Montavon, S.~Lapuschkin, C.~Anders, and K.-R. Müller,
  ``Explaining deep neural networks and beyond: A review of methods and
  applications,'' {\em Proceedings of the IEEE}, vol.~109, no.~3, pp.~247--278,
  2021.

\bibitem{bansal2020sam}
N.~Bansal, C.~Agarwal, and A.~Nguyen, ``Sam: the sensitivity of attribution
  methods to hyperparameters. in 2020 ieee,'' in {\em CVF Conference on
  Computer Vision and Pattern Recognition Workshops (CVPRW)}, pp.~11--21, 2020.

\bibitem{nourelahi2022explainable}
M.~Nourelahi, L.~Kotthoff, P.~Chen, and A.~Nguyen, ``How explainable are
  adversarially-robust cnns?,'' {\em arXiv preprint arXiv:2205.13042}, 2022.

\bibitem{NEURIPS2018_294a8ed2}
J.~Adebayo, J.~Gilmer, M.~Muelly, I.~Goodfellow, M.~Hardt, and B.~Kim, ``Sanity
  checks for saliency maps,'' in {\em Advances in Neural Information Processing
  Systems} (S.~Bengio, H.~Wallach, H.~Larochelle, K.~Grauman, N.~Cesa-Bianchi,
  and R.~Garnett, eds.), vol.~31, Curran Associates, Inc., 2018.

\bibitem{yeh2019fidelity}
C.-K. Yeh, C.-J. Hsieh, A.~Suggala, D.~Inouye, and P.~Ravikumar, ``On the
  (in)fidelity and sensitivity of explanations,'' in {\em Advances in Neural
  Information Processing Systems (NeurIPS)}, vol.~32, 2019.

\bibitem{He2016ResNet}
K.~He, X.~Zhang, S.~Ren, and J.~Sun, ``Deep residual learning for image
  recognition,'' in {\em IEEE Conference on Computer Vision and Pattern
  Recognition}, pp.~770--778, 2016.

\bibitem{Simonyan2014VGG}
K.~Simonyan and A.~Zisserman, ``Very deep convolutional networks for
  large-scale image recognition,'' {\em arXiv preprint}, vol.~arXiv:1409.1556,
  2014.

\bibitem{Szegedy2015Inception}
C.~Szegedy, V.~Vanhoucke, S.~Ioffe, J.~Shlens, and Z.~Wojna, ``Rethinking the
  inception architecture for computer vision,'' in {\em IEEE Conference on
  Computer Vision and Pattern Recognition}, pp.~2818--2826, 2016.

\bibitem{dosovitskiy2020image}
A.~Dosovitskiy, L.~Beyer, A.~Kolesnikov, D.~Weissenborn, X.~Zhai,
  T.~Unterthiner, M.~Dehghani, M.~Minderer, G.~Heigold, S.~Gelly, J.~Uszkoreit,
  and N.~Houlsby, ``An image is worth 16x16 words: Transformers for image
  recognition at scale,'' in {\em International Conference on Learning
  Representations (ICLR)}, 2021.
\newblock arXiv:2010.11929.

\bibitem{dvc}
J.~Elson, J.~J. Douceur, J.~Howell, and J.~Saul, ``Asirra: A captcha that
  exploits interest-aligned manual image categorization,'' in {\em Proceedings
  of 14th ACM Conference on Computer and Communications Security (CCS)},
  Association for Computing Machinery, Inc., October 2007.

\bibitem{ILSVRC15}
O.~Russakovsky, J.~Deng, H.~Su, J.~Krause, S.~Satheesh, S.~Ma, Z.~Huang,
  A.~Karpathy, A.~Khosla, M.~Bernstein, A.~C. Berg, and L.~Fei-Fei, ``{ImageNet
  Large Scale Visual Recognition Challenge},'' {\em International Journal of
  Computer Vision (IJCV)}, vol.~115, no.~3, pp.~211--252, 2015.

\bibitem{parkhi12a}
O.~M. Parkhi, A.~Vedaldi, A.~Zisserman, and C.~V. Jawahar, ``Cats and dogs,''
  in {\em IEEE Conference on Computer Vision and Pattern Recognition}, 2012.

\bibitem{tschandl2018ham10000}
P.~Tschandl, C.~Rosendahl, and H.~Kittler, ``The ham10000 dataset: A large
  collection of multi-source dermatoscopic images of common pigmented skin
  lesions,'' {\em Scientific data}, vol.~5, no.~1, pp.~1--9, 2018.

\bibitem{fei2004learning}
L.~Fei-Fei, R.~Fergus, and P.~Perona, ``Learning generative visual models from
  few training examples: An incremental bayesian approach tested on 101 object
  categories,'' in {\em 2004 Conference on Computer Vision and Pattern
  Recognition Workshop}, pp.~178--178, IEEE, 2004.

\bibitem{vedaldi2008quick}
A.~Vedaldi and S.~Soatto, ``Quick shift and kernel methods for mode seeking,''
  in {\em European Conference on Computer Vision}, pp.~705--718, Springer,
  2008.

\bibitem{rbo}
W.~Webber, A.~Moffat, and J.~Zobel, ``A similarity measure for indefinite
  rankings,'' {\em ACM Trans. Inf. Syst.}, vol.~28, Nov. 2010.

\bibitem{Dosovitskiy2020ViT}
A.~Dosovitskiy, L.~Beyer, A.~Kolesnikov, D.~Weissenborn, X.~Zhai,
  T.~Unterthiner, M.~Dehghani, M.~Minderer, G.~Heigold, S.~Gelly, J.~Uszkoreit,
  and N.~Houlsby, ``An image is worth 16x16 words: Transformers for image
  recognition at scale,'' in {\em International Conference on Learning
  Representations (ICLR)}, 2021.

\bibitem{choi2024icev2}
H.~Choi, S.~Jin, and K.~Han, ``Icev2: Interpretability, comprehensiveness, and
  explainability in vision transformer,'' {\em International Journal of
  Computer Vision}, pp.~1--18, 2024.

\bibitem{Kou1995JPEG}
W.~Kou, ``{JPEG Compression Standard},'' in {\em Digital Image Compression},
  vol.~333 of {\em The Springer International Series in Engineering and
  Computer Science}, Springer, Boston, MA, 1995.

\bibitem{Yitzhaky1999direct}
Y.~Yitzhaky and N.~S. Kopeika, ``{Comparison of direct blind deconvolution
  methods for motion-blurred images},'' {\em Applied Optics}, vol.~38, no.~20,
  pp.~4325--4332, 1999.

\bibitem{Goodfellow2015FGSM}
I.~J. Goodfellow, J.~Shlens, and C.~Szegedy, ``{Explaining and Harnessing
  Adversarial Examples},'' {\em arXiv preprint arXiv:1412.6572}, 2015.

\bibitem{Madry2018PGD}
A.~Madry, A.~Makelov, L.~Schmidt, D.~Tsipras, and A.~Vladu, ``{Towards Deep
  Learning Models Resistant to Adversarial Attacks},'' {\em arXiv preprint
  arXiv:1706.06083}, 2018.

\bibitem{Carlini2017CW}
N.~Carlini and D.~Wagner, ``{Towards evaluating the robustness of neural
  networks},'' {\em 2017 IEEE Symposium on Security and Privacy (SP)},
  pp.~39--57, 2017.

\bibitem{achanta2012slic}
R.~Achanta, A.~Shaji, K.~Smith, A.~Lucchi, P.~Fua, and S.~S{\"u}sstrunk, ``Slic
  superpixels compared to state-of-the-art superpixel methods,'' {\em IEEE
  Transactions on Pattern Analysis and Machine Intelligence}, vol.~34, no.~11,
  pp.~2274--2282, 2012.

\bibitem{felzenszwalb2004efficient}
P.~F. Felzenszwalb and D.~P. Huttenlocher, ``Efficient graph-based image
  segmentation,'' {\em International Journal of Computer Vision}, vol.~59,
  no.~2, pp.~167--181, 2004.

\bibitem{kendall1938tau}
M.~G. Kendall, ``A new measure of rank correlation,'' {\em Biometrika},
  vol.~30, no.~1/2, pp.~81--93, 1938.

\bibitem{spearman1904proof}
C.~Spearman, ``The proof and measurement of association between two things,''
  {\em The American Journal of Psychology}, vol.~15, no.~1, pp.~72--101, 1904.

\bibitem{kendall1939}
M.~G. Kendall and B.~B. Smith, ``The problem of m rankings,'' {\em The Annals
  of Mathematical Statistics}, vol.~10, no.~3, pp.~275--287, 1939.

\bibitem{smilkov2017smoothgrad}
D.~Smilkov, N.~Thorat, B.~Kim, F.~Vi{\'e}gas, and M.~Wattenberg, ``Smoothgrad:
  removing noise by adding noise,'' in {\em Workshop on Visualization for Deep
  Learning, ICML}, 2017.

\bibitem{hegde2022probabilistic}
H.~Hegde, S.~Kejriwal, and J.~Pujara, ``Probabilistic pixel attribution:
  Interpreting deep models with statistical inference,'' in {\em Proceedings of
  the IEEE/CVF Conference on Computer Vision and Pattern Recognition (CVPR)},
  pp.~10899--10908, 2022.

\bibitem{madry2018towards}
A.~Madry, A.~Makelov, L.~Schmidt, D.~Tsipras, and A.~Vladu, ``Towards deep
  learning models resistant to adversarial attacks,'' in {\em International
  Conference on Learning Representations (ICLR)}, 2018.

\end{thebibliography}

\clearpage
\newpage
\appendix
\section{Appendix Section}
\subsection{Why is the RBO Better than $l_1$ Norm for Measuring Changes in CAM Outputs}
\label{app1}
\begin{figure*}[htbp]
    \centering
    \includegraphics[width=0.6\textwidth]{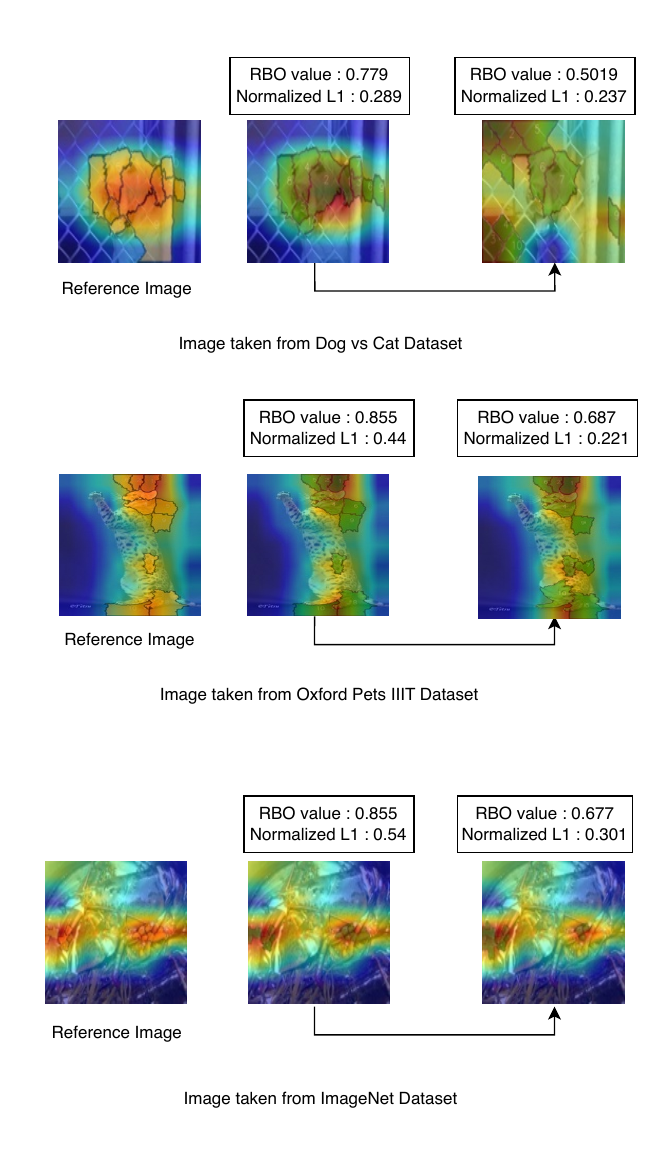}
    \caption{\textit{Comparison of Rank-Biased Overlap (RBO) and normalized $l_1$ norm for measuring changes in CAM outputs under different noise levels.} Examples from Dogs vs. Cats, Oxford IIIT Pets, and ImageNet datasets show that RBO values decrease significantly when the CAM output changes, accurately reflecting differences in segment ranking. In contrast, $l_1$ norm values do not always increase with significant changes and can even decrease, indicating inconsistencies in measuring CAM intensity changes. This highlights the superiority of RBO in capturing meaningful variations in heatmap structures.}
    \label{fig:L1vsRBO}
\end{figure*}

In this section, we provide a detailed comparison of Rank-Biased Overlap (RBO) and the $l_1$ distances for evaluating changes in CAM outputs. Figure~\ref{fig:L1vsRBO} illustrates this comparison across three datasets: ImageNet, Dogs vs. Cats, and Oxford IIIT Pets. When assessing changes in CAM outputs, the RBO metric focuses on the ranking similarity of segments, while the $l_1$ distance is the pixel-wise differences. The following examples demonstrate the effectiveness of RBO in capturing significant changes, whereas the $l_1$ distance can sometimes produce misleading results.

For a random image taken from the Dogs vs. Cats dataset, under minimal changes in the CAM outputs, the RBO value is 0.779, and the normalized $l_1$ is 0.289. In the case of a significant change, the RBO value drops to 0.5019, indicating substantial reordering of the segments. However, the normalized $l_1$ distance decreases to 0.237 instead of increasing, failing to capture the extent of the change, thus showing an instability.

For another random image from the Oxford IIIT Pets dataset, with minimal changes, the RBO value is 0.855, and the normalized $l_1$ distance is 0.44. For a minimal visual change in the CAM map, the RBO value decreases to 0.687, reflecting the rank changes. In contrast, the normalized $l_1$  reduces to 0.221, providing an inconsistent measure of the change's magnitude.
\newpage

Random image from the ImageNet dataset is taken, with minor CAM variations, the RBO value is 0.855, and the normalized $l_1$ distance is 0.54. When the CAM outputs change significantly, the RBO value drops to 0.677, showing a clear deviation, but the normalized $l_1$ values decrease to 0.301, suggesting an unreliable interpretation of the visual difference.

Different noise types, such as salt-and-pepper noise, can cause changes in pixel values without significantly affecting the CAM heatmap's structure. For example, even though the $l_1$ distance might increase due to pixel-level perturbations, the overall CAM distribution and ranking may remain unchanged. This inconsistency leads to potential inaccuracies when using the $l_1$ norm-based distance measurement as a metric for deciding the robustness of a particular CAM method. In contrast, RBO remains robust by focusing on the relative ranks of regions within the CAM outputs, ensuring a more reliable assessment of actual changes.

These examples highlight that RBO consistently reflects the extent of changes in CAM maps by analyzing rank-based similarities. In contrast, the $l_1$ norm can produce misleading results, failing to capture meaningful deviations in the heatmap structure. Therefore, RBO is a more suitable metric for evaluating CAM output changes, especially when dealing with various noise types and dataset complexities.

\newpage
\subsection{\Rev{Violin Plot Distribution of RBO Scores Across Models and Datasets}}
\label{appendix:violin-distributions}

\Rev{To provide a comprehensive view of the distribution of CAM robustness scores across models and datasets, we include violin plots in this section. These plots visualize the spread and central tendency of Rank-Biased Overlap (RBO) scores computed for each CAM method under various perturbations.}

\begin{figure}[H]
    \centering
    \includegraphics[height=0.85\textheight]{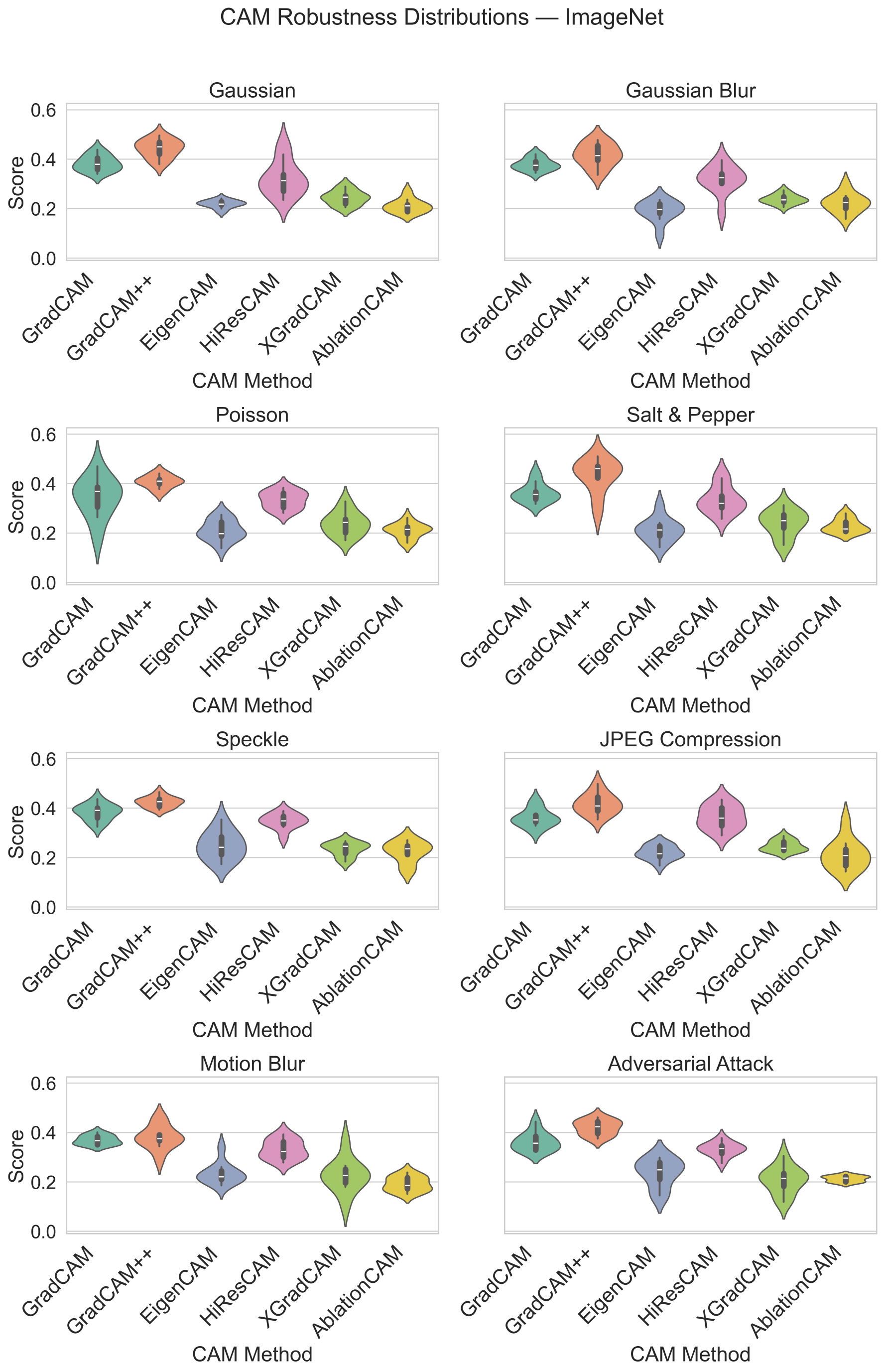}
    \caption{\Rev{RBO distribution across CAM methods on the ImageNet dataset (ResNet-50).}}
    \label{fig:resnet50_violin_imagenet}
\end{figure}

\begin{figure}[H]
    \centering
    \includegraphics[height=0.85\textheight]{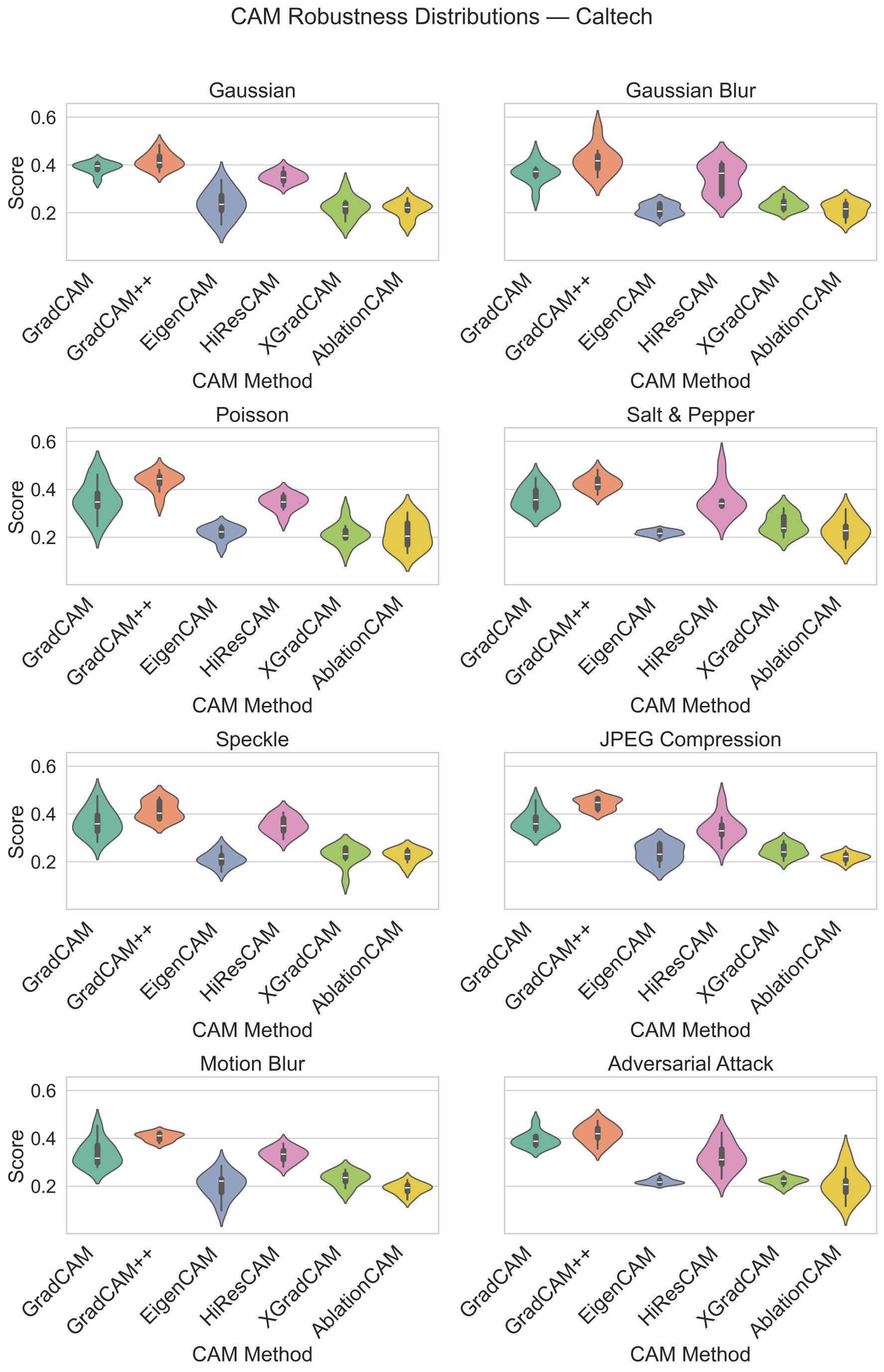}
    \caption{\Rev{RBO distribution across CAM methods on the Caltech dataset (ResNet-50).}}
    \label{fig:resnet50_violin_caltech}
\end{figure}

\begin{figure}[H]
    \centering
    \includegraphics[height=0.85\textheight]{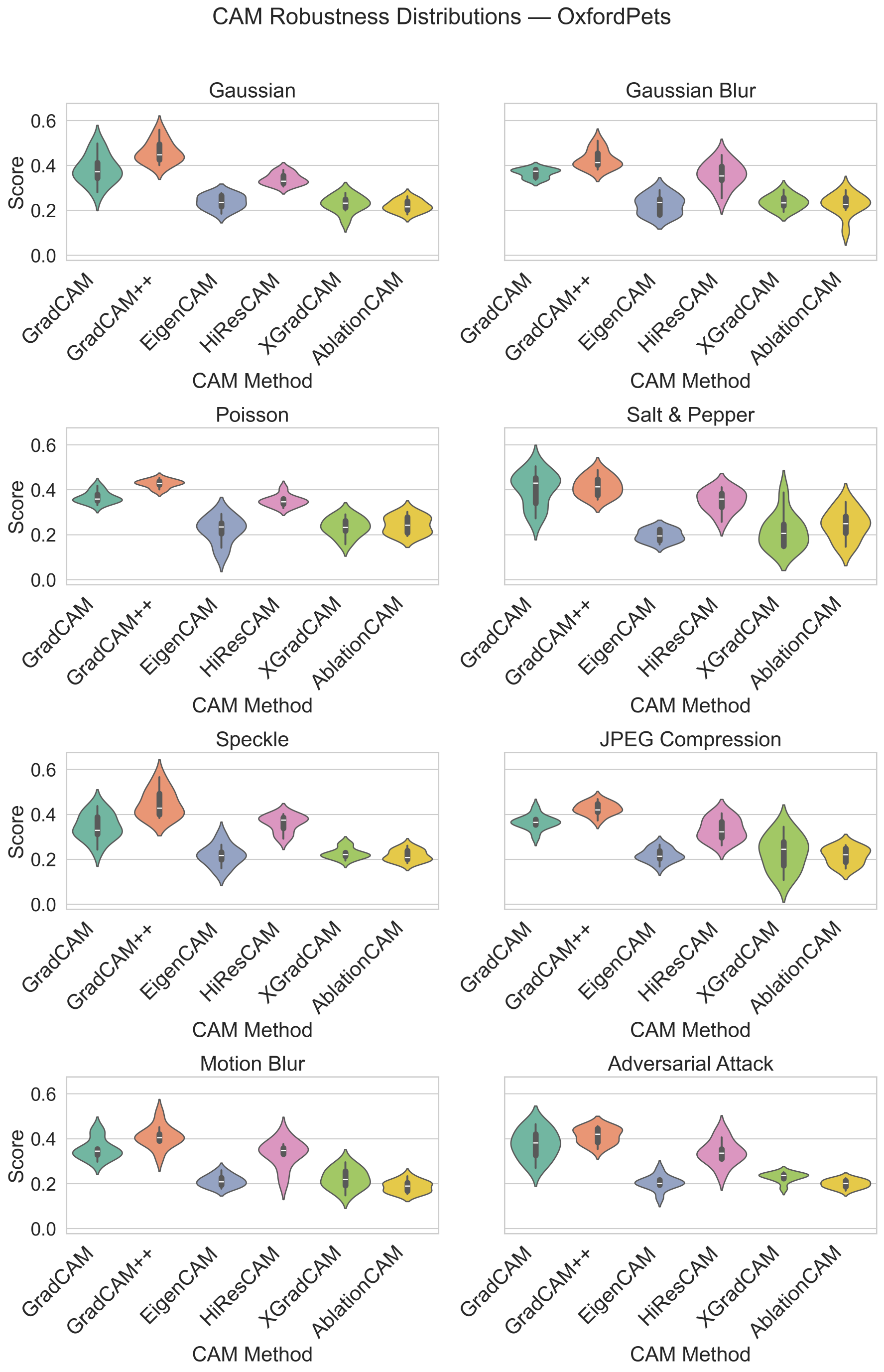}
    \caption{\Rev{RBO distribution across CAM methods on the OxfordPets dataset (ResNet-50).}}
    \label{fig:resnet50_violin_oxfordpets}
\end{figure}

\begin{figure}[H]
    \centering
    \includegraphics[height=0.85\textheight]{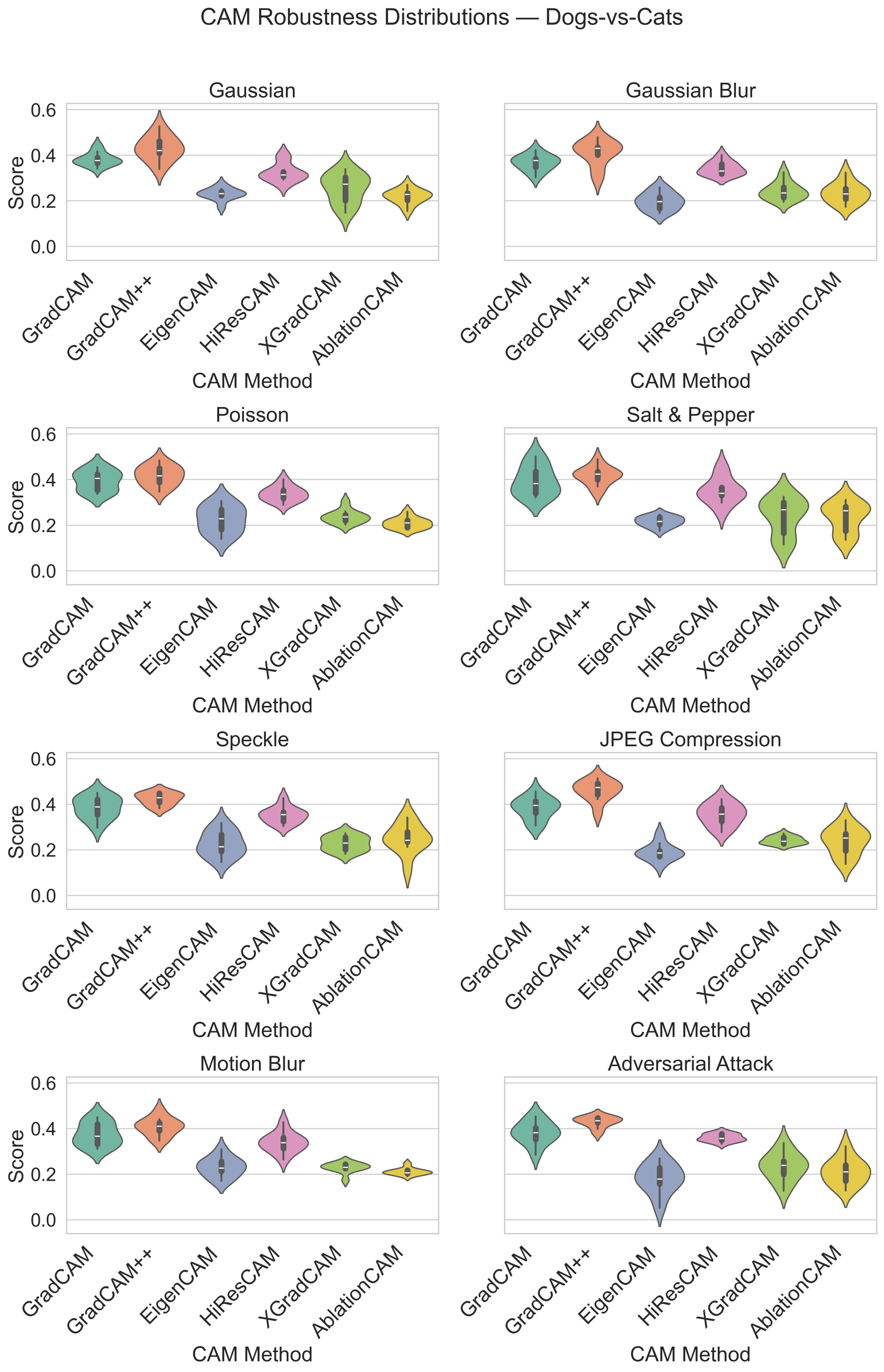}
    \caption{\Rev{RBO distribution across CAM methods on the Dogs-vs-Cats dataset (ResNet-50).}}
    \label{fig:resnet50_violin_dvc}
\end{figure}

\begin{figure}[H]
    \centering
    \includegraphics[height=0.85\textheight]{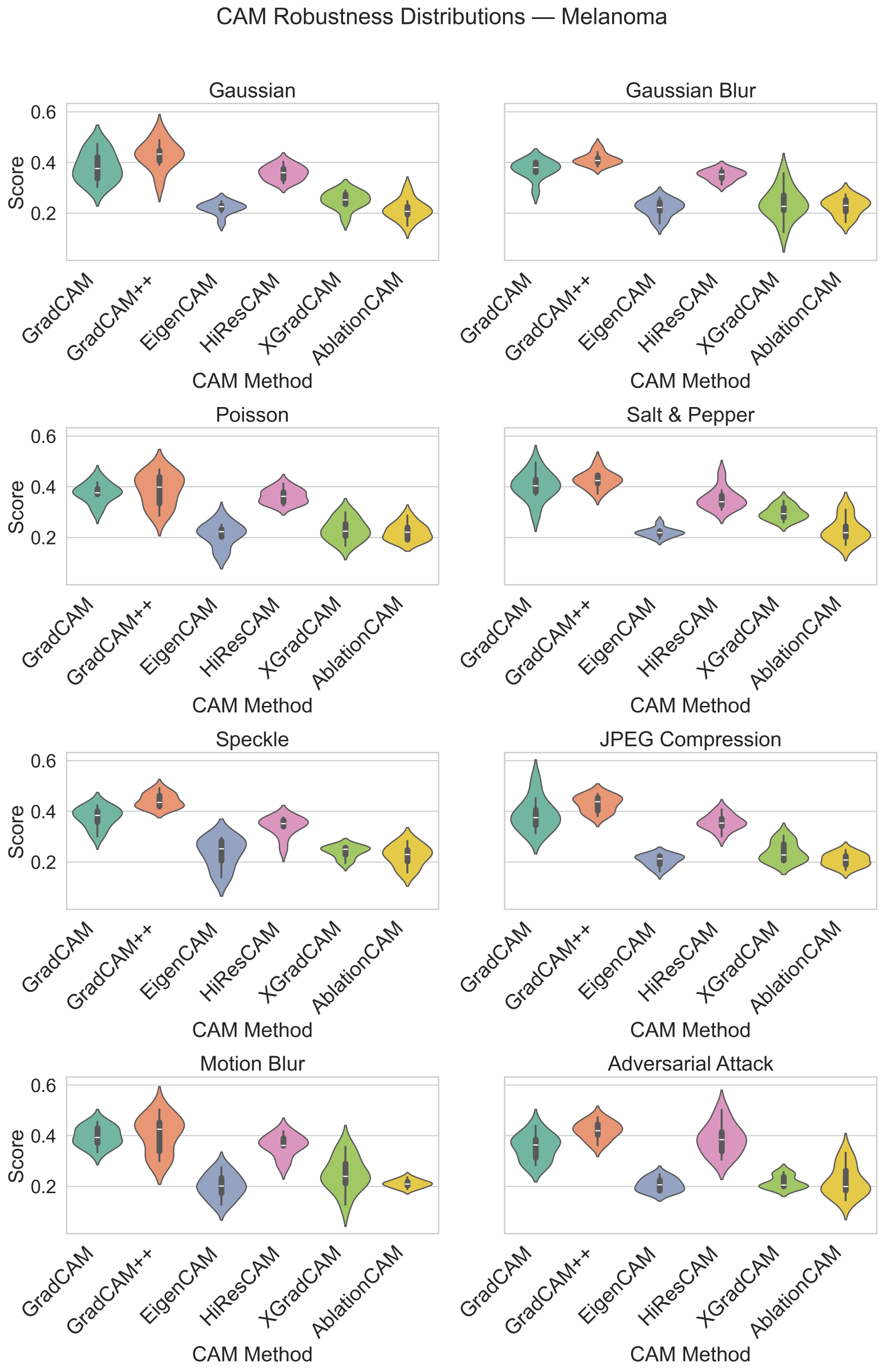}
    \caption{\Rev{RBO distribution across CAM methods on the Melanoma dataset (ResNet-50).}}
    \label{fig:resnet50_violin_melanoma}
\end{figure}

\begin{figure}[H]
    \centering
    \includegraphics[height=0.85\textheight]{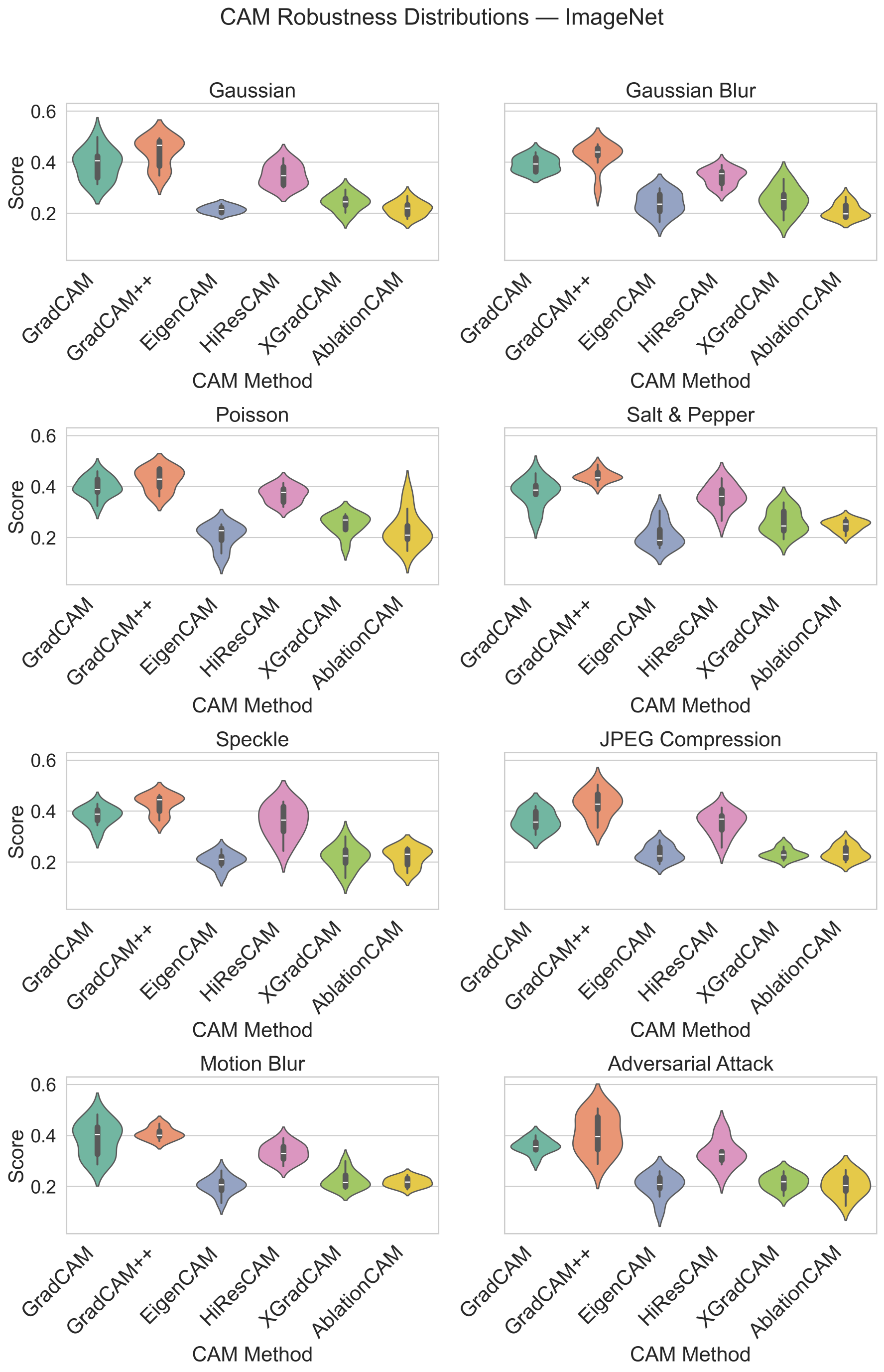}
    \caption{\Rev{RBO distribution across CAM methods for VGG19 on ImageNet.}}
    \label{fig:vgg19_imagenet_violin}
\end{figure}

\begin{figure}[H]
    \centering
    \includegraphics[height=0.85\textheight]{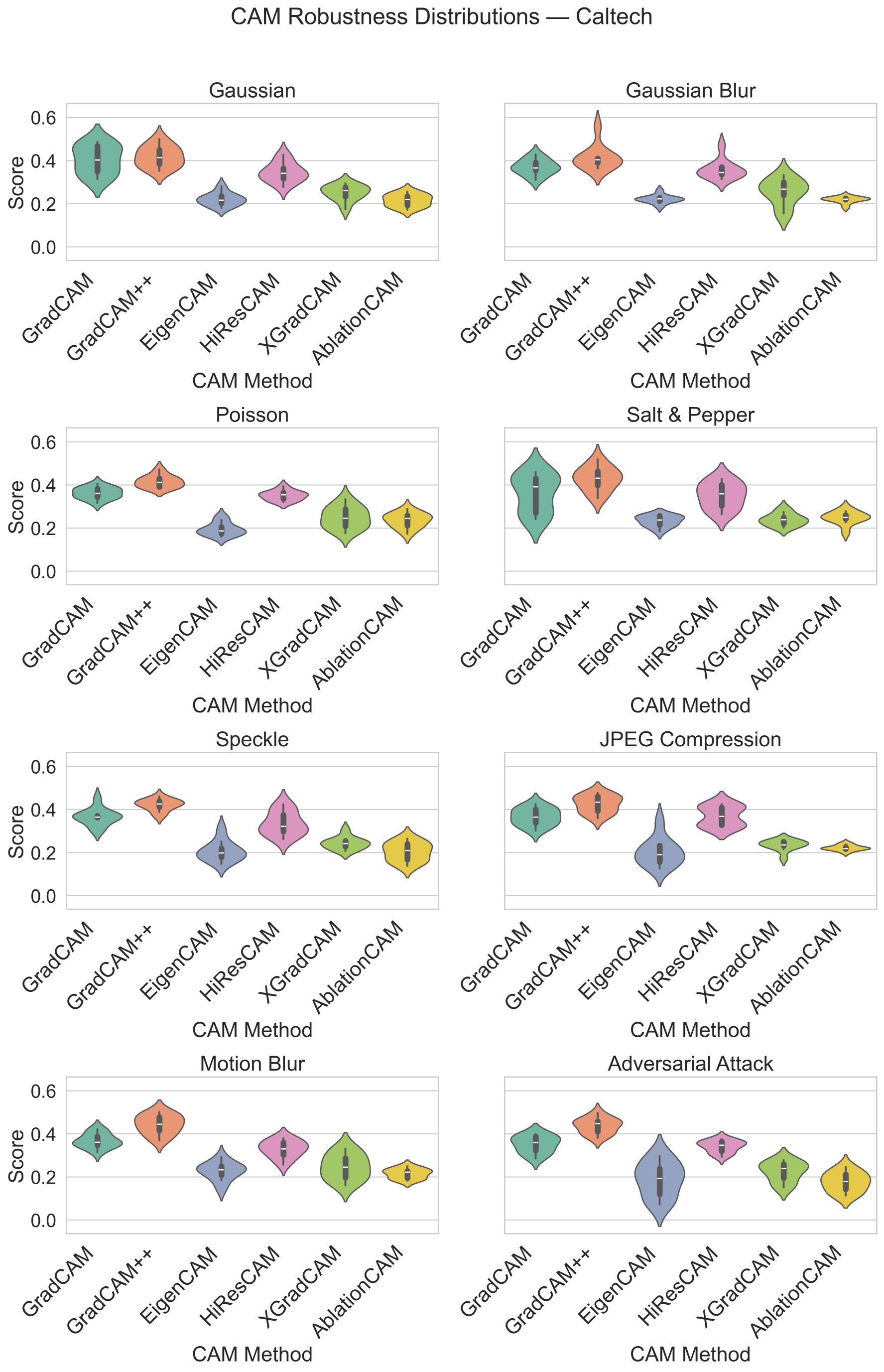}
    \caption{\Rev{RBO distribution across CAM methods for VGG19 on Caltech.}}
    \label{fig:vgg19_caltech_violin}
\end{figure}

\begin{figure}[H]
    \centering
    \includegraphics[height=0.85\textheight]{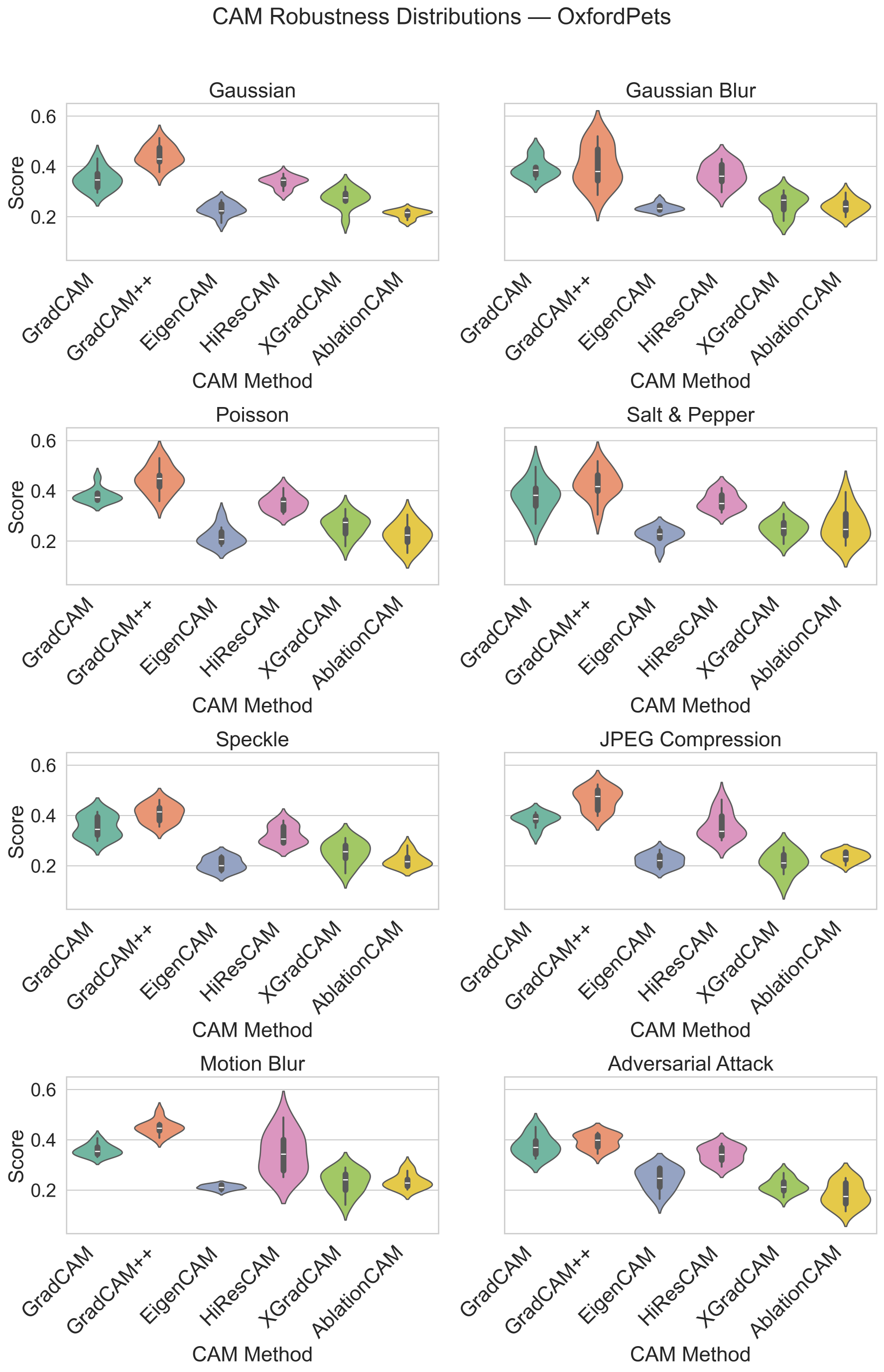}
    \caption{\Rev{RBO distribution across CAM methods for VGG19 on OxfordPets.}}
    \label{fig:vgg19_oxfordpets_violin}
\end{figure}

\begin{figure}[H]
    \centering
    \includegraphics[height=0.85\textheight]{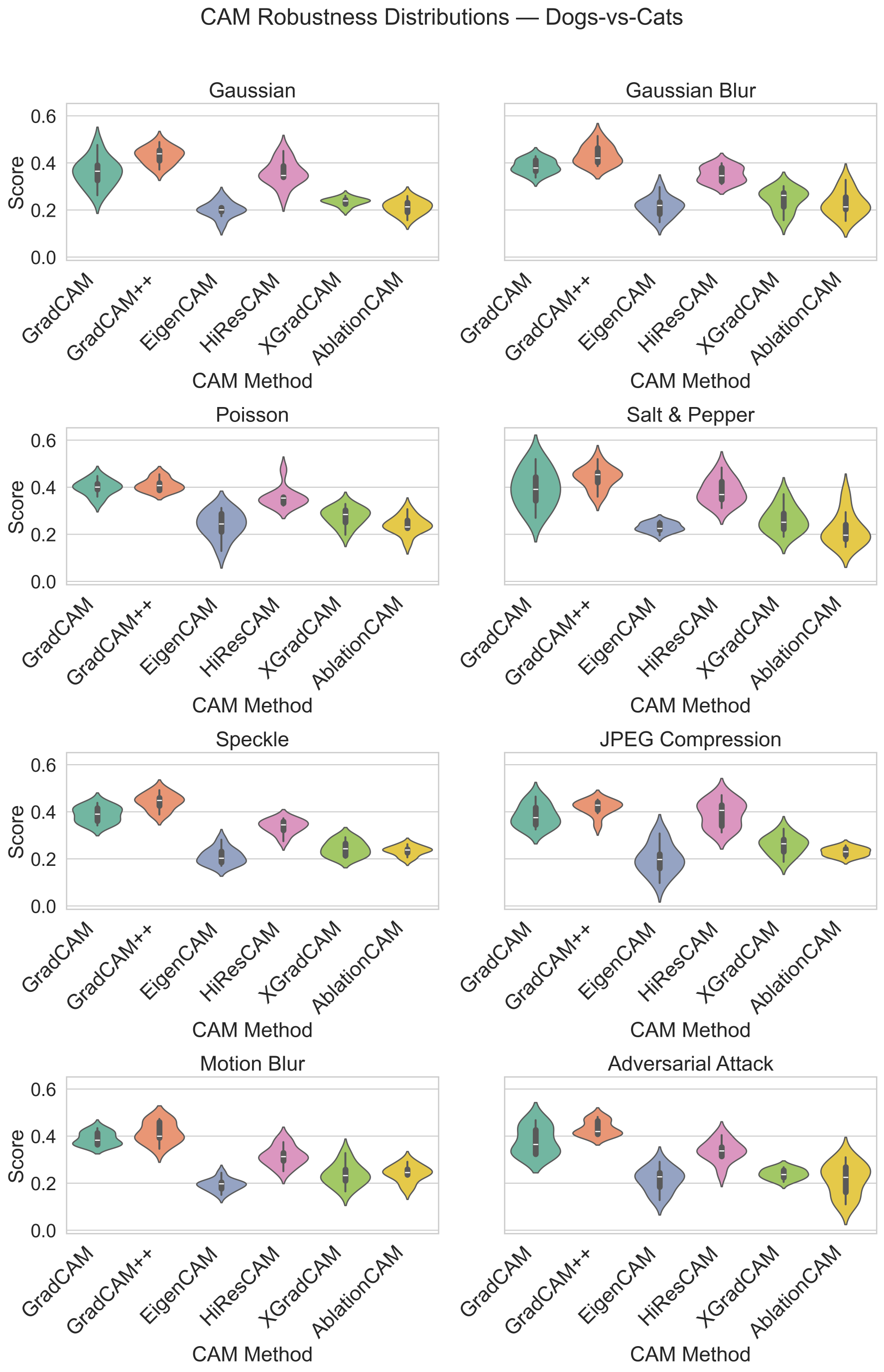}
    \caption{\Rev{RBO distribution across CAM methods for VGG19 on Dogs-vs-Cats.}}
    \label{fig:vgg19_dogsvscats_violin}
\end{figure}

\begin{figure}[H]
    \centering
    \includegraphics[height=0.85\textheight]{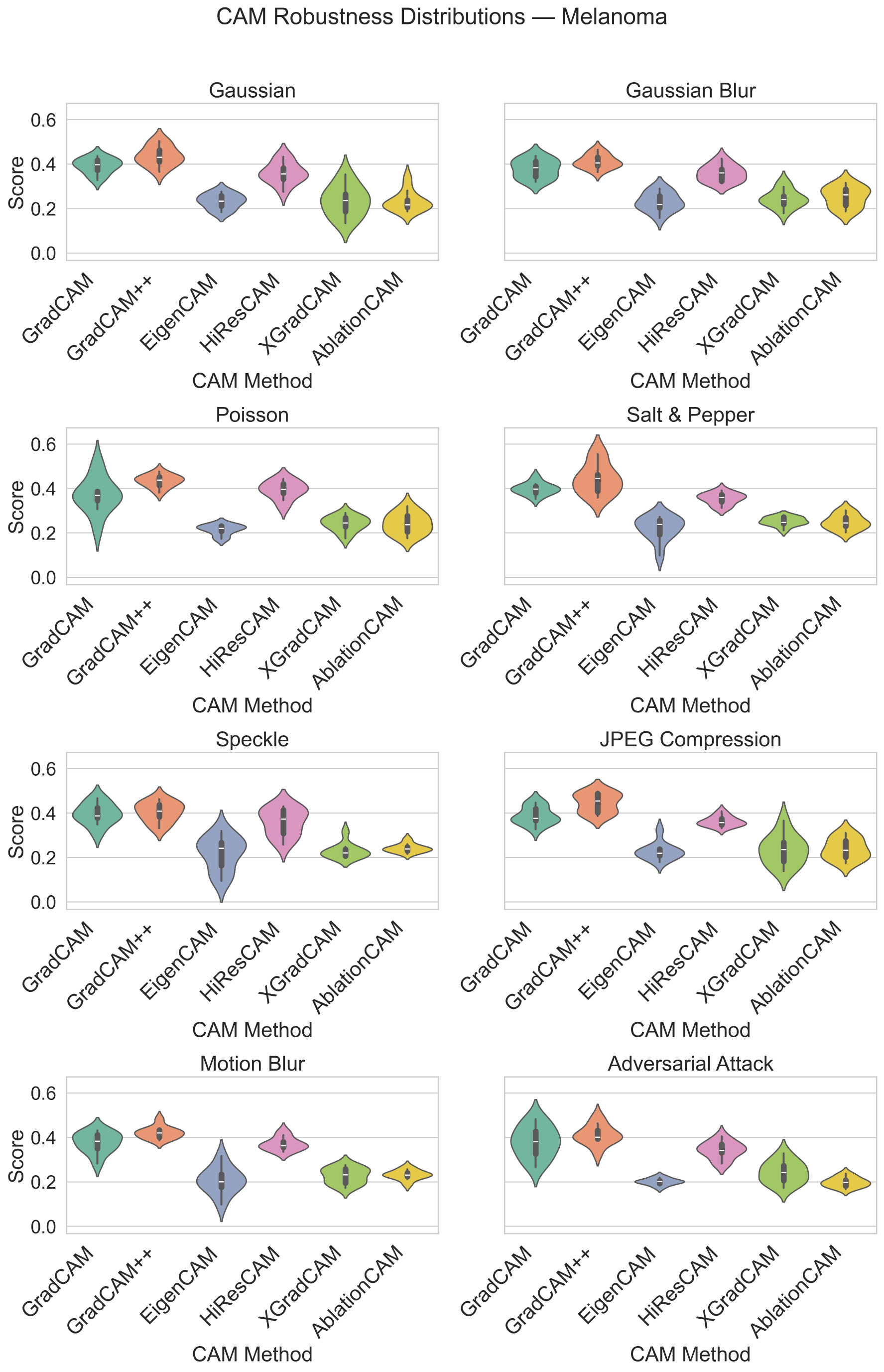}
    \caption{\Rev{RBO distribution across CAM methods for VGG19 on Melanoma.}}
    \label{fig:vgg19_melanoma_violin}
\end{figure}

\begin{figure}[H]
    \centering
    \includegraphics[height=0.85\textheight]{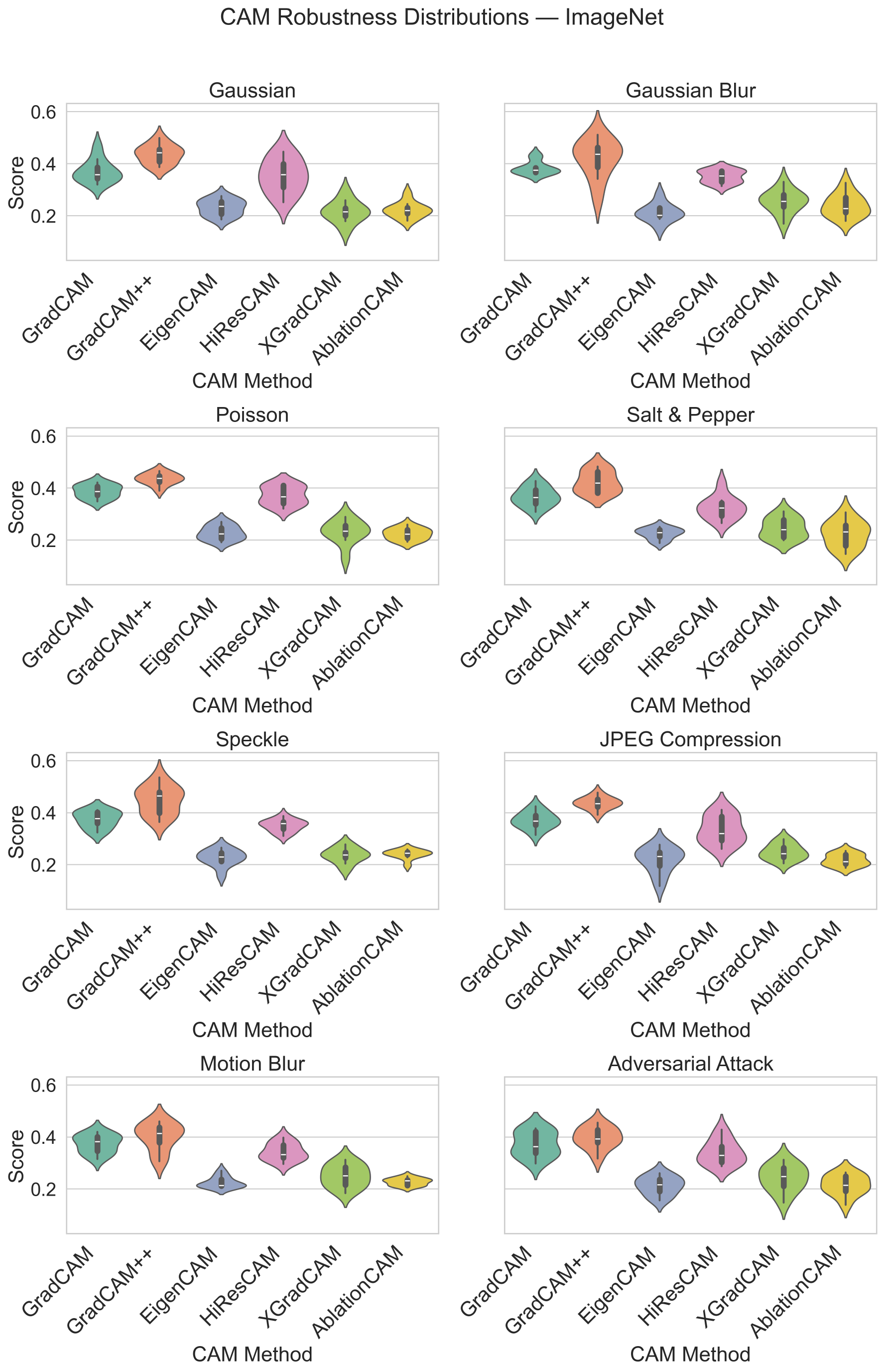}
    \caption{\Rev{RBO distribution across CAM methods for InceptionV3 on the ImageNet dataset.}}
    \label{fig:inception_imagenet_violin}
\end{figure}

\begin{figure}[H]
    \centering
    \includegraphics[height=0.85\textheight]{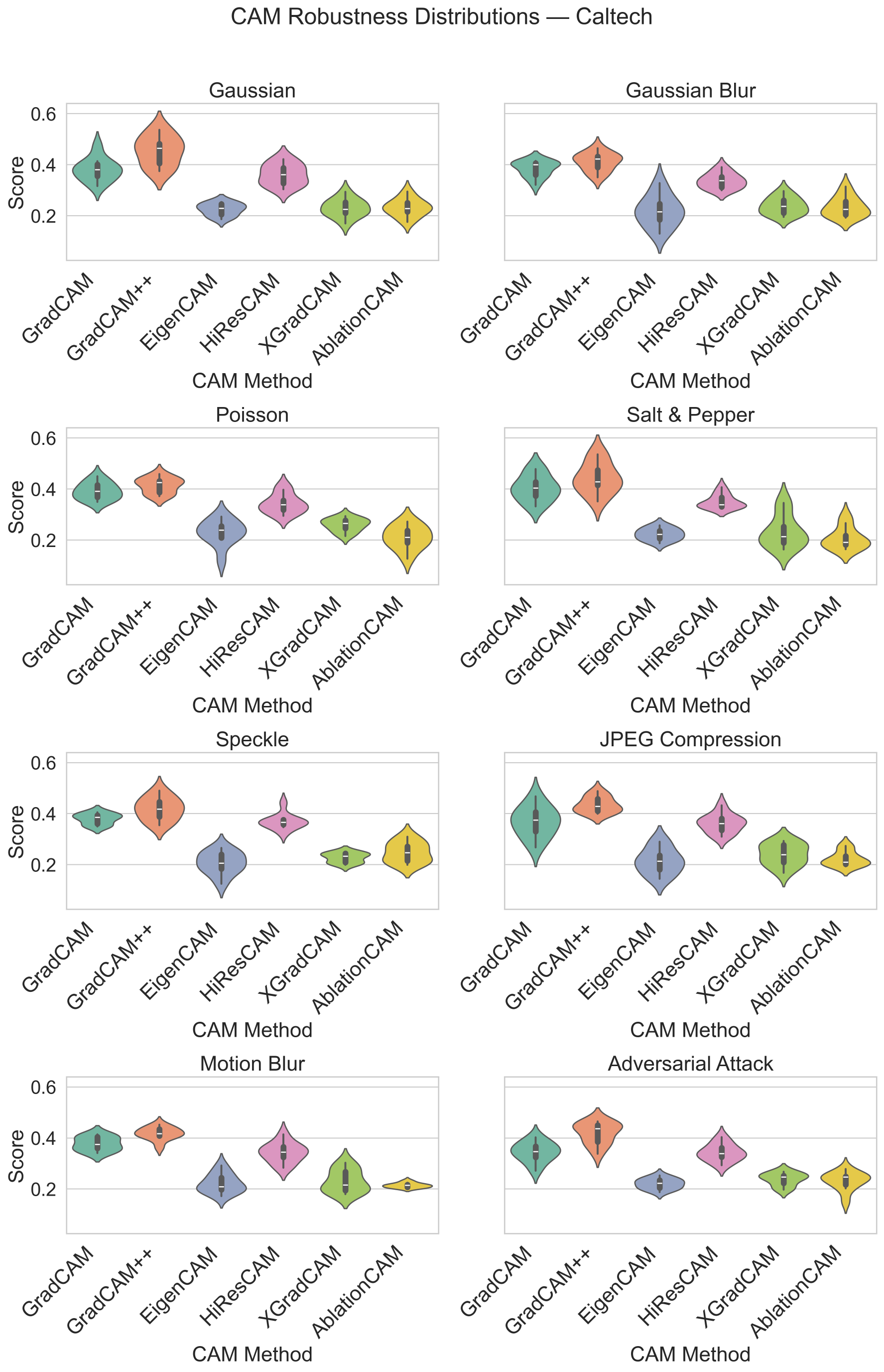}
    \caption{\Rev{RBO distribution across CAM methods for InceptionV3 on the Caltech dataset.}}
    \label{fig:inception_caltech_violin}
\end{figure}

\begin{figure}[H]
    \centering
    \includegraphics[height=0.85\textheight]{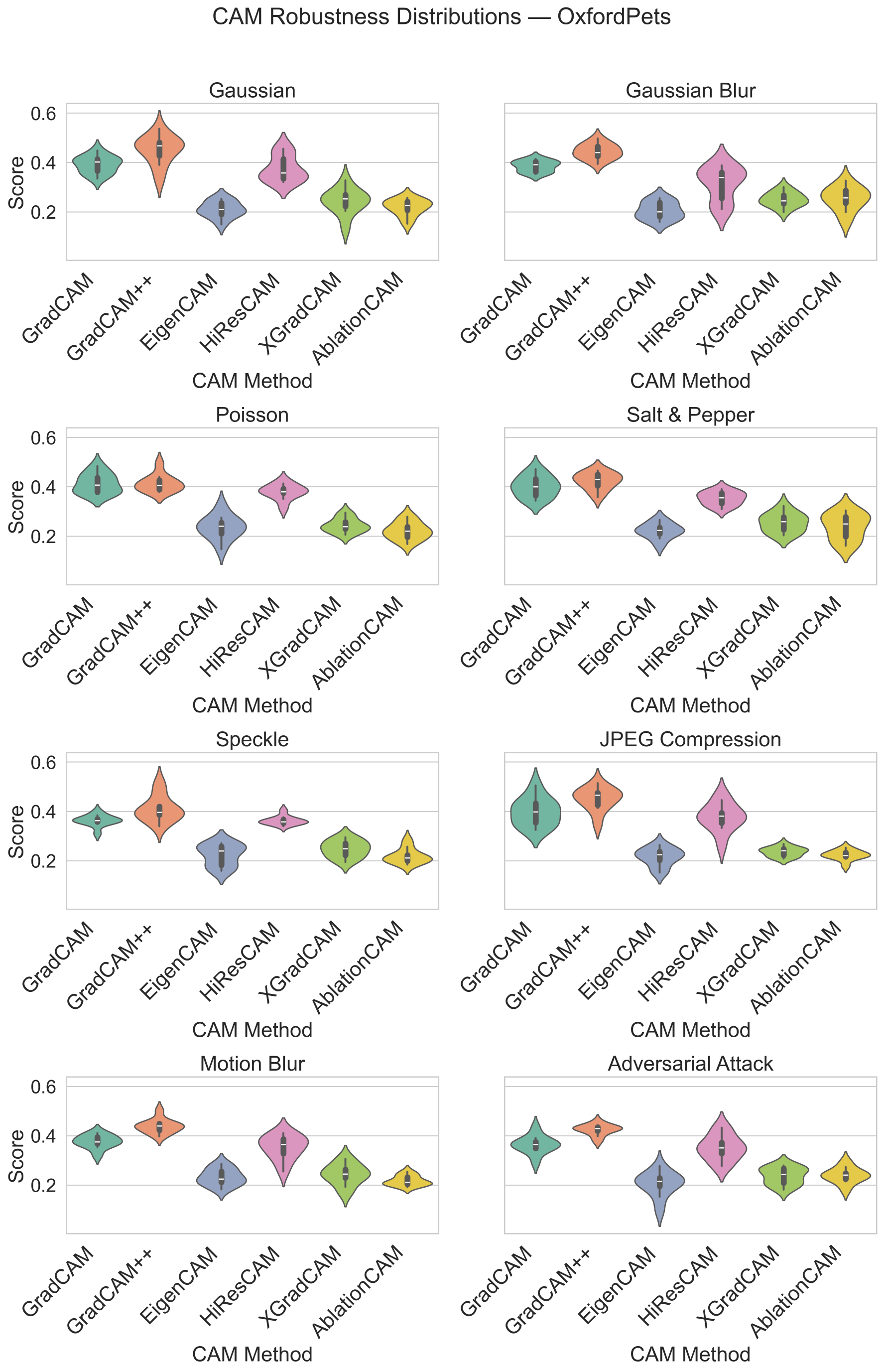}
    \caption{\Rev{RBO distribution across CAM methods for InceptionV3 on the OxfordPets dataset.}}
    \label{fig:inception_oxfordpets_violin}
\end{figure}

\begin{figure}[H]
    \centering
    \includegraphics[height=0.85\textheight]{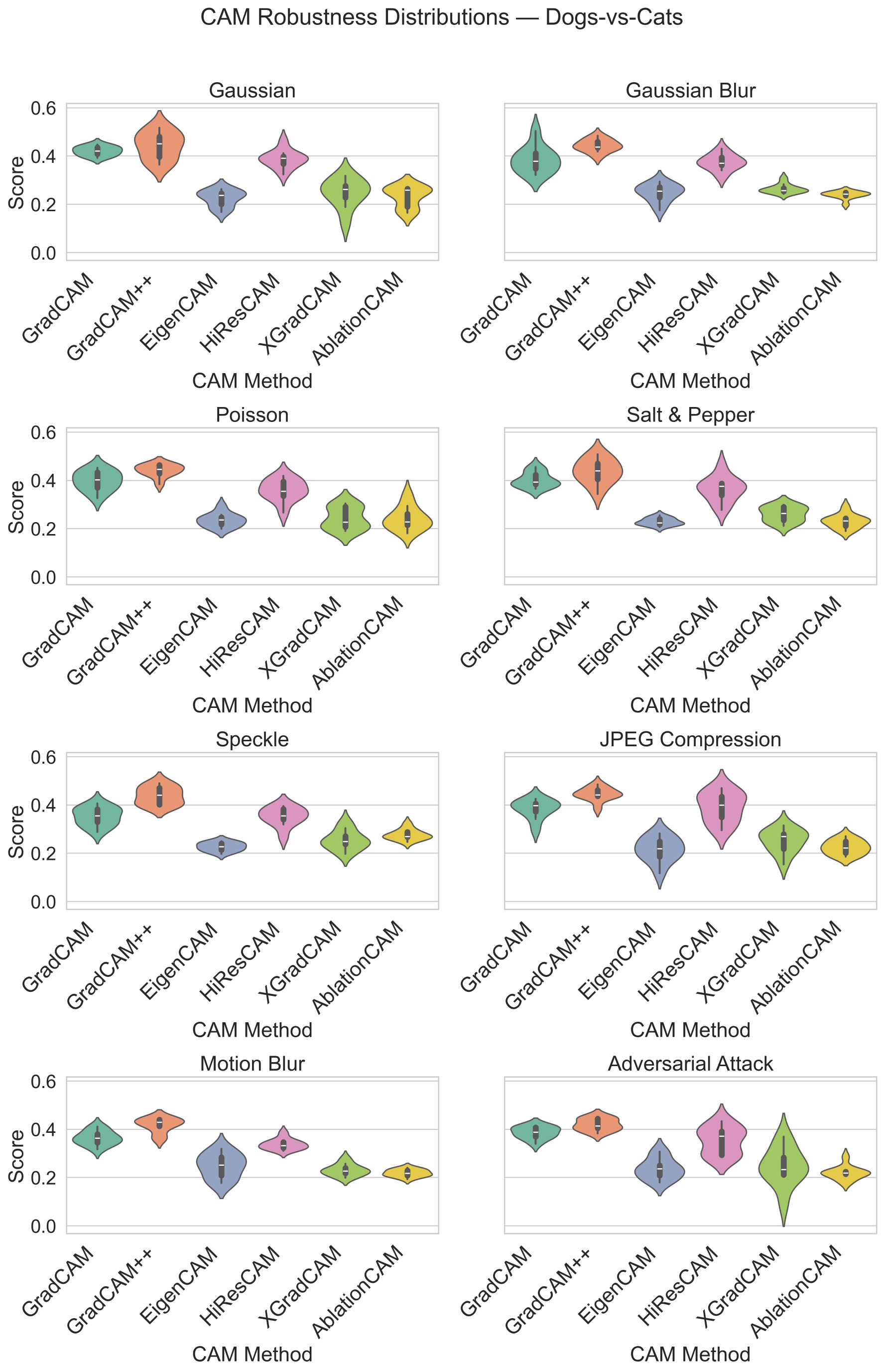}
    \caption{\Rev{RBO distribution across CAM methods for InceptionV3 on the Dogs-vs-Cats dataset.}}
    \label{fig:inception_dogsvscats_violin}
\end{figure}

\begin{figure}[H]
    \centering
    \includegraphics[height=0.85\textheight]{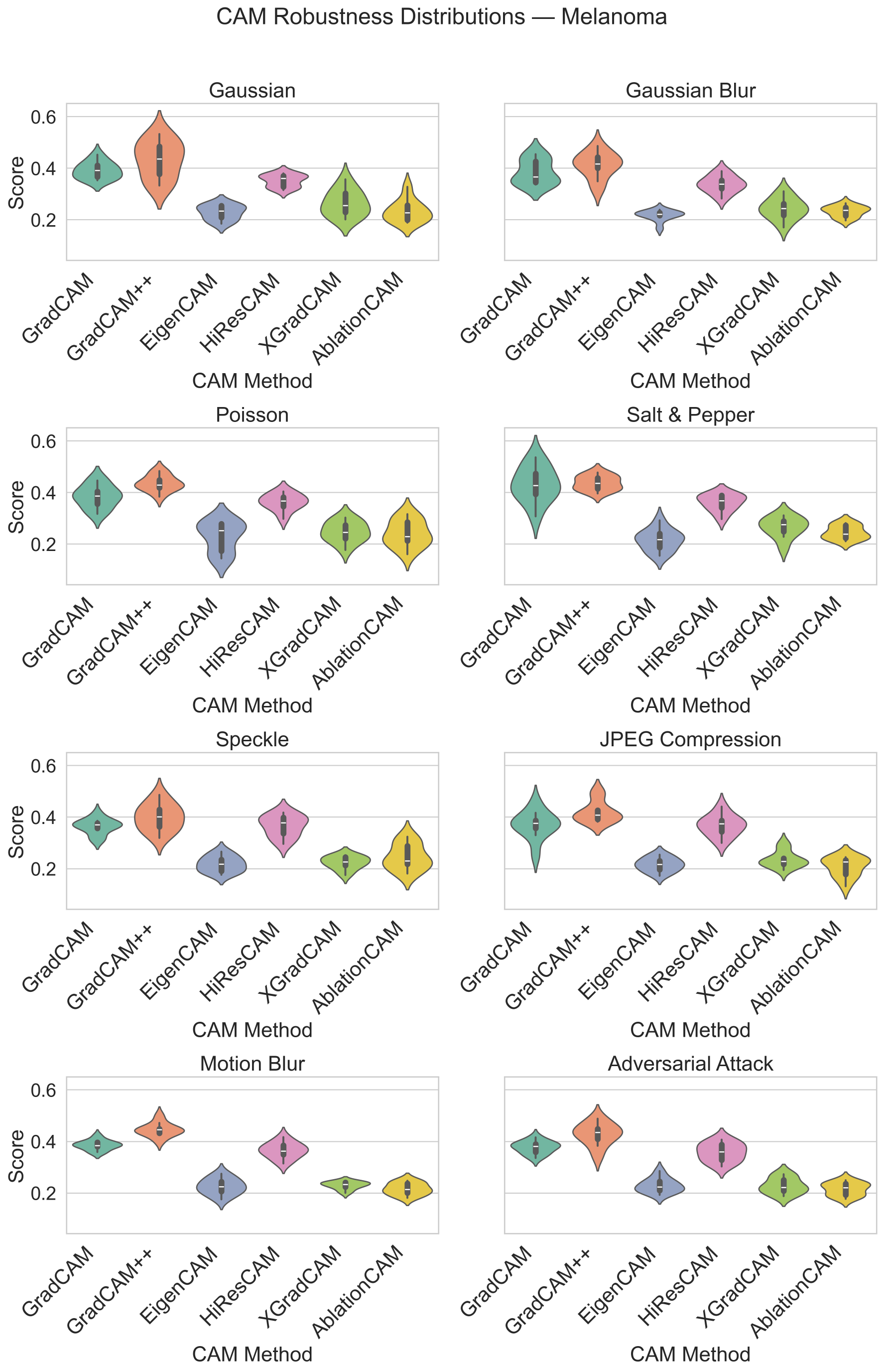}
    \caption{\Rev{RBO distribution across CAM methods for InceptionV3 on the Melanoma dataset.}}
    \label{fig:inception_melanoma_violin}
\end{figure}

\begin{figure}[H]
    \centering
    \includegraphics[width=0.85\textwidth]{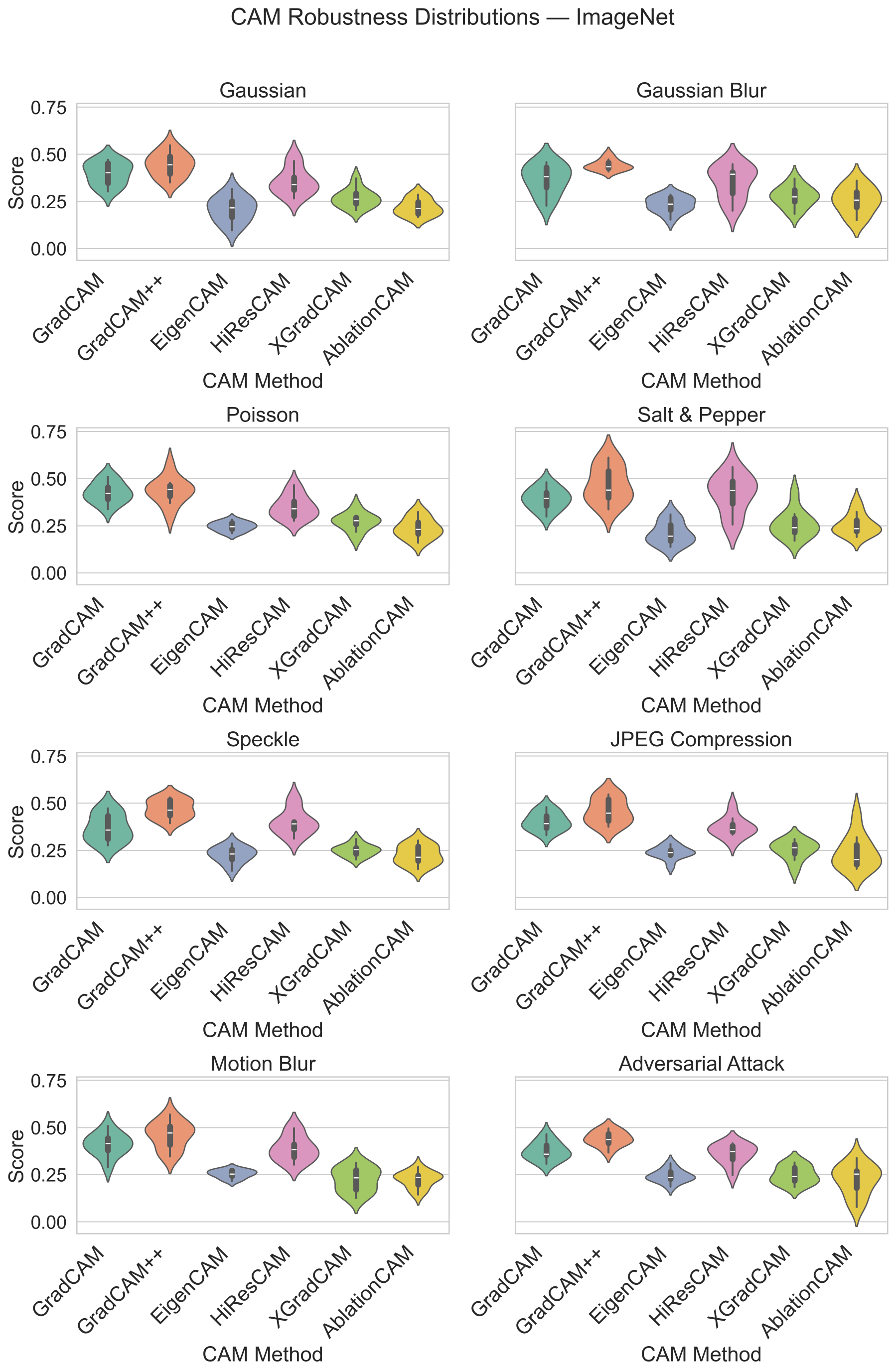}
    \caption{\Rev{RBO distribution across CAM methods for ViT on ImageNet.}}
    \label{fig:vit_violin_imagenet}
\end{figure}

\begin{figure}[H]
    \centering
    \includegraphics[width=0.85\textwidth]{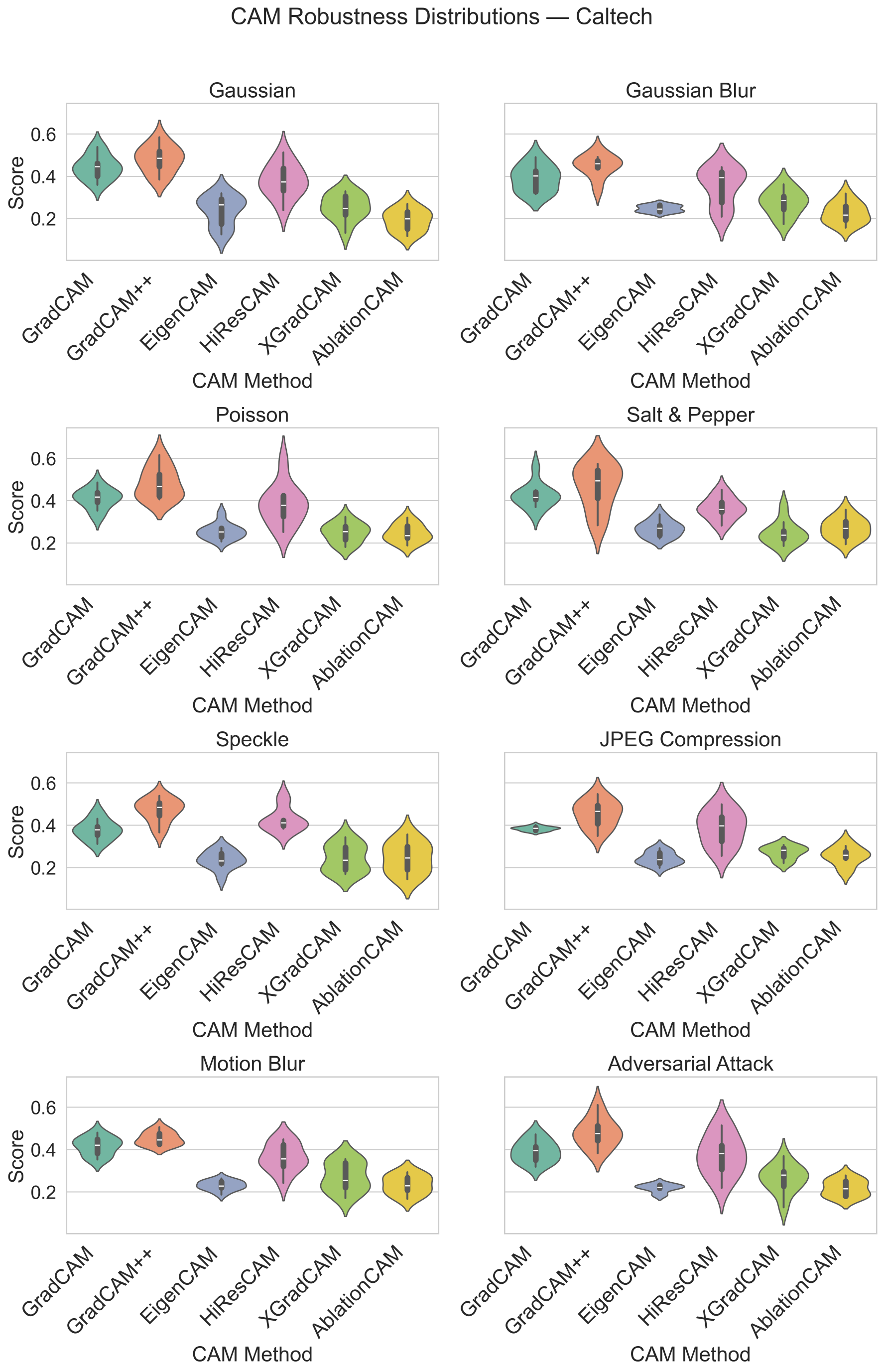}
    \caption{\Rev{RBO distribution across CAM methods for ViT on Caltech.}}
    \label{fig:vit_violin_caltech}
\end{figure}

\begin{figure}[H]
    \centering
    \includegraphics[width=0.85\textwidth]{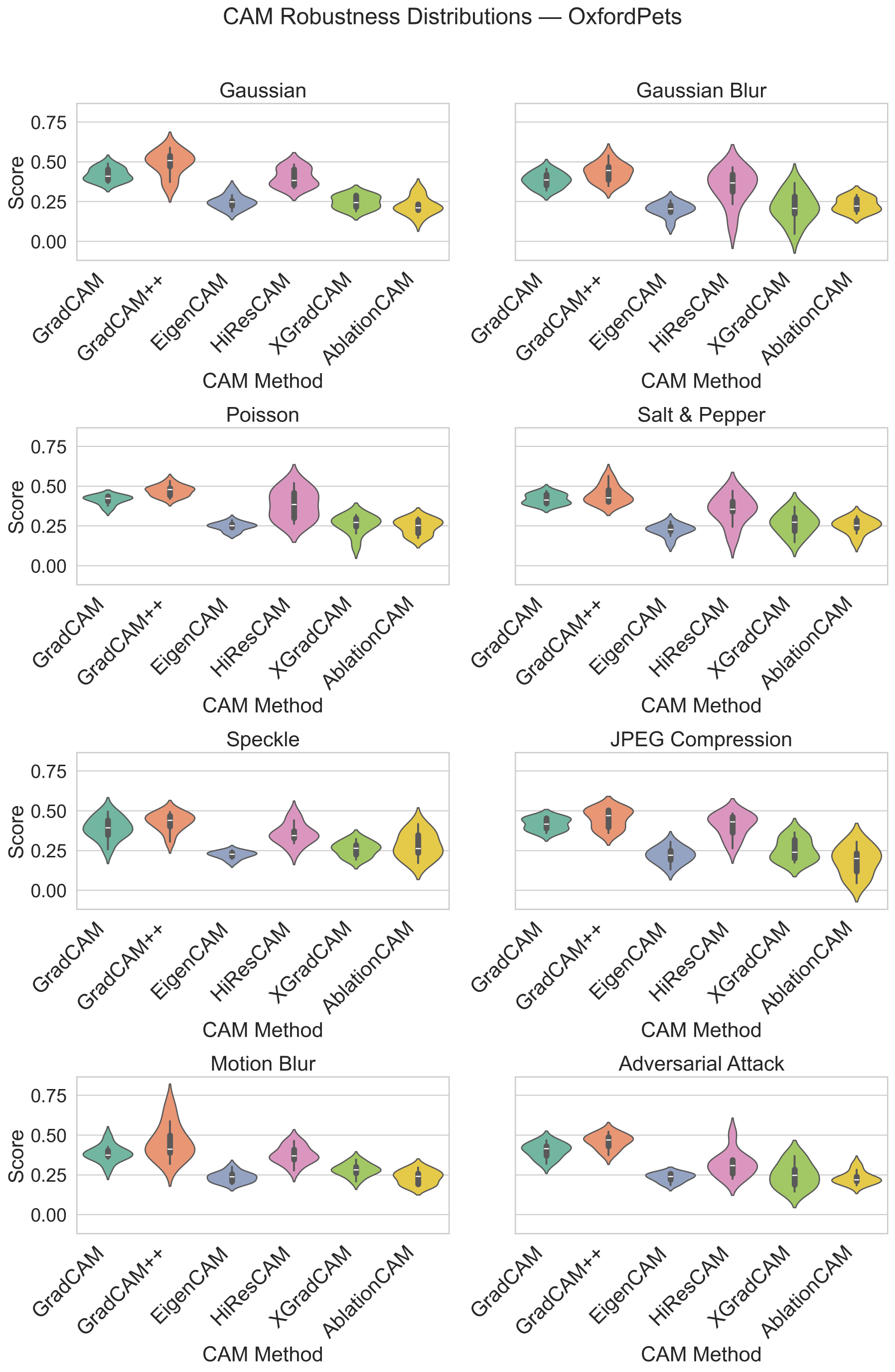}
    \caption{\Rev{RBO distribution across CAM methods for ViT on Oxford Pets.}}
    \label{fig:vit_violin_oxfordpets}
\end{figure}

\begin{figure}[H]
    \centering
    \includegraphics[width=0.85\textwidth]{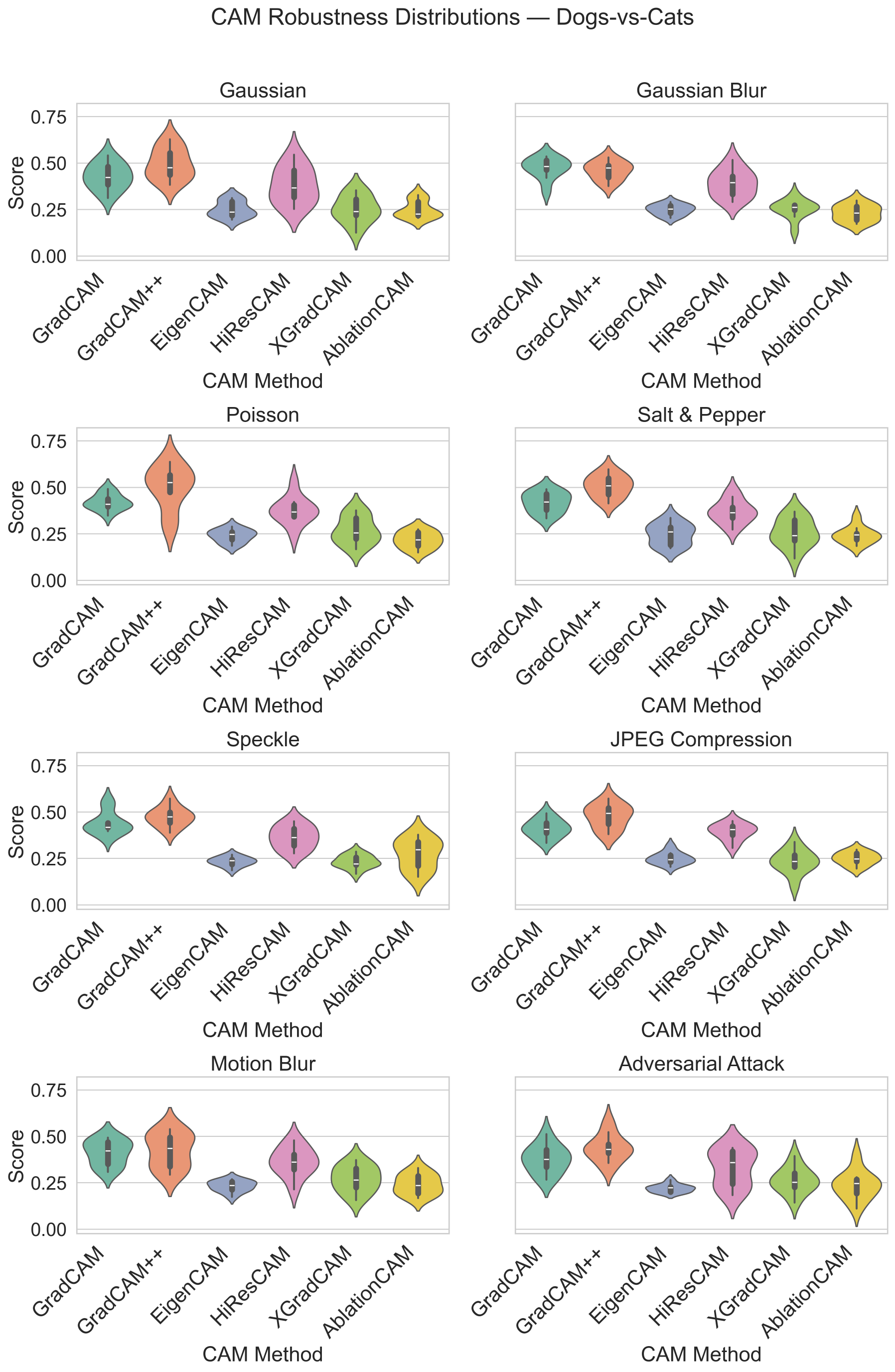}
    \caption{\Rev{RBO distribution across CAM methods for ViT on Dogs-vs-Cats.}}
    \label{fig:vit_violin_dogsvscats}
\end{figure}

\begin{figure}[H]
    \centering
    \includegraphics[width=0.85\textwidth]{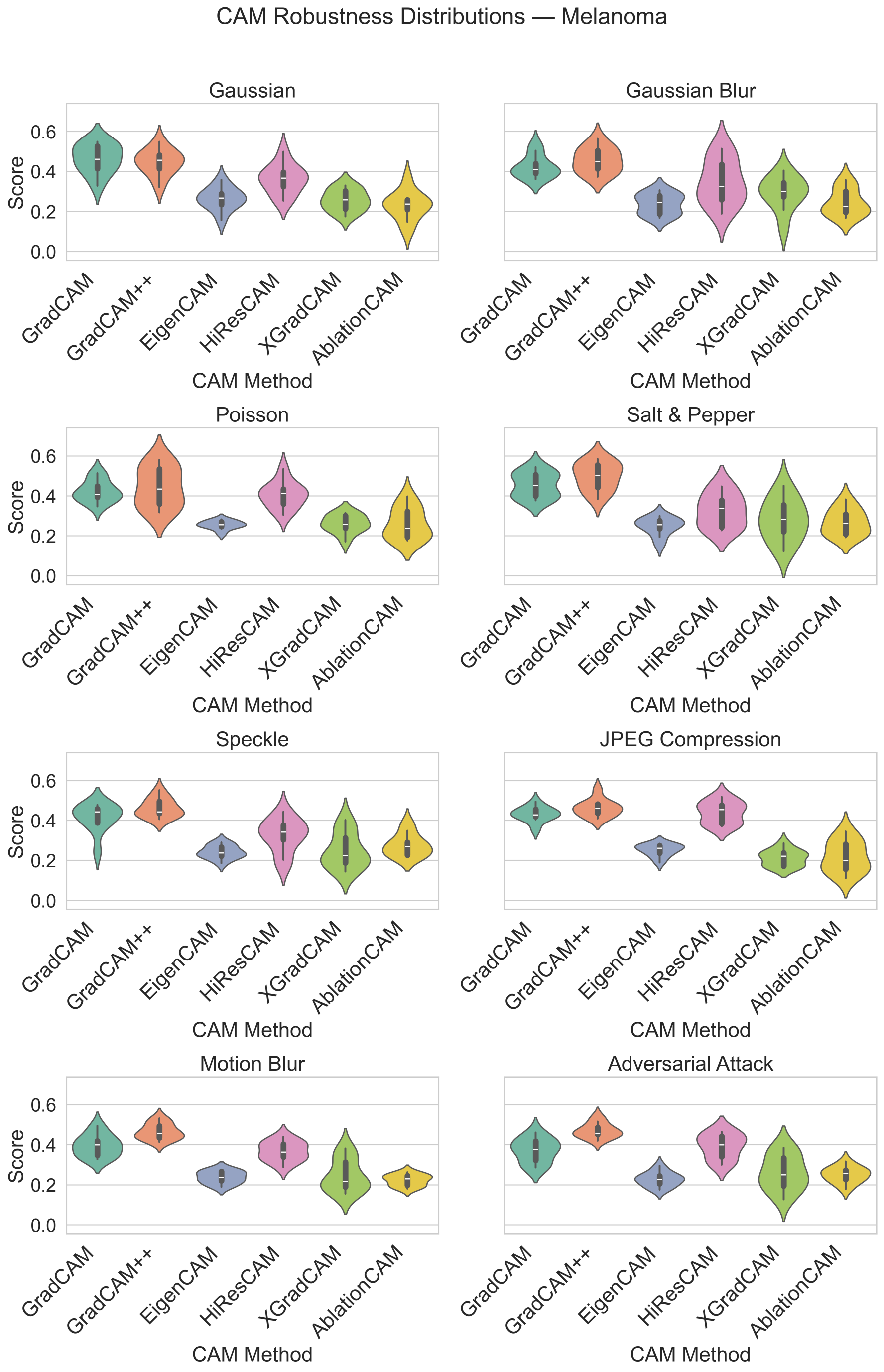}
    \caption{\Rev{RBO distribution across CAM methods for ViT on Melanoma.}}
    \label{fig:vit_violin_melanoma}
\end{figure}

\Rev{Each violin plot corresponds to a specific model evaluated across five datasets. In total, we present \textbf{20 violin plots} for four models \textit{(ResNet50, VGG19, InceptionV3, ViT)} evaluated on five datasets \textit{(ImageNet, Caltech, Oxford Pets, Dogs-vs-Cats, Melanoma)}.
These plots allow for intuitive visual comparison of:
\begin{itemize}
\item CAM method robustness within each model-dataset pair.
\item The variability in robustness due to dataset complexity and model architecture.
\item Outliers and shifts in the distribution under different noise types.
\end{itemize}
\textbf{Discussion.}
As observed from the violin plots in Figures~\ref{fig:resnet50_violin_imagenet}–\ref{fig:vit_violin_melanoma}, we note the following key patterns:
\begin{itemize}
\item \textbf{ViT-based CAMs are more unstable.} The ViT model consistently shows greater variance in RBO scores across all five datasets, indicating a higher sensitivity of CAMs to input perturbations. This suggests that saliency explanations generated from ViT architectures are inherently less stable than those from convolution-based models.
\item \textbf{EigenCAM and AblationCAM exhibit lower robustness.} Across all models (ResNet50, VGG19, InceptionV3, and ViT), EigenCAM and AblationCAM generally record lower median RBO values, reflecting weaker consistency under perturbations.
\item \textbf{GradCAM++ demonstrates the highest robustness.} In contrast, GradCAM++ consistently yields higher RBO scores across datasets and models, indicating that it is the most robust CAM method in terms of stability against noise and distortions.
\item \textbf{Architecture influences robustness trends.} While overall trends are preserved across models, the degree of variability differs. For instance, InceptionV3 shows more consistent scores across CAM methods compared to VGG19 or ViT, hinting at the interaction between model architecture and saliency robustness.
\end{itemize}
Together, these results underscore the importance of evaluating explanation stability across both CAM methods and architectures, with GradCAM++ emerging as a more reliable method under perturbations, and ViT-based models requiring caution due to their inherent instability in CAM generation.}

% \begin{tcolorbox}[colframe=blue, colback=lightblue, boxrule=1pt, sharp corners, enhanced jigsaw]
% \end{tcolorbox}

% \begin{tcolorbox}[colframe=blue, colback=lightblue, boxrule=1pt, sharp corners, enhanced jigsaw]
% \end{tcolorbox}

% \begin{tcolorbox}[colframe=blue, colback=lightblue, boxrule=1pt, sharp corners, enhanced jigsaw]
% \end{tcolorbox}

% \begin{tcolorbox}[colframe=blue, colback=lightblue, boxrule=1pt, sharp corners, enhanced jigsaw]
% \end{tcolorbox}

% \begin{tcolorbox}[colframe=blue, colback=lightblue, boxrule=1pt, sharp corners, enhanced jigsaw]
% \end{tcolorbox}

\clearpage

% \begin{tcolorbox}[colframe=blue, colback=lightblue, boxrule=1pt, sharp corners, enhanced jigsaw]
% \end{tcolorbox}

% \begin{tcolorbox}[colframe=blue, colback=lightblue, boxrule=1pt, sharp corners, enhanced jigsaw]
% \end{tcolorbox}

% \begin{tcolorbox}[colframe=blue, colback=lightblue, boxrule=1pt, sharp corners, enhanced jigsaw]
% \end{tcolorbox}

% \begin{tcolorbox}[colframe=blue, colback=lightblue, boxrule=1pt, sharp corners, enhanced jigsaw]
% \end{tcolorbox}

% \begin{tcolorbox}[colframe=blue, colback=lightblue, boxrule=1pt, sharp corners, enhanced jigsaw]
% \end{tcolorbox}

\clearpage

% \begin{tcolorbox}[colframe=blue, colback=lightblue, boxrule=1pt, sharp corners, enhanced jigsaw]
% \end{tcolorbox}

% \begin{tcolorbox}[colframe=blue, colback=lightblue, boxrule=1pt, sharp corners, enhanced jigsaw]
% \end{tcolorbox}

% \begin{tcolorbox}[colframe=blue, colback=lightblue, boxrule=1pt, sharp corners, enhanced jigsaw]
% \end{tcolorbox}

% \begin{tcolorbox}[colframe=blue, colback=lightblue, boxrule=1pt, sharp corners, enhanced jigsaw]
% \end{tcolorbox}

% \begin{tcolorbox}[colframe=blue, colback=lightblue, boxrule=1pt, sharp corners, enhanced jigsaw]
% \end{tcolorbox}

\clearpage

% \begin{tcolorbox}[colframe=blue, colback=lightblue, boxrule=1pt, sharp corners, enhanced jigsaw]
% \end{tcolorbox}

% \begin{tcolorbox}[colframe=blue, colback=lightblue, boxrule=1pt, sharp corners, enhanced jigsaw]
% \end{tcolorbox}

% \begin{tcolorbox}[colframe=blue, colback=lightblue, boxrule=1pt, sharp corners, enhanced jigsaw]
% \end{tcolorbox}

% \begin{tcolorbox}[colframe=blue, colback=lightblue, boxrule=1pt, sharp corners, enhanced jigsaw]
% \end{tcolorbox}

% \begin{tcolorbox}[colframe=blue, colback=lightblue, boxrule=1pt, sharp corners, enhanced jigsaw]
% \end{tcolorbox}

\clearpage

\subsection{\Rev{Segmentation Method Parameters}}
\label{appendix:segmentation_parameters}

\Rev{In our experiments, we used three commonly adopted image segmentation algorithms: Felzenszwalb, SLIC, and QuickShift. To ensure a comparable number of segments across different methods, we carefully selected the parameters listed in Table~\ref{tab:segmentation_parameters}. These settings were chosen based on empirical observations and the need to balance spatial coherence and segment granularity.}

\begin{table}[htbp]
\centering
\caption{\Rev{Parameters used for different segmentation methods}}
\label{tab:segmentation_parameters}
% \begin{tcolorbox}[colframe=blue, colback=lightblue, boxrule=1pt, sharp corners, enhanced jigsaw, width = 0.8\textwidth]
\begin{tabular}{|c|c|c|}
\hline
\textbf{Segmentation Method} & \textbf{Parameter} & \textbf{Value} \\
\hline
\multirow{3}{*}{Felzenszwalb} 
  & \texttt{scale}         & 100 \\
  & \texttt{sigma}         & 0.5 \\
  & \texttt{min\_size}     & 50 \\
\hline
\multirow{4}{*}{SLIC} 
  & \texttt{n\_segments}   & 120 \\
  & \texttt{compactness}   & 10.0 \\
  & \texttt{sigma}         & 1.0 \\
  & \texttt{start\_label}  & 0 \\
\hline
\multirow{3}{*}{QuickShift} 
  & \texttt{kernel\_size}  & 10 \\
  & \texttt{max\_dist}     & 200 \\
  & \texttt{ratio}         & 0.5 \\
\hline
\end{tabular}
% \end{tcolorbox}
\end{table}

\Rev{These parameter settings ensured that all segmentation approaches produced a reasonably similar number of regions for fair comparative analysis in our CAM robustness evaluation.}

\subsection{\Rev{Calculation of Kendall’s W}}
\label{sec:kendellW}
\Rev{To quantify the agreement among different rank correlation metrics (RBO, Kendall’s $\tau$, and Spearman’s $\rho$) for the robustness ranking of CAM methods under each perturbation type, we compute Kendall’s coefficient of concordance (W). This statistic reflects the degree of consensus among raters (here, ranking metrics), ranging from 0 (no agreement) to 1 (complete agreement).
Let:
\begin{itemize}
    \item $m$ = number of raters (here, 3 ranking metrics),
    \item $n$ = number of items being ranked (here, 6 CAM methods),
    \item $R_{ij}$ = rank of CAM method $j$ by rater $i$,
    \item $\bar{R}_j$ = average rank for item $j$ across raters,
    \item $\bar{R}$ = grand mean of all ranks.
\end{itemize}
Kendall’s W is computed as:
\[
W = \frac{12 \sum_{j=1}^{n} (\bar{R}_j - \bar{R})^2}{m^2(n^3 - n)}
\]
For each perturbation type (Gaussian Blur, Motion Blur, Adversarial Attack), we construct a matrix of shape $m \times n$ (i.e., $3 \times 6$), where each row contains the ranking of CAM methods according to RBO, Kendall’s $\tau$, and Spearman’s $\rho$.
The computed Kendall’s W values are:
\begin{itemize}
    \item \textbf{Gaussian Blur:} $W = 0.97$, $p = 0.0003$
    \item \textbf{Motion Blur:} $W = 1.00$, $p = 0.0001$
    \item \textbf{Adversarial Attack:} $W = 1.00$, $p = 0.0001$
\end{itemize}
These high values indicate strong agreement across ranking metrics, validating the consistency and reliability of the CAM method rankings.}

\subsection{\Rev{Evaluation on Probabilistic CAMs}}

\Rev{We extend our evaluation framework to include stochastic and probabilistic CAM methods to assess their robustness under noise. VarGrad~\cite{adebayo2018sanity} captures uncertainty in saliency by computing the variance of gradients over multiple noisy forward passes. SmoothGrad~\cite{smilkov2017smoothgrad} averages gradients across noise-perturbed inputs to enhance visual coherence. CAPE~\cite{hegde2022probabilistic} models the saliency map as a probability distribution, enabling uncertainty-aware CAM generation.}

\Rev{While our method remains deterministic in design, it is agnostic to the CAM generation strategy. We apply the same robustness pipeline to evaluate RBO scores for both deterministic (e.g., GradCAM~\cite{selvaraju2017grad}) and stochastic CAMs. This allows fair comparison and extends our evaluation protocol to uncertainty-aware explanations. As shown in Table~\ref{tab:probabilistic-cam-extended}, the probabilistic CAM methods \textit{SmoothGrad}, \textit{VarGrad}, and \textit{CAPE} consistently exhibit higher robustness scores than \textit{EigenCAM} across both natural and adversarial perturbations. In most cases, their performance closely matches that of \textit{GradCAM++}, the most robust deterministic method identified in earlier analyses. This trend confirms two important insights:
\begin{itemize}
\item Effectiveness of Probabilistic CAMs: Stochastic approaches that aggregate multiple noisy passes (SmoothGrad, VarGrad) or model saliency as a distribution (CAPE) inherently offer more stable and noise-tolerant explanations.
\item Generalizability of the Evaluation Framework: The proposed robustness metric based on RBO reliably extends to stochastic methods without any modification. This highlights its applicability as a model- and method-agnostic tool for benchmarking explanation robustness across a wide range of CAM generation techniques.
\end{itemize}
Overall, this demonstrates that our framework effectively distinguishes between stable and unstable explanations not only for traditional CAM methods but also for uncertainty-aware and probabilistic approaches, providing a comprehensive way for the study of robustness for future CAM methods.}

\begin{table}[H]
\centering
\caption{\Rev{Robustness Metric ($Consistency \times Responsiveness$) for Stochastic and Traditional CAM Methods for ResNet50 model}}
% \begin{tcolorbox}[colframe=blue, colback=lightblue, boxrule=1pt, sharp corners, enhanced jigsaw]
\label{tab:probabilistic-cam-extended}
\resizebox{\textwidth}{!}{%
\begin{tabular}{llccccc}
\toprule
\textbf{Dataset} & \textbf{Noise Type} & \textbf{SmoothGrad} & \textbf{VarGrad} & \textbf{CAPE} & \textbf{GradCAM++} & \textbf{EigenCAM} \\
\midrule
ImageNet & Gaussian         & 0.374 ± 0.04 & 0.366 ± 0.05 & 0.384 ± 0.05 & 0.419 ± 0.05 & 0.236 ± 0.03 \\
         & Gaussian Blur    & 0.372 ± 0.03 & 0.363 ± 0.04 & 0.380 ± 0.04 & 0.412 ± 0.04 & 0.238 ± 0.02 \\
         & Motion Blur      & 0.368 ± 0.03 & 0.359 ± 0.03 & 0.376 ± 0.04 & 0.414 ± 0.06 & 0.229 ± 0.05 \\
         & JPEG Compression & 0.371 ± 0.04 & 0.361 ± 0.04 & 0.378 ± 0.03 & 0.418 ± 0.05 & 0.235 ± 0.02 \\
         & FGSM (Adv.)      & 0.364 ± 0.05 & 0.353 ± 0.05 & 0.372 ± 0.04 & 0.411 ± 0.04 & 0.224 ± 0.06 \\
         & PGD (Adv.)       & 0.365 ± 0.05 & 0.350 ± 0.05 & 0.369 ± 0.03 & 0.411 ± 0.04 & 0.224 ± 0.06 \\
\midrule
Melanoma & Gaussian         & 0.386 ± 0.06 & 0.374 ± 0.05 & 0.392 ± 0.05 & 0.431 ± 0.05 & 0.246 ± 0.06 \\
         & Gaussian Blur    & 0.380 ± 0.03 & 0.368 ± 0.04 & 0.386 ± 0.04 & 0.423 ± 0.04 & 0.243 ± 0.05 \\
         & Motion Blur      & 0.374 ± 0.04 & 0.362 ± 0.04 & 0.380 ± 0.04 & 0.417 ± 0.06 & 0.236 ± 0.05 \\
         & JPEG Compression & 0.379 ± 0.06 & 0.365 ± 0.04 & 0.386 ± 0.03 & 0.426 ± 0.05 & 0.242 ± 0.04 \\
         & FGSM (Adv.)      & 0.371 ± 0.05 & 0.359 ± 0.04 & 0.376 ± 0.04 & 0.412 ± 0.04 & 0.231 ± 0.03 \\
         & PGD (Adv.)       & 0.371 ± 0.05 & 0.356 ± 0.05 & 0.373 ± 0.03 & 0.412 ± 0.04 & 0.231 ± 0.03 \\
\bottomrule
\end{tabular}%
}
% \end{tcolorbox}
\end{table}

\subsection{\Rev{Performance on Adversarially Trained Models}}

\Rev{While adversarial training~\cite{madry2018towards} is a well-established strategy for improving classification robustness, its impact on CAM robustness remains underexplored. The proposed framework evaluates CAM robustness independently of the model training procedure and is thus fully compatible with adversarially trained models. We tested our method on a ResNet-50 model trained adversarially on ImageNet and measured the RBO-based robustness scores across various noise types.}

\Rev{Table~\ref{tab:adv-rbo} reports these results. Despite increased classification robustness, CAM explanations remain sensitive to perturbations, highlighting the importance of dedicated metrics for explanation stability. Notably, the robustness metric under adversarial perturbations is significantly lower than for other noise types, even though the robustness scores across various CAM methods under standard corruptions (such as Gaussian, Poisson, and Blur) are roughly comparable to those from the standard (non-adversarially trained) ResNet-50 model. This suggests that CAM methods, when applied to adversarially trained models, are less vulnerable to explanation degradation under adversarial attacks.}

\begin{table}[H]
\centering
\caption{\Rev{Robustness Score ($Consistency \times Responsiveness$) for different CAMs under Different Noise Types on \textit{Adversarially Trained ResNet-50}} (ImageNet)}
\label{tab:adv-rbo}
% \begin{tcolorbox}[colframe=blue, colback=lightblue, boxrule=1pt, sharp corners, enhanced jigsaw]
\resizebox{\textwidth}{!}{
\begin{tabular}{lcccccc}
\toprule
\textbf{Noise Type} & \textbf{AblationCAM} & \textbf{EigenCAM} & \textbf{GradCAM} & \textbf{GradCAM++} & \textbf{HiResCAM} & \textbf{XGradCAM} \\
\midrule
Gaussian            & $0.210 \pm 0.03$ & $0.215 \pm 0.02$ & $0.374 \pm 0.04$ & $0.419 \pm 0.05$ & $0.344 \pm 0.06$ & $0.236 \pm 0.03$ \\
Gaussian Blur       & $0.213 \pm 0.04$ & $0.216 \pm 0.06$ & $0.372 \pm 0.03$ & $0.412 \pm 0.04$ & $0.339 \pm 0.05$ & $0.238 \pm 0.02$ \\
Poisson             & $0.216 \pm 0.03$ & $0.220 \pm 0.05$ & $0.375 \pm 0.06$ & $0.417 \pm 0.03$ & $0.347 \pm 0.03$ & $0.237 \pm 0.04$ \\
Salt \& Pepper      & $0.229 \pm 0.04$ & $0.218 \pm 0.04$ & $0.378 \pm 0.05$ & $0.421 \pm 0.06$ & $0.346 \pm 0.05$ & $0.239 \pm 0.05$ \\
Speckle             & $0.218 \pm 0.05$ & $0.214 \pm 0.06$ & $0.370 \pm 0.03$ & $0.416 \pm 0.03$ & $0.341 \pm 0.04$ & $0.231 \pm 0.03$ \\
JPEG Compression    & $0.212 \pm 0.06$ & $0.216 \pm 0.03$ & $0.371 \pm 0.04$ & $0.418 \pm 0.05$ & $0.343 \pm 0.05$ & $0.235 \pm 0.02$ \\
Motion Blur         & $0.208 \pm 0.03$ & $0.213 \pm 0.04$ & $0.368 \pm 0.03$ & $0.414 \pm 0.06$ & $0.336 \pm 0.04$ & $0.229 \pm 0.05$ \\
Adversarial Attack  & $0.203 \pm 0.10$ & $0.211 \pm 0.08$ & $0.364 \pm 0.08$ & $0.411 \pm 0.07$ & $0.332 \pm 0.06$ & $0.224 \pm 0.09$ \\
\bottomrule
\end{tabular}
}
% \end{tcolorbox}
\end{table}

\end{document}